\documentclass[review]{elsarticle}

\usepackage[dvipsnames]{xcolor}

\usepackage{amsmath}
\usepackage{amssymb}
\usepackage{amstext}
\usepackage{appendix}
\usepackage{booktabs}
\usepackage{graphicx}
\usepackage{lineno}
\usepackage{mathtools}
\usepackage{multirow}
\usepackage{natbib}
\usepackage{setspace}
\usepackage{tabularx}
\usepackage{epstopdf}
\usepackage{pdfpages}
\usepackage{pdftexcmds}
\usepackage{longtable}
\usepackage{chngcntr}
\usepackage{mathrsfs}
\usepackage{euscript}

\modulolinenumbers[1]

\journal{Computers and Industrial Engineering}

\biboptions{authoryear}

\begin{document}

\begin{frontmatter}
\singlespacing
\title{Resource-constrained multi-project scheduling \\with activity and time flexibility}
\author[mymainaddress,mysecondaryaddress,mythirdaddress]{Viktoria A. Hauder\corref{mycorrespondingauthor}}
\author[mymainaddress,mythirdaddress]{Andreas Beham}
\author[mymainaddress,mythirdaddress]{Sebastian Raggl}
\author[mysecondaryaddress]{\\Sophie N. Parragh}
\author[mythirdaddress,myfourthaddress]{Michael Affenzeller}
\address[mymainaddress]{Josef Ressel Center for Adaptive Optimization in Dynamic Environments, \\University of Applied Sciences Upper Austria, Hagenberg}
\address[mysecondaryaddress]{Institute of Production and Logistics Management, \\Johannes Kepler University, Linz, Austria}
\address[mythirdaddress]{Heuristic and Evolutionary Algorithms Laboratory, \\University of Applied Sciences Upper Austria, Hagenberg}
\address[myfourthaddress]{Institute of Formal Models and Verification, Johannes Kepler University, Linz, Austria}
\cortext[mycorrespondingauthor]{Corresponding author email address: \texttt{viktoria.hauder@fh-hagenberg.at} \\ The financial support by the Austrian Federal Ministry for Digital and Economic Affairs and the National Foundation for Research, Technology and Development, and the Christian Doppler Research Association is gratefully acknowledged. The work described in this paper was also done within the project Logistics Optimization in the Steel Industry (LOISI) \#855325 within the funding program Smart Mobility 2015 organized by the Austrian Research Promotion Agency (FFG) and funded by the Governments of Styria and Upper Austria. \\ \\ \textit{Former working title: ``On constraint programming for a new flexible project scheduling problem with resource constraints''.}}

\begin{abstract}
Project scheduling in manufacturing environments often requires flexibility in terms of the selection and the exact length of alternative production activities. Moreover, the simultaneous scheduling of multiple lots is mandatory in many production planning applications. To meet these requirements, a new resource-constrained project scheduling problem (RCPSP) is introduced where both decisions (activity flexibility and time flexibility) are integrated. Besides the minimization of makespan, two new alternative objectives are presented: maximization of balanced length of selected activities (time balance) and maximization of balanced resource utilization (resource balance). New mixed integer and constraint programming (CP) models are proposed for the developed integrated flexible project scheduling problem. Benchmark instances on an already existing flexible RCPSP and the newly developed problem are solved to optimality. The real-world applicability of the suggested CP models is shown by additionally solving a large industry case.
\end{abstract}
 
\end{frontmatter}

\newpage

%\linenumbers

\onehalfspacing
\pagenumbering{arabic}

\section{Introduction and motivation}\label{sec:Introduction}
Project scheduling is an essential operational planning area in different business sectors where precedences between activities and the access to limited resources have to be taken into account. Besides a large number of applications such as research and development or software development, one important example is the scheduling of manufacturing activities \citep{artigues2013resource}. The underlying optimization problem is the well-known NP-hard resource-constrained project scheduling problem (RCPSP), characterizing a project where all included activities have to be scheduled in such a way that resource constraints, processing times, and precedence relations are respected. The most common objective is the minimization of the makespan \citep{hartmann2010survey}.

In a manufacturing context, decision-makers sometimes have the flexibility to choose between different manufacturing activities or decide on the exact length of them. Precisely these flexibility descriptions motivate the RCPSP presented in this work, faced by many different manufacturing industries. For the production of multiple lots (=jobs), there are several alternative production activities per lot. Due to existing technological requirements, alternative activities are aggregated to a number of alternative routes per lot. For every lot, one production route has to be selected, an individual delivery date has to be considered and early delivery is not permitted. Besides machines or manpower, also typical logistics renewable resources such as vehicles or intermediate storages (buffers) have to be considered since they are strongly limited in many factories. Every activity of every lot demands at least one scarce resource, e.g. a transporting, storaging or machining activity. Due to the consideration of all existing resources, all activities of one lot have to be sequenced directly one after another. This means that idle times between the activities within one lot are not allowed, since they would result in a temporary ficticious disappearance of the production lot. E.g., if the machining activity of lot 1 is finished, the next one in the precedence relationship, e.g. the storaging activity of lot 1 has to start immediately. However, routes (multiple activities) of different lots can be scheduled in parallel, competing for the available production resources. E.g., the storaging activity of lot 2 can start before the storaging activity of lot 1 is completed. Moreover, different lots use different production activities and the sequence of utilized resources is not identical for all lots. With special regard to logistics activities such as storaging, minimum and maximum allowed processing times are specified for activities. Thus, processing times are of variable length, i.e. the start and the end time per activity have to be decided during the optimization process (time flexibility). As a result, the treated problem integrates the two major decisions of activity selection and processing time determination. In addition, we introduce two new objective functions: maximization of balanced length of activity processing times (time balance), and maximization of balanced resource utilization (resource balance).

The described production process may be found in various production industries. Three different examples the authors know from several cooperations with production managers are the steel-, the food-, and the glass processing industry, to name just a few. The production of steel slabs is typically processed lot-wise and includes several logistics activities, e.g. a storaging activity between necessary cooling, reheating or transporting activities. Moreover, in many steel plants a minimum and maximum allowed processing time is given as it does not influence the quality of such a product whether the storaging activity of one lot is performed for one hour or two days in a slabyard. Furthermore, this time flexibility gives production managers an often demanded additional degree of freedom in their production planning possibilities. The same holds for the glass processing industry where e.g. transporting and storaging activities are necessary in the production process.

In the very different food processing industry, similar patterns can be found. Industrial bakeries for example typically prescribe a minimum purchase quantity for business customers to be able to carry out a lot-wise production. Furthermore, besides recipes for which precise time specifications have to be met, for other ones the production activities are allowed to vary up to one hour or even more, e.g. the cooling down of hot baked goods. For deep-frozen baked goods, there is an even higher time flexibility: the baked good has to be at least stored in a freezer for a certain amount of time but can be stored up to weeks or even longer if necessary. The same examples also apply to forbidden early deliveries. There are business customers who do not give their door keys to carriers for security reasons, and therefore only allow deliveries beginning at their opening hours. Some steel and glass industry customers (e.g. automobile industry) also do not accept early delivery, overall resulting in deliveries without earliness.

Furthermore, objective functions directly or indirectly influence financial burdens of companies, e.g. balancing the working time of all employees increases the acceptance of work schedules and balancing resource allocation avoids idle resources \citep{matl2017workload,rieck2012mixed}. However, to the best of our knowledge, time balancing purposes concerning activities have not been considered yet in the scientific literature on RCPSP. For balancing resource utilization over time, resources are weighted by introducing costs \citep{li2018effective,neumann1999resource} although in many companies, there are often several equivalent resources (e.g. two deep freezing facilities of an industrial bakery or two slab yards of a steel manufacturer) with the only goal of avoiding uneven utilizations. Therefore, the two above presented new objectives are introduced.

The described choices between various alternative activities are known as flexible project scheduling \citep{beck2000constraint,tao2017scheduling}. Related to this, \cite{tao2017scheduling} present the RCPSP with alternative activity chains (RCPSP-AC) and develop a simulated annealing algorithm for solving it. They describe that more efficient solution methods should be investigated for the RCPSP-AC in order to tackle related real-world problems. Moreover, they point out the necessity of an officially available benchmark set for the RCPSP-AC. Motivated by these suggestions, the new problem extensions described above, and the success of constraint programming (CP) for solving scheduling problems \citep{laborie2018ibm,schnell2017generalization}, we develop and solve mixed integer programming (MIP) and CP models for the RCPSP-AC, its new multi-project version (RCMPSP-AC), and a new version featuring multiple projects as well as time flexibility. The latter we denote resource-constrained multi-project scheduling problem with alternative activity chains and time flexibility (RCMPSP-ACTF). To be more precise, the contribution of this work is fivefold:
\begin{itemize}\vspace{-0,3cm}
	\item A new resource-constrained multi-project scheduling problem with alternative activity chains and time flexibility (RCMPSP-ACTF) is introduced and a related MIP model is proposed. It is based on the work of \cite{tao2017scheduling} who consider alternative activity chains in a single-project environment without time flexibility. \vspace{-0,3cm}
	\item Two new objective functions aiming at well-balanced solutions are presented.\vspace{-0,3cm}
	\item CP solution models are developed for the RCPSP-AC, its multi-project version, and the RCMPSP-ACTF.\vspace{-0,3cm}
	\item MIP and CP modeling approaches are compared on newly generated benchmark data sets, illustrating the advantage of using our new CP models: all benchmarks of the RCPSP-AC and many instances of the newly presented problems are solved to optimality. \vspace{-0,3cm}
	\item With the CP based models, a large industry case with more than 600 activities is solved to optimality and the impact of using the proposed alternative objective functions is evaluated.\vspace{-0,2cm}
\end{itemize}
The remainder of this paper is organized as follows. Section \ref{sec:Literature} gives an extensive review of related literature. The developed problem formulation is introduced in Section \ref{sec:MIP}, while Section \ref{sec:CP} provides a detailed description of the created constraint programming solution approach. In Section \ref{sec:ComputationalExperiments}, the computational results for the generated benchmark data and the real-world case study are discussed. Finally, Section \ref{sec:Conclusion} gives concluding remarks and suggestions for further research.

\section{Literature overview}\label{sec:Literature}
The scheduling of manufacturing activities has been extensively investigated in the last decades since its efficient management is of high relevance in practice \citep{russell2019multi}. One area of such scheduling problems are flexible manufacturing systems (FMS), considering flexibility within the production process \citep{blazewicz2019handbook}. Examples for FMS are flexible job shops (FJS), flexible or hybrid flow shops (FFS) or resource-constrained FMS. FJS and FFS typically consider machines as resources. In FJS, different jobs can consist of different activities (=operations) which can be performed on alternative machines \citep{rajabinasab2011dynamic}. In FFS, all jobs consist of the same activities but they can also be executed on alternative machines \citep{ruiz2010hybrid}. For the FJS and FFS, only one job can be handled at a time by each resource, there are no precedence constraints between the jobs and storage capacities (buffers) are assumed to be unlimited (unlimited idle times between activities are allowed) or zero (no idle times are allowed since there is no capacity) \citep{blazewicz2019handbook}. However, as described in Section \ref{sec:Introduction}, in the production environments considered in this work the production activities require additional resources with limited capacities such as storage areas. They all have capacities (i.e. no idle times between activities within one job are allowed since the capacities are modeled by activities that demand them; idle times between different jobs are allowed), and multiple activities and jobs can be handled at a time by each resource. Moreover, there are alternative activities for every job, all of them are in a precedence relationship and they have flexible processing times, overall resulting in a resource-constrained FMS which corresponds to a flexible RCPSP \citep{blazewicz2019handbook} where several scheduling issues have to be newly mastered together.

The RCPSP is a well-researched topic, creating a schedule with starting times for all activities. Basically, there are fixed processing times, activity demands for limited resources, and precedence relations between all activities, typically represented by acyclic activity-on-node (AON) networks \citep{hans2007hierarchical,johnson1979computers}. Besides the standard objective of makespan minimization, also a wide variety of alternative objectives has been studied. An example that is also tackled in this work is resource leveling, where varying resource utilization over time is minimized \citep{li2018effective}. One of the many extensions which is examined in this work is related to time, considering an additional flexibility or a limitation of processing times \citep{artigues2017strength}. Examples related to time extensions are idle times \citep{allahverdi2016survey}, uncertain activity durations \citep{moradi2019robust}, flexible resource usage durations \citep{naber2014mip}, generalized precedence relations (GPR) \citep{schnell2016efficient}, and setup times \citep{vanhoucke2019resource}.

A great variety of exact and heuristic solution methods has already been investigated. As the detailed presentation of algorithmic methods and the many existing extensions of the RCPSP is outside the scope of this work, the subsequent review is limited to flexible and multi-project scheduling and constraint programming. The interested reader is referred to the works of \cite{artigues2013resource,blazewicz2019handbook,hartmann2010survey,schwindt2015handbook} and \cite{weglarz2012project} for an additional comprehensive examination.

\subsection{Flexible and multi-project resource-constrained project scheduling}\label{sec:LiteratureFlexibleMulti}
Flexible project scheduling as a generalization of the RCPSP has proven to be also NP-hard \citep{blazewicz1983scheduling,tao2017scheduling}. It deals with the selection of the best out of multiple alternative activities \citep{beck2000constraint,burgelman2018maximising}. Already \cite{pritsker1966gert} has shown that besides so-called AND nodes which imply the selection of all successor nodes, also OR nodes can be introduced. OR nodes allow flexibility, as one out of multiple existing successor nodes is chosen. \cite{johannes2005complexity} proofed that the minimization of weighted completion times under consideration of OR precedence constraints is already NP-hard with one single resource. \cite{capek2012production} considered an RCPSP with alternative process plans for wire harnesses production. They proposed an integer linear programming (ILP) model and a heuristic algorithm for real-world applicability. \cite{kellenbrink2015scheduling} examined an aircraft turnaround process and proposed a genetic algorithm (GA) for the optimization of the developed flexible project structure. \cite{vanhoucke2016approach} considered bidirectional relations besides AND/OR ones. They developed a satisfiability approach and showed its competitiveness on well-known benchmark datasets. \cite{tao2017scheduling} studied an AND/OR network under consideration of alternative activity chains (RCPSP-AC), which builds the base for the development of our problem formulations. They showed that their RCPSP-AC is a generalization of the multi-mode RCPSP (MRCPSP) and proposed a simulated annealing procedure. \cite{tao2018multi} extended the RCPSP-AC by integrating it into a bi-objective MRCPSP. They solved the new problem with a hybrid metaheuristic, consisting of a tabu search procedure and the NSGA II and compared it with the solutions generated with an exact solver. \cite{tao2018stochastic} introduced an RCPSP with hierarchical alternatives and stochastic activity durations and proposed a metaheuristic framework consisting of sample average approximation and an evolutionary algorithm. \cite{burgelman2018maximising} presented a new flexible MRCPSP under maximization of weighted alternative execution modes and proposed different ILP formulations. \cite{servranckx2019tabu} proposed an RCPSP with alternative subgraphs and solved it with a tabu search. Moreover, they extended the problem by introducing different strategies for the consideration of uncertainty for this new problem \citep{servranckx2019strategies}. \cite{birjandi2019optimizing} introduced a new nonlinear MIP model for an RCPSP with multiple routes (RCPSP-MR) and solved it with a hybrid metaheuristic based on particle swarm optimization (PSO) and a GA. \cite{birjandi2019fuzzy} presented a fuzzy extension of the RCPSP-MR with flexible activities under uncertain conditions and proposed a hybrid approach consisting of distribution rules, PSO and a GA.

The resource-constrained multi-project scheduling problem (RCMPSP) as another generalization of the RCPSP considers a set of multiple projects $l \in \{1,...,L\}$ and activities $j \in N_l=\{1,...,n+1\}$ per project \citep{hartmann2010survey,lova2000multicriteria}. There are two possible ways for the representation of this multi-project variant. Through the single-project (SP) approach, all projects are cumulated into an AON network with one common dummy source and sink node \citep{lova2001analysis}. \cite{pritsker1969multiproject} were the first to suggest an additional sink node per project for the consideration of a due date per project within the SP approach. With the multi-project (MP) approach, source and sink nodes for every project are considered, and the connecting elements are commonly shared resources. A recent example is the one of \cite{asta2016combining} who worked on a multi-mode RCMPSP under consideration of the SP and the MP approach and proposed a combination of monte-carlo and hyper-heuristic methods. \cite{chakrabortty2017resource} suggested an evolutionary local search method based on priority rules for the MP approach. An already existing example for a manufacturing application of an SP approach for the RCMPSP is the work of \cite{voss2007hybrid}. They modeled a hybrid flow shop scheduling problem as a multi-mode RCMPSP. Stages with variable capacities, waiting times between stages and renewable resources were considered. All existing stages must be traversed and different resources must be used. In their work, every machine only processed one job at a time, production routes were identical for every job and processing times were fixed. The objectives were the minimization of weighted tardiness and a maximized resource utilization. They applied dispatching rules for solving large real-world instances.

\subsection{Scheduling and constraint programming}\label{sec:LiteratureCP}
Constraint programming is a powerful optimization technique especially for combinatorial problems and thus, also for scheduling and real-world problems \citep{baptiste2012constraint}. \cite{bockmayr2005constraint} presented the general functionality of CP. They showed similarities of CP and MIP methods such as the generation of branching trees. One dissimilarity is constraint propagation, which is part of CP and removes all values from domains, which cannot take part in any feasible solution. CP has already been efficiently applied to different domains such as the deficiency problem \citep{altinakar2016comparison}, project driven manufacturing \citep{banaszak2009constraint}, ship scheduling and inventory management \citep{goel2015constraint}, operating theatres \citep{wang2015scheduling}, resource-constrained FMS \citep{novas2014integrated}, and the RCPSP \citep{liess2008constraint}. Recent works also showed the compatibility of the CP methodology with the RCPSP. \cite{schnell2016efficient,schnell2017generalization} developed and analyzed exact algorithms, Boolean Satisfiability Solving and CP approaches for the MRCPSP with GPR and presented new best solutions. \cite{kreter2017using} developed new CP models and a special propagator for the RCPSP with general temporal constraints and calendar constraints and provided optimal solutions for all benchmarks sets. \cite{kreter2018mixed} developed new MIP and CP models for resource availability cost problems and solved all open benchmarks to optimality.

\subsection{Identified research gap}\label{sec:LiteratureGap}
Considering the described literature, the presented scheduling problem in Section \ref{sec:Introduction} can be modeled as a resource-constrained multi-project scheduling problem since precedence relations, limited renewable resources and multiple lots (projects) have to be taken into account. Moreover, flexibility concerning activity selection like in \cite{tao2017scheduling} and activity processing times has to be incorporated. To the best of our knowledge, the integration of multiple projects and flexible processing times has not been considered yet in an RCPSP context. As the already described work of \cite{tao2017scheduling} on the RCPSP-AC builds the base for the development of our work, we call our new problem the resource-constrained \textit{multi-project} scheduling problem with \textit{alternative activity chains} and \textit{time flexibility} (RCMPSP-ACTF). Two new objectives for the RCMPSP-ACTF, minimizing different processing time buffers and peak usages of resources are introduced and the necessary additional constraints are presented. CP models are developed for the RCPSP-AC, its new multi-project extension (RCMPSP-AC) and the new RCMPSP-ACTF and they are tested on newly generated benchmark data. The developed models also provide decision support for a steel industry company partner, illustrated by the presented real-world case study in this work.

\section{Problem definition}\label{sec:MIP}
In this section, the MIP formulations for the new RCMPSP-ACTF are presented. Section \ref{MIP-Notations} provides basic assumptions and necessary notations, including an exemplarily AON network that is used to model the introduced problem. In Sections \ref{MIP-makespan} - \ref{MIP_resourcebalance}, we formally define the RCMPSP-ACTF and the two new objectives alongside the necessary constraints for time and resource balance.

\subsection{Notations}\label{MIP-Notations}
\begin{figure}
	\centering\includegraphics[width=0.7\textwidth]{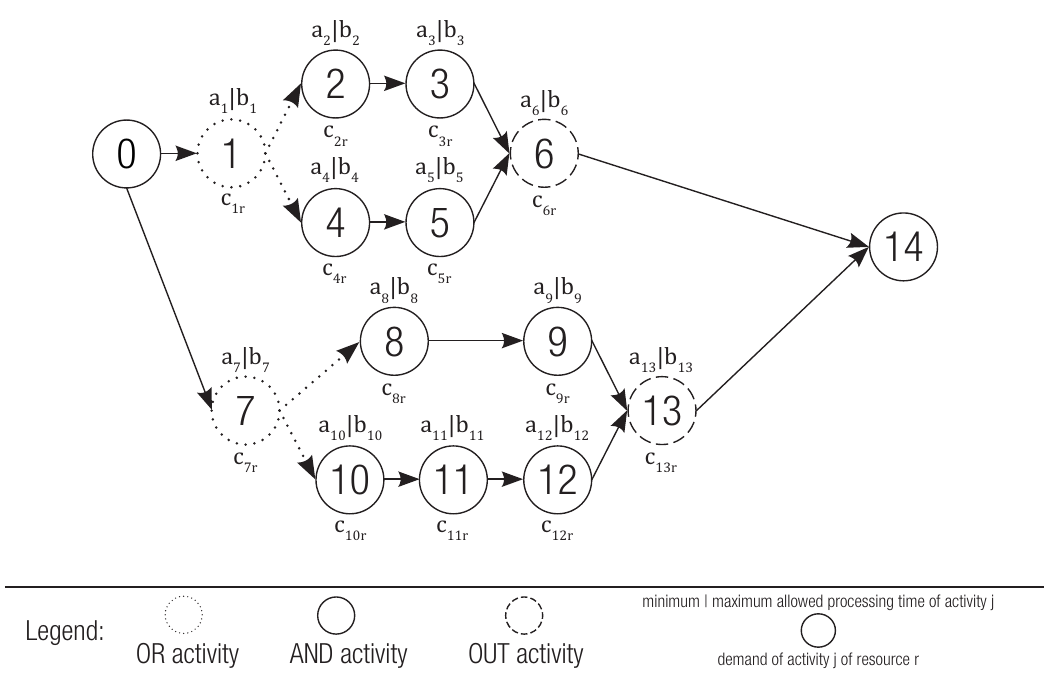}
	\caption{Example AND/OR/OUT network for the RCMPSP-ACTF.}
	\label{FIG_Illustrative-AON_RCMPSP-ACTF}
\end{figure}

For the RCMPSP-ACTF, all operations which have to be performed are called activities (or tasks or nodes) and are distinguished from the term job (or lot or project), which corresponds to a customer order. Precedence relations are defined using an AON project network. Every activity $i,j \in \{0,...,n+1\}$ has a specific task, except nodes $0$ and $n+1$, which are dummy production process source and sink nodes. ``Activity 1'' could for example be the heating and ``activity 2'' the transportation of a manufactured product. For the consideration of one sink node per lot and thus, per project, the subset $\mathcal{L} \subseteq \mathcal{J}$ is introduced. Every sink node corresponds to one customer delivery activity and therefore represents the completion of one lot. This kind of consideration complies with the well-known SP approach with one sink node per project \citep{lova2001analysis,pritsker1969multiproject} as explained in Section \ref{sec:Literature}. Related to this, we need a mandatory new activity type $p_j=2$ which we call OUT activity in addition to the two activity types AND/OR ($p_j=1 / p_j=0$) introduced by \cite{tao2017scheduling}. It ensures that idle times between all activities within one lot are forbidden (since all intermediate storages have to be modeled as resources due to their scarcity in manufacturing environments) and it guarantees that no additional activities of one lot appear in the production schedule if they do not belong to the chosen alternative route of the optimization (see the following Section \ref{MIP-makespan} for a detailed description). Related to this, all activities have minimum and maximum allowed processing times $a_i$ and $b_i$ now. The complete list of notations is given in Table \ref{TAB_MIPnotations}. In order to illustrate the new RCMPSP-ACTF, we give an example AON network in Figure \ref{FIG_Illustrative-AON_RCMPSP-ACTF}, demonstrating the alternative activity chains and time flexibility for multiple projects. Nodes 0 and 14 are the dummy source and sink nodes with zero processing times and no demands. With nodes 1 and 7, the production of the respective lots is started and with customer delivery activities 6 and 13, it is finished, i.e. every lot has its own start node and a customer delivery activity.

\begin{table}[!h]
	\caption{MIP model notations.}
	\vspace{1.0ex}
	\label{TAB_MIPnotations}
	\begin{tabularx}{\textwidth}{lX}
		\hline
		\multicolumn{2}{l}{\textbf{Indices}}	
		%\rule{0pt}{15pt}
		\\
		$i,j$ 		& 	Activities 	
		\\
		%$l$			&	Lot
		%\\
		$t,\tau$	&	Time periods
		\\
		$r$			&	Resource	
		\\
		\multicolumn{2}{l}{\textbf{Sets}}	
		\\
		$\mathcal{J}$	&	Set of activities $i,j \in \{0,1,..,n+1\}$
		\\
		$\mathcal{L}$	&	Set of delivery activities $\mathcal{L} \subseteq \mathcal{J}$
		\\
		$\mathcal{T}$	&	Set of time slots $t,\tau \in \{0,1,..,T\}$
		\\
		$\mathcal{R}$	&	Set of renewable resources $r \in \{0,1,..,R\}$
		\\
		$\mathcal{R^*}$	&	Set of renewable resources $\mathcal{R^*} \subseteq \mathcal{R}$ considered for resource balance
		\\
		\multicolumn{2}{l}{\textbf{Parameters}}	
		\\	
		$A_{ij}$	&	Activity-adjacency matrix $A_{ij}=1$ if $j$ is the direct successor of $i$ and 0 otherwise
		\\
		$p_j$		&	Node type $p_j=1$ if activity $j$ is an AND activity, $p_j=0$ if activity $j$ is an OR activity, $p_j=2$ if activity $j$ is an OUT activity
		\\
		$a_j$		&	Minimum allowed processing time (duration) of activity $j$
		\\
		$b_j$		&	Maximum allowed processing time (duration) of activity $j$
		\\
		$d_j$		&	Delivery time (due date) of activity $j \in \mathcal{L}$
		\\
		$C_r$		&	Capacity of resource $r$
		\\
		$c_{jr}$	&	Demand of activity $j$ of resource $r$
		\\
		$M$			&	Big M (large constant)
		\\
		\multicolumn{2}{l}{\textbf{Decision variables}}
		\\
		$x_j$		&	=1 if activity $j$ is selected and 0 otherwise
		\\
		$s_{jt}$	&	=1 if activity $j$ is selected to start at time slot $t$ and 0 otherwise
		\\
		$w_{jt}$	&	=1 if activity $j$ is selected to be worked on (processed) at time slot $t$ and 0 otherwise
		\\
		$y_{jt}$	&	=1 if activity $j$ is selected to be completed at time slot $t$ and 0 otherwise
		\\
		$B$			&	Largest time buffer of the whole network
		\\
		$S$			&	Smallest time buffer of the whole network
		\\
		$u_r$		&	Maximum resource utilization (peak usage) of resource $r \in \mathcal{R^*}$
		\\
		\hline
	\end{tabularx}
\end{table}

\subsection{RCMPSP-ACTF MIP model: makespan minimization}\label{MIP-makespan}
In the following, we present the new multi-project scheduling problem with alternative activity chains and time flexibility. It extends the RCPSP-AC of \cite{tao2017scheduling}. New variables and constraints are introduced for the consideration of multiple lots, alternative activity chains and time flexibility. Non-renewable resources (e.g. money) are not explicitly considered, since they typically do not exist in production applications \citep{blazewicz2019handbook}. However, they can be included into the model in a staightforward way if necessary.

\nolinenumbers \noindent \textit{Minimize}
\begin{equation}
\sum_{t \in \mathcal{T}} t \cdot y_{n+1t}  																	\label{OF1}
\end{equation}\vspace{-0,5cm}
\\ \textit{subject to}%\vspace{-0,5cm}
\begin{equation}
s_{00} = 1,																							\label{NB2}
\end{equation}\vspace{-0,5cm}
\begin{equation}
\sum_{t \in \mathcal{T}} s_{it}= x_i 		\qquad  \forall \; i \in \mathcal{J},					\label{NB3}
\end{equation}\vspace{-0,5cm}
\begin{equation}
\sum_{t \in \mathcal{T}} y_{it} = x_i 		\qquad  \forall \; i \in \mathcal{J},					\label{NB4}
\end{equation}\vspace{-0,5cm}
\begin{equation}
\sum_{j \in \mathcal{J}} A_{ij} \cdot x_j = x_i 	\qquad \forall \; i \in \mathcal{J}, \; if \; p_i=0, \label{NB5}	
\end{equation}\vspace{-0,5cm}
\begin{equation}
A_{ij} \cdot x_i \leq x_j 								\qquad \forall \; i,j \in \mathcal{J}, \; if \; p_i=1,		\label{NB6}
\end{equation}\vspace{-0,5cm}
\begin{equation}
\sum_{i \in \mathcal{J}} A_{ij} \cdot x_j = x_j 		\qquad \forall \; j \in \mathcal{J}, \; if \; p_j=2, 		\label{NB7}
\end{equation}\vspace{-0,5cm}
\begin{equation}
\sum_{j \in \mathcal{J}} A_{ij} \cdot s_{jt} = y_{it} 	\qquad \forall \; i \in \mathcal{J}, \;  t \in \mathcal{T}, \; if \; p_i=0 \; and \; p_j \leq 1,		\label{NB8}
\end{equation}\vspace{-0,5cm}
\begin{equation}
y_{it} \cdot A_{ij} = s_{jt} \qquad \forall \; i,j \in \mathcal{J}, \;  t \in \mathcal{T}, \; if \; p_i=1 \; and \; p_j \leq 1,						\label{NB9}
\end{equation}\vspace{-0,5cm}
\begin{equation}	
\sum_{i \in \mathcal{J}} A_{ij} \cdot y_{it} = s_{jt} 	\qquad \forall \; j \in \mathcal{J}, \; t \in \mathcal{T},  \; if \; p_j=2, 					\label{NB10}
\end{equation}\vspace{-0,5cm}
\begin{equation}
\sum_{t \in \mathcal{T}} t \cdot y_{it}  \leq \sum_{t \in \mathcal{T}} t \cdot s_{n+1t} 	\qquad \forall \; i \in \mathcal{L},					\label{NB11}
\end{equation}\vspace{-0,3cm}
\begin{equation}
\sum_{\tau=1}^{t} (s_{i\tau}-y_{i\tau}) = w_{it} 		\qquad \forall \; i \in \mathcal{J}, t \in \mathcal{T},										\label{NB12}
\end{equation}\vspace{-0,5cm}
\begin{equation}
a_i \cdot x_i \leq \sum_{\tau \in \mathcal{T}} w_{i\tau}	\qquad \forall \; i \in \mathcal{J},														\label{NB13}
\end{equation}\vspace{-0,5cm}
\begin{equation}
b_i \cdot x_i \geq \sum_{\tau \in \mathcal{T}} w_{i\tau}	\qquad \forall \; i \in \mathcal{J},														\label{NB14}
\end{equation}\vspace{-0,5cm}
\begin{equation}
\sum_{t \in \mathcal{T}} t \cdot y_{it} \geq d_i 		\qquad \forall \; i \in \mathcal{L},														\label{NB15}
\end{equation}\vspace{-0,5cm}
\begin{equation}
\sum_{i \in \mathcal{J}} (w_{it} \cdot c_{ir}) \leq C_r	\qquad r \in \mathcal{R}, t \in \mathcal{T},												\label{NB16}
\end{equation}\vspace{-0,5cm}
\begin{equation}
x_i \in \{0,1\} 										\qquad \forall \; i \in \mathcal{J},														\label{NB17}
\end{equation}\vspace{-0,5cm}
\begin{equation}
s_{it}, w_{it}, y_{it} \in \{0,1\} \qquad \forall \; i \in \mathcal{J}, t \in \mathcal{T}.															\label{NB18}
\end{equation}
Objective function (1) minimizes the makespan of the whole production process. The new condition (2) starts the production process with the first activity at the first time slot. Restrictions (3)-(4) define that every activity has to be started and finished exactly once. With constraints (5)-(10), flexibility in terms of alternative activity chains and processing times is determined. If an activity is an OR node ($p_i=0$), only one of its successors in the project network must be selected with conditions (5). If an activity is an AND node ($p_i=1$), all successors have to be selected via restrictions (6). New constraints (7) with the flexibility type $p_j=2$ are necessary since they satisfy together with new constraints (10) the prohibition of selecting additional production nodes besides one activity chain per lot and the mandatory requirement of forbidden idle times between activity processing times within every lot. Conditions (8) and (9) are also new and guarantee the possibility of flexible processing times for AND/OR activities. It is assured that within one route, an activity $j$ has to be started at the finishing time of predecessor activity $i$ (no idle times are allowed) and that only activities can be selected which are in a precedence relationship. New constraints (11) ensure that all lots have to be finished before the whole project (production process) can be finished. With the new restrictions (12) it is guaranteed that every time slot $t$ which is used for the processing of one activity $i$ has to be between its decided start and end time. New conditions (13)-(14) ensure that the flexible processing time for every activity complies with its defined minimum and maximum allowed processing times. The new constraints (15) make sure that the production of one lot cannot be finished earlier than its delivery time and thus that tardiness is allowed but earliness is not. Conditions (16) represent the capacity restrictions for all renewable resources. Constraints (17)-(18) define all decision variables as Boolean ones.

\subsection{RCMPSP-ACTF MIP model: time balance maximization}\label{MIP_timebalance}
We now present an alternative objective that concerns the imbalance between time buffers, i.e. the duration (length) of different activities. As a result, the time balance of activities is maximized. Decision variables $B$ and $S$ decide on the largest and smallest time buffers within the whole project. Similar approaches can for example be found for the Vehicle Routing Problem with route balancing where the difference between route lengths is minimized \citep{matl2017workload}.

\vspace{-0.5cm}
\nolinenumbers \begin{equation}
\textit{Minimize} \qquad B-S															\label{OF19}
\end{equation}
Objective function (\ref{OF19}) minimizes the different lengths of activity durations. In addition to the already presented restrictions (\ref{NB2})-(\ref{NB18}) in Section \ref{MIP-makespan}, three further conditions are necessary:

\vspace{-0.5cm}
\nolinenumbers
\begin{equation}%\vspace{-0.1cm}
\sum_{t \in \mathcal{T}} w_{it}+(1-x_i) \cdot M - x_i \cdot a_i \geq S		\qquad \forall \; i \in \mathcal{J},				\label{NB20}
\end{equation}%\vspace{-0.8cm}
\begin{equation}
\sum_{t \in \mathcal{T}} w_{it}-x_i \cdot a_i \leq B						\qquad \forall \; i \in \mathcal{J},				\label{NB21}
\end{equation}%\vspace{-0.7cm}
\begin{equation}
B\geq 0, S \geq 0.															\label{NB22}
\end{equation}
With the newly introduced constraints (\ref{NB20})-(\ref{NB21}), minimum and maximum time buffers are connected to working times of activities. Conditions (\ref{NB22}) restrict decision variables to be of non-negative values.

\subsection{RCMPSP-ACTF MIP model: resource balance maximization}\label{MIP_resourcebalance}
The third objective function aims at balanced resource utilization. In existing works on resource leveling problems which come closest to our problem formulation, total weighted sums of squared resource usages are considered for the minimization of varying resource utilization over time and weights are represented by unit costs of resources \citep{li2018effective,neumann1999resource}. In our work, costs are not considered, since we do not discriminate between different resources but consider them equally important. However, motivated by the fact that resource balancing may not be meaningful for all of the considered resources (e.g. small vehicles versus large warehouses), we allow to select a subset of resources $\mathcal{R^*} \subseteq \mathcal{R}$ which are considered in the objective function. We use the decision variable $u_r$ for the concerned balancing resources $r \in \mathcal{R^*}$ to denote the maximum concurrent usage of one resource $r$. The resource balance objective can now be formulated as follows:

\vspace{-0.5cm}
\nolinenumbers \begin{equation}%\vspace{-0.2cm}
\textit{Minimize} \qquad \text{max}_{r \in \mathcal{R^*}} \; (u_r / C_r)									\label{OF23}
\end{equation}
Objective function (\ref{OF23}) minimizes the different peak usage of concerned renewable resources $\mathcal{R^*}$. In addition to constraints (\ref{NB2})-(\ref{NB15}) and (\ref{NB17})-(\ref{NB18}), the following conditions are necessary:

\vspace{-0.5cm}
\nolinenumbers \begin{equation}
\sum_{i \in \mathcal{J}} (w_{it} \cdot c_{ir}) \leq u_r		\qquad \forall\; r \in \mathcal{R^*}, t \in \mathcal{T},								\label{NB24}
\end{equation}%\vspace{-0.8cm}
\begin{equation}
u_r \leq C_r												\qquad \forall \; r \in \mathcal{R^*},													\label{NB25}
\end{equation}%\vspace{-0.8cm}
\begin{equation}
\sum_{i \in \mathcal{J}} (w_{it} \cdot c_{ir}) \leq C_r		\qquad \forall \; r \in \mathcal{R} \setminus \mathcal{R^*}, t \in \mathcal{T},			\label{NB26}
\end{equation}%\vspace{-0.8cm}
\begin{equation}
u_r \geq 0	 												\qquad \forall \; r \in \mathcal{R^*}.													\label{NB27}
\end{equation}
With restrictions (24)-(26), conditions (16) are replaced. With the new constraints, capacity restrictions are satisfied for all renewable resources. Constraints (27) guarantee non-negative values for the new decision variable.

\section{Constraint programming solution approach}\label{sec:CP}
Motivated by the recent success of CP based exact methods \citep{vilim2015failure}, we now propose CP models for the RCPSP-AC, its multi-project version and the RCMPSP-ACTF which can be solved by the CP Optimizer of IBM ILOG CPLEX. We first describe the main building blocks of the CP Optimizer and our developments in order to fit this modeling framework in Section \ref{sec:CP-notations}. Thereafter, in Sections \ref{CP-RCPSP-AC} - \ref{CP-resourceBalance}, the developed CP models are presented.

\subsection{Constraint Programming: Modeling developments and notations} \label{sec:CP-notations}
Besides the possibility of implementing MIP models, IBM ILOG CPLEX also provides the constraint programming framework CP Optimizer. \cite{laborie2018ibm} described its main ingredients and illustrated its performance on scheduling and other real-world problems. In the following, we describe the CP Optimizer functions and expressions that are necessary to develop our CP models. For all standard functions and expressions, we refer to Appendix \ref{AppendixA}.

\textit{Decision expressions:} With the decision variable \textit{interval($w_j$)}, an interval of time $w$ (a range or duration) is expressed for every activity $j$. Intervals are flexible in two ways: First, intervals can be of variable length (=time flexibility). Second, activities can be left unperformed, which is necessary as there are alternative routes and therefore some activities which have to be skipped (=alternative activity chains). The project horizon $T$ is now used as a constant, which limits the maximum length of the interval decision variable in contrary to the MIP models in Section \ref{sec:MIP} where $t \in \{1,...,T\}$ was a necessary index for decision variables. Alternative activity chains are considered by the statement \texttt{optional}. As a result, \texttt{dvar} \textit{interval($w_j$)} \texttt{optional in 0..T} is introduced.

\begin{figure}
	\centering\includegraphics[width=1.00\textwidth]{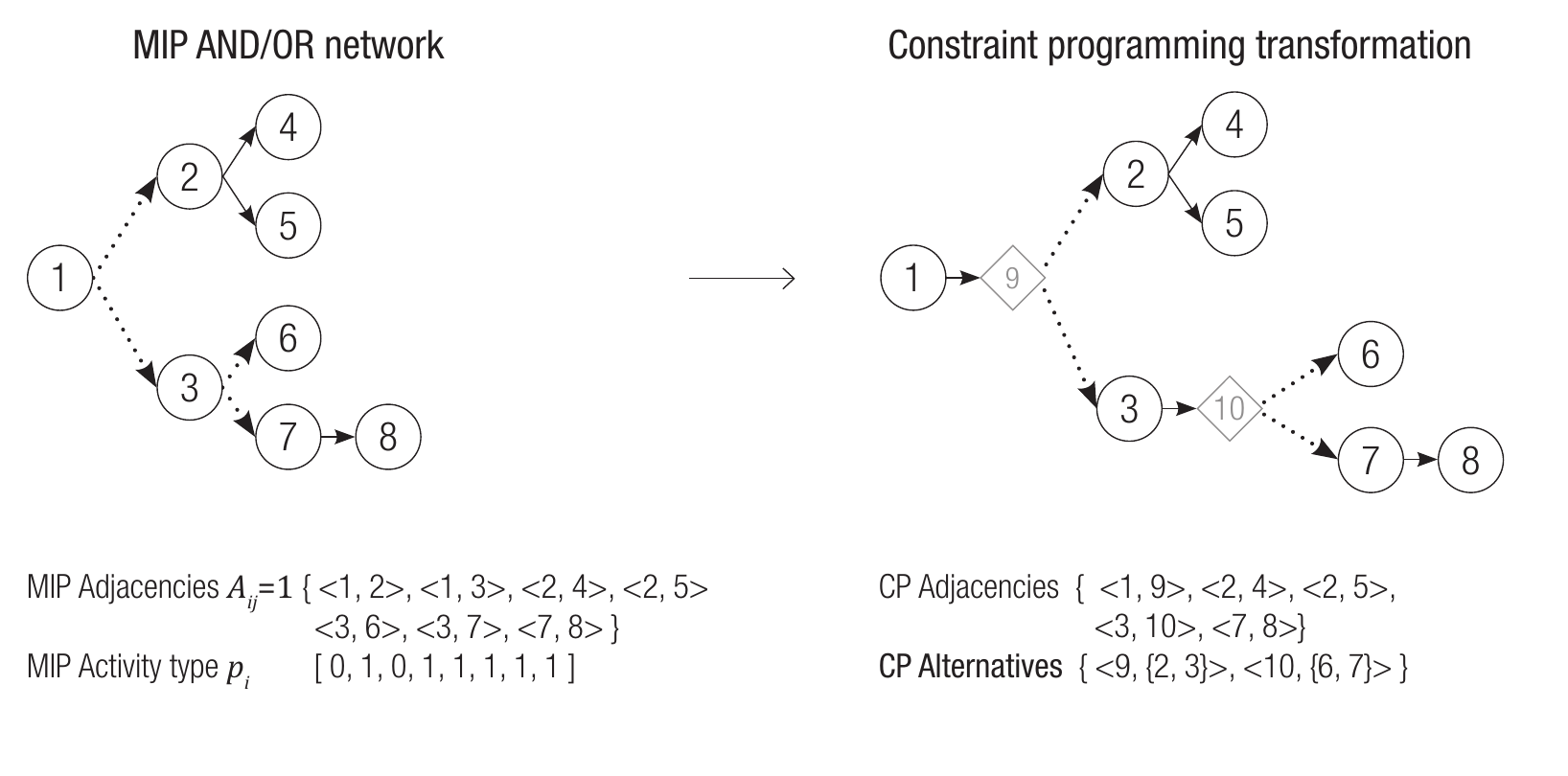}
	\caption{Introduction of meta nodes for the CP \texttt{alternative()} expression.}
	\label{FIG_MIP-CP-Transformation_alternative} %\vspace{length}
\end{figure}

\textit{Alternative expressions:} With the \texttt{alternative}$(w_i,\{w_j,w_k\})$ expression, the possibility of choosing between different alternative successor activities $j$ and $k$ is modeled. If node $i$ is present, exactly \textit{one} out of multiple alternative successor nodes $\{j,k\}$ can be selected and the selected alternative successor node \textit{must} start and end together with node $i$. However, in typical scheduling applications, and also in the RCMPSP-ACTF, a node $i$ cannot always start and end together with a successor node $\{j,k\}$ due to specified precedence relations and time restrictions. Thus, for every OR node $i$ which implies a decision on a successor activity, one dummy alternative node $meta_a$ has to be introduced. In Figure \ref{FIG_MIP-CP-Transformation_alternative}, the introduction of the necessary $meta_a$ nodes inspired by the work of \cite{tao2017scheduling} 
is illustrated. On the left, an example AND/OR network with the necessary MIP adjacencies and flexibility types is depicted, including the OR relations \{$<1,2>,<1,3>$\} and \{$<3,6>,<3,7>$\} in the MIP adjacencies for nested OR nodes 1 and 3. For the CP transformation on the right, the \texttt{alternative} expression has to be introduced: The dummy meta node 9 is necessary since it can start and end together with the selected successor 2 or 3 in contrary to node 1, which has to be finished before the start of node 2 or 3. Exactly the same logic is applied for the nested OR node 3 with meta node 10. In contrary to the MIP adjacencies, the OR relations for the nested OR nodes 1 and 3 cannot be inserted in the CP adjacencies: Instead, the relations of OR nodes with the $meta_a$ nodes \{$<1,9>,<3,10>$\} have to be inserted in the CP adjacencies and the relations of the $meta_a$ nodes with real nodes (9, \{2, 3\}) and (10, \{6, 7\}) have to be transferred to the CP alternatives.

\begin{figure}
	\centering\includegraphics[width=1.00\textwidth]{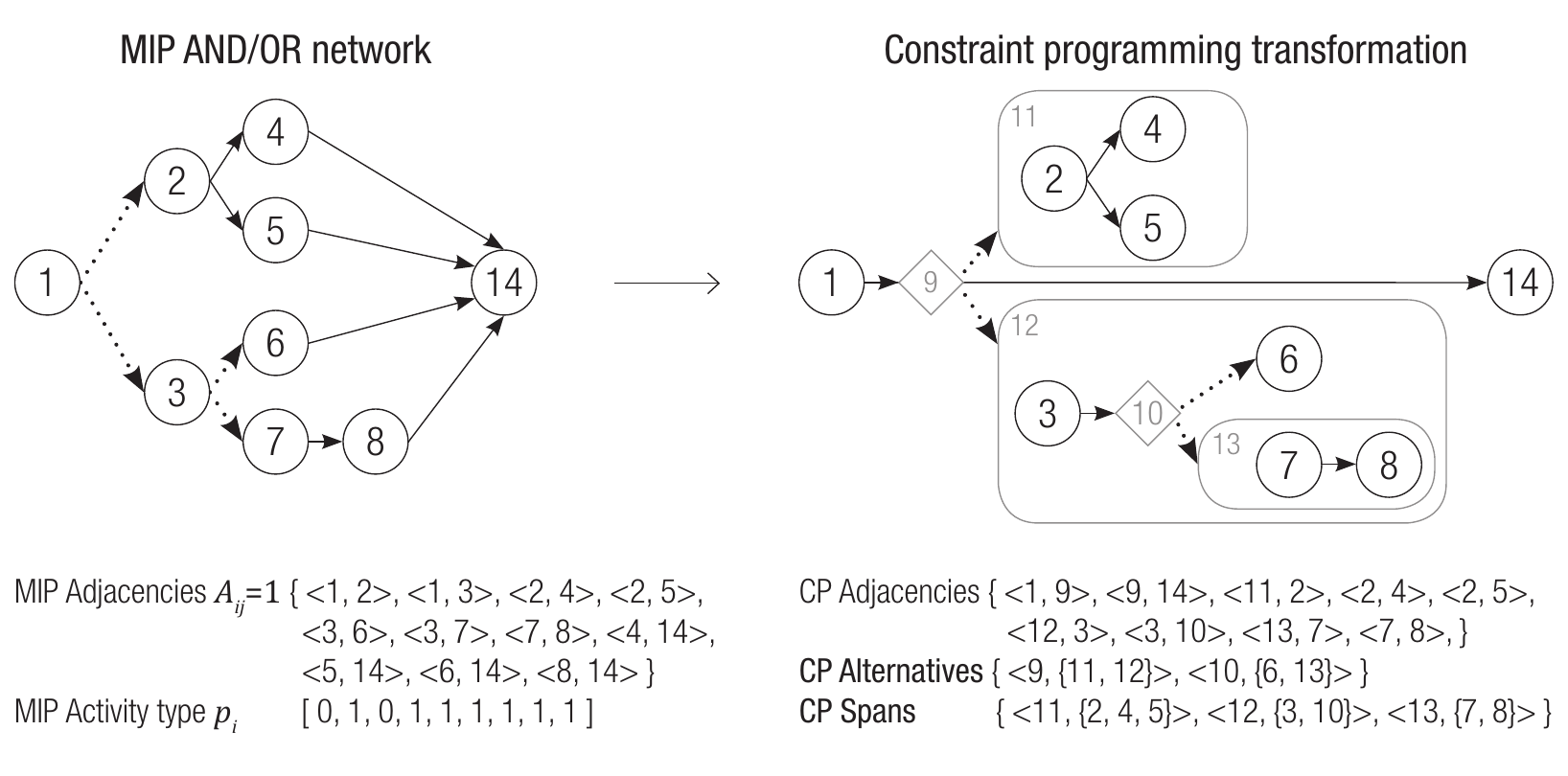}
	\caption{Introduction of meta nodes for the CP \texttt{alternative()} and \texttt{span()} expression.}
	\label{FIG_MIP-CP-Transformation_span}
\end{figure}

\textit{Span expressions:} With the \texttt{span}$(w_i,\{w_j,w_k\})$ expression, all activities $j$ and $k$ are included (spanned), if an activity $i$ is selected. In an AON network with nested OR nodes and one common end node, which is only allowed to be started after \textit{all} predecessors are scheduled, this expression is necessary. The before described \texttt{alternative()} expression is not applicable, since it allows only one predecessor and one successor node to start and end together but not all activities can be included. For every node that implies a decision on a successor activity chain, i.e. a relation which includes more than one node, one dummy span activity of the type $meta_s$ has to be introduced. In Figure \ref{FIG_MIP-CP-Transformation_span}, an example MIP network with one common end node 14 is shown on the left and the corresponding CP transformation is presented on the right. The OR nodes 1 and 3 are linked to $meta_a$ alternative nodes 9 and 10 as already described for Figure \ref{FIG_MIP-CP-Transformation_alternative}. However, the new common project end node 14 can only start if all predecessor nodes are finished. Hence, for every OR node, which includes more than one successor node, one additional $meta_s$ node is necessary: OR node 1 has two successor relations: The first relation has three nodes \{2, 4, 5\} and thus, gets $meta_s$ node 11. The second with four nodes \{3, 6, 7, 8\} is linked to $meta_s$ node 12. The nested OR node 3 also has two successor relations. The first relation includes only one successor node \{6\} and therefore, it does not require additional meta nodes. The second successor relation \{7, 8\} contains more than one node and thus, the $meta_s$ node 13 is introduced. The span nodes are linked to the first node of a span successor relation in the CP adjacencies, e.g. span node 11 is linked to node 2. All further span and alternative relations have to be transferred to CP alternatives and spans instead of an inclusion into the adjacencies.

\textit{Resource function:} The \texttt{cumulFunction} $q_r=\sum_{j \in \mathcal{J}:c_{jr}\geq0}\texttt{pulse}(w_j,c_{jr})$ expresses the cumulative usage $q$ of a renewable resource $r \in \mathcal{R}$ over time for every activity $j$. It only counts the demand $c_{jr}$ of one activity if the corresponding optional \textit{interval}$(w_j)$ is used and only allows an accumulation if resource capacities $C_r$ are exhausted. The temporal change of the resource usage in dependence of the demand $c_{jr}$ is initiated by the integrated function \texttt{pulse}. As a result, the consideration of alternative activity chains and time flexibility is taken care of in this expression. The following Table \ref{CPnotations} gives all notations, which are used for the development of our new CP models.
\begin{table}[!h]
	\caption{CP model notations.}
	\vspace{1.0ex}
	\label{CPnotations}
	\begin{tabularx}{\textwidth}{lX}
		\hline
		\multicolumn{2}{l}{\textbf{Indices}}	
		\\
		$i,j$ 		& 	Activities
		\\			
		$r$			&	Resource	
		\\
		\multicolumn{2}{l}{\textbf{Sets}}	
		\\
		$\mathcal{A}$	&	Set of adjacencies between activities $(i,j)$
		\\
		$\mathcal{J}$	&	Set of activities $i,j \in \{0,1,..,n+1\}$
		\\
		$\mathcal{L}$	&	Set of delivery activities $\mathcal{L} \subseteq \mathcal{J}$
		\\		
		$\mathcal{M}$	&	Set of meta activities $\mathcal{M} \subseteq \mathcal{J}$
		\\			
		$\mathcal{R}$	&	Set of renewable resources $r \in \{0,1,..,R\}$
		\\
		$\mathcal{R^*}$	&	Set of renewable resources $\mathcal{R^*} \subseteq \mathcal{R}$ considered for resource balance
		\\	
		$\mathcal{N}$	&	Set of non-renewable resources $r \in \{0,1,..,N\}$
		\\
		\multicolumn{2}{l}{\textbf{Parameters}}	
		\\	
		$a_j$		&	Minimum allowed processing time (duration) of activity $j$
		\\
		$b_j$		&	Maximum allowed processing time (duration) of activity $j$
		\\			
		$D_j$		&	Fixed processing time (duration) of activity $j$
		\\
		$d_j$		&	Delivery time (due date) of activity $j \in \mathcal{L}$
		\\
		$T$			&	Time horizon
		\\			
		$C_r$		&	Capacity of resource $r$
		\\
		$c_{jr}$	&	Demand of activity $j$ of resource $r$
		\\
		$S_j$		&	Alternative start activities of alternative activity $j$			
		\\
		$E_j$		&	Alternative end activities of alternative activity $j$
		\\
		$G_j$		&	Possible span activities (=selection relation) of alternative activity $j$
		\\			
		\multicolumn{2}{l}{\textbf{Decision variables}}
		\\
		$q_r$			&	Cumulative resource usage of renewable resources $r \in \mathcal{R}$ over time
		\\
		$u_r$		&	Maximum resource utilization (peak usage) of resource $r \in \mathcal{R^*}$
		\\
		$w_j$		&	Optional interval decision variable: selection of activity $j$ for the production process and assignment of start, duration, and end time (interval) for every selected activity $j$
		\\
		\hline
	\end{tabularx}
\end{table}

\subsection{RCPSP-AC and RCMPSP-AC CP Model}\label{CP-RCPSP-AC}
We now propose a CP formulation for the RCPSP with alternative activity chains (RCPSP-AC) of \cite{tao2017scheduling} and its multi-project version, the RCMPSP-AC. To the best of our knowledge, this is the first time that this problem is modeled and solved with CP. It allows us to solve all RCPSP-AC benchmark instances to optimality. For the multi-project version, we use the well-known SP approach as already introduced for the MIP formulation of the RCMPSP-ACTF in Section \ref{sec:MIP}. This means that we do not change the model of the RCPSP-AC but the precedence relations, i.e. the project structure itself, to obtain the computationally more expensive RCMPSP-AC (for a detailed explanation we refer to Section \ref{sec:InstanceGeneration}).

\nolinenumbers \noindent \textit{Minimize} %\vspace{-0,5cm}
\begin{equation}
\texttt{{endOf}\big($w_{n+1}$\big)}														\label{OF28}
\end{equation}\vspace{-1cm}
\\ \textit{subject to}%\vspace{-0,5cm}
\begin{equation}
\texttt{{startOf}\big($w_0$\big)} = 1,														\label{NB29}
\end{equation}\vspace{-0,8cm}
\begin{equation}
\texttt{{presenceOf}\big($w_0$\big)} = 1, 													\label{NB30}
\end{equation}\vspace{-0,8cm}
\begin{equation}
\texttt{{lengthOf}\big($w_i$\big)} = D_i \qquad \forall\; i \in \mathcal{J}, 			\label{NB31}
\end{equation}\vspace{-0,8cm}
\begin{equation}
\texttt{{alternative}\big($w_i,\{w_a \in S_i$\}\big)}  \qquad \forall\; i \in \mathcal{M},	\label{NB32}
\end{equation}\vspace{-0,8cm}
\begin{equation}
\texttt{{span}\big($w_i,\{w_s \in G_i$\}\big)}  \qquad \forall\; i \in \mathcal{M},		\label{NB33}
\end{equation}\vspace{-0,8cm}
\begin{equation}
\texttt{{endBeforeStart}\big($w_i,w_j$\big)}  \qquad \forall\; i,j \in \mathcal{A},	\label{NB34}
\end{equation}\vspace{-0,8cm}
\begin{equation}
\texttt{{presenceOf}\big($w_i$\big)} = \texttt{{presenceOf}\big($w_j$\big)}, \qquad \forall\; i,j \in \mathcal{A},											\label{NB35}
\end{equation}\vspace{-0,8cm}
\begin{equation}
q_r \leq C_r	\qquad \forall\;  r \in \mathcal{R},										\label{NB36}
\end{equation}\vspace{-0,8cm}
\begin{equation}
\sum_{j \in \mathcal{J}\setminus\mathcal{M}} \texttt{{presenceOf}\big($w_j$\big)} \cdot c_{jr} \leq C_r	\qquad \forall \; r \in \mathcal{N}.												  \label{NB37}
\end{equation}
With objective function (\ref{OF28}), the makespan of the project is minimized. Restrictions (\ref{NB29})-(\ref{NB30}) determine the start of the project with the first activity at the first time slot of the project. Conditions (\ref{NB31}) define that fixed processing times of all activities have to be satisfied. Constraints (\ref{NB32})-(\ref{NB33}) guarantee activity selection flexibility for nested AND/OR relations. Restrictions (\ref{NB34})-(\ref{NB35}) ensure that the precedence relations between different activities are met and that idle times between the scheduling of different activities are allowed. With constraints (\ref{NB36})-(\ref{NB37}), capacity limits for renewable and non-renewable resources are satisfied.

\subsection{RCMPSP-ACTF CP Model: makespan minimization}\label{CP-makespan}
In this section, we present the CP model for the RCMPSP with alternative activity chains and time flexibility. As already stated for the MIP model of this new problem, flexible processing times are considered and idle times are not allowed within the production processes of single lots. However, the parallel production of multiple lots and thus, the concurrent scheduling of multiple activities of different lots is allowed. Non-renewable resources are not considered. Nevertheless, they can be included easily since they can be modeled in the same way as in the case of the RCPSP-AC (see constraints \ref{NB37}).

\nolinenumbers \noindent \textit{Minimize} %\vspace{-0,5cm}
\begin{equation}
\texttt{{endOf}\big($w_{n+1}$\big)}														\label{OF38}
\end{equation}\vspace{-1cm}
\\ \textit{subject to}%\vspace{-0,5cm}
\begin{equation}
\texttt{{startOf}\big($w_0$\big)} = 1,														\label{NB39}
\end{equation}\vspace{-0,8cm}
\begin{equation}
\texttt{{presenceOf}\big($w_0$\big)} = 1,													\label{NB40}
\end{equation}\vspace{-0,8cm}
\begin{equation}
\texttt{{presenceOf}\big($w_{n+1}$\big)} = 1,												\label{NB41}
\end{equation}\vspace{-0,8cm}
\begin{equation}
\texttt{{presenceOf}\big($w_i$\big)} = 1	\qquad \forall \; i \in \mathcal{L},			\label{NB42}
\end{equation}\vspace{-0,8cm}
\begin{equation}
\texttt{{lengthOf}\big($w_i$\big)} \geq a_i	\qquad \forall \; i \in \mathcal{J},			\label{NB43}
\end{equation}\vspace{-0,8cm}
\begin{equation}
\texttt{{lengthOf}\big($w_i$\big)} \leq b_i	\qquad \forall \; i \in \mathcal{J},			\label{NB44}
\end{equation}\vspace{-0,8cm}
\begin{equation}
\texttt{{endOf}\big($w_i$\big)} \geq d_i	\qquad \forall \; i \in \mathcal{L},			\label{NB45}
\end{equation}\vspace{-0,8cm}
\begin{equation}
\texttt{{endOf}\big($w_i$\big)} \leq T	\qquad \forall \; i \in \mathcal{L},				\label{NB46}
\end{equation}\vspace{-0,8cm}
\begin{equation}
\texttt{{alternative}\big($w_i,\{w_a \in S_i$\}\big)}  \qquad \forall\; i \in \mathcal{M},	\label{NB47}
\end{equation}\vspace{-0,8cm}
\begin{equation}
\texttt{{endAtStart}\big($w_i,w_a$\big)}	\qquad \forall \; i \in \mathcal{M}, a \in E_i,												\label{NB48}
\end{equation}\vspace{-0,8cm}
\begin{equation}
\texttt{{endBeforeStart}\big($w_i,w_{n+1}$\big)}	\qquad \forall \; i,j \in \mathcal{L},	\label{NB49}
\end{equation}\vspace{-0,8cm}
\begin{equation}
\texttt{{endAtStart}\big($w_i,w_j$\big)}	\qquad \forall \; i,j \in \mathcal{A},			\label{NB50}
\end{equation}\vspace{-0,8cm}
\begin{equation}
\texttt{{presenceOf}\big($w_i$\big)} = \texttt{{presenceOf}\big($w_j$\big)}	\qquad \forall \; i,j \in \mathcal{A},																				\label{NB51}
\end{equation}\vspace{-0,8cm}
\begin{equation}
q_r \leq C_r	\qquad \forall\;  r \in \mathcal{R}.										\label{NB52}
\end{equation}
With objective (\ref{OF38}), the makespan of the whole production process is minimized. Restrictions (\ref{NB39})-(\ref{NB42}) define the start and end of the project and guarantee the production of all lots. Conditions (\ref{NB43})-(\ref{NB44}) enable flexible processing times. The processing time of every selected activity has to comply with minimum and maximum allowed durations. Constraints (\ref{NB45}) allow tardiness for every lot. The overall production planning time is set to the predefined horizon $T$ in restrictions (\ref{NB46}). Conditions (\ref{NB47})-(\ref{NB48}) specify alternative production routes for every lot. One out of multiple existing meta alternative start and end activities has to be chosen and idle times between meta activities within one lot are forbidden. Restrictions (\ref{NB49})-(\ref{NB50}) allow idle times between the production activities of different lots and forbid idle times between production activities within one lot. Constraints (\ref{NB51}) guarantee that precedence relations are adhered to. With constraints (\ref{NB52}), capacity restrictions for renewable resources are satisfied.

\subsection{RCMPSP-ACTF CP Model: time balance maximization}\label{CP-timeBalance}
We now show how to model the objective of time balance maximization:

\vspace{-1cm}
\nolinenumbers \begin{equation}
\textit{Minimize } \big(\text{max}_{i \in \mathcal{J} \setminus \mathcal{M}} \texttt{lengthOf}(w_i)-a_i\big) - \big(\text{min}_{i \in \mathcal{J} \setminus \mathcal{M}} \texttt{lengthOf}(w_i)-a_i\big)		\label{OF53}\vspace{-0,3cm}
\end{equation}
With objective function (\ref{OF53}), the difference between time buffer lengths of all activities is minimized and thus, time balance is maximized. There is no need for additional decision variables or changed restrictions in contrary to the MIP model in Section \ref{MIP_timebalance}. Instead, the new objective function is employed together with the presented CP restrictions (\ref{NB39})-(\ref{NB52}).

\subsection{RCMPSP-ACTF CP Model: resource balance maximization}\label{CP-resourceBalance}
The objective of resource balance maximization is modeled as follows:

\vspace{-0,5cm}
\nolinenumbers \begin{equation}
\textit{Minimize} \qquad \text{max}_{r \in \mathcal{R^*}} \big(u_r / C_r \big)  \label{OF54} \vspace{-0,1cm}
\end{equation}
Objective function (\ref{OF54}) minimizes the difference between the resource usage of single resources and thus, maximizes load balancing between all resources. Besides the already introduced constraints (\ref{NB39})-(\ref{NB51}), it requires the following additional ones:

\vspace{-0,5cm}
\nolinenumbers \begin{equation}
q_r \leq u_r \qquad \forall \, r \in \mathcal{R^*},  \label{NB55} %\vspace{-0,1cm}
\end{equation}\vspace{-0,8cm}
\begin{equation}
u_r \leq C_r \qquad \forall \, r \in \mathcal{R^*},  \label{NB56} %\vspace{-0,1cm}
\end{equation}\vspace{-0,8cm}
\begin{equation}
q_r \leq C_r \qquad \forall \, r \in \mathcal{R} \setminus \mathcal{R^*}.  \label{NB57} %\vspace{-0,1cm}
\end{equation}
The new restrictions (\ref{NB55})-(\ref{NB57}) replace constraints (\ref{NB52}). They restrict the peak usage of all concerned resources to the defined capacity limits.

\section{Computational experiments}\label{sec:ComputationalExperiments}
The MIP and CP models are implemented in OPL and the CPLEX 12.9.0 MIP solver and CP Optimizer are used to solve them. All experiments are carried out on a virtual machine Intel(R) Xeon(R) CPU E5-2660 v4, 2.00GHz with 28 logical processors, Microsoft Windows 10 Education. Since \cite{tao2017scheduling} introduce a limit of 3.600 seconds for their runs and we derive results for benchmark instances generated as described in their work, we use the same limit for our optimization runs. We first describe the test design in Section \ref{sec:InstanceGeneration} and then present and discuss the obtained results for the benchmark sets and the industry case study in Sections \ref{sec:RCPSP-AC_Results}-\ref{sec:CaseStudy_Results}.

\subsection{Instance generation}\label{sec:InstanceGeneration}
\begin{figure}
	\centering\includegraphics[width=0.40\textwidth]{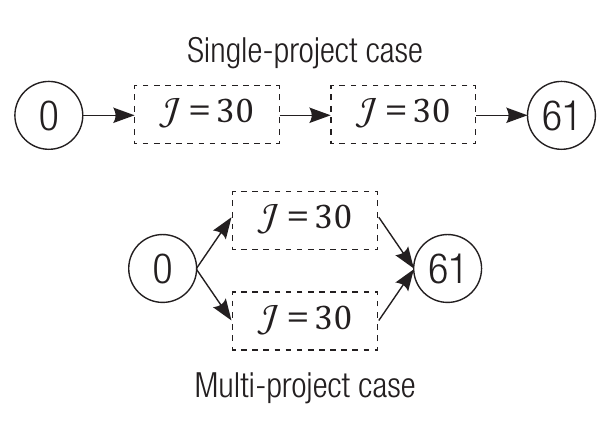}
	\caption{Benchmark instance generation for the RCPSP-AC and the RCMPSP-AC.}
	\label{FIG_InstanceGeneration}
\end{figure}

For the RCPSP-AC, we base our single-project instances on the information given in \cite{tao2017scheduling}, since their employed instances are not available and they describe the necessity of an officially available benchmark set in their work. However, they fully present one instance in their work, consisting of a project structure with $\mathcal{J}=30$ nodes in total and including five nested OR nodes. Following \cite{tao2017scheduling}, to obtain five instance groups with 15 instances each, this project structure is multiplied by 2, 3, 4 and 5, resulting in instance groups with 30, 60, 90, 120, and 150 nodes besides two additional dummy nodes for each instance (start and end node). As described in their work, processing times and demands for all resources are randomly generated. In order to obtain multi-project instances, the single-project instances of \cite{tao2017scheduling} are extended by arranging several project structures in parallel instead of in sequence (see Figure \ref{FIG_InstanceGeneration} for an example). We note that the MIP and CP models for the RCPSP-AC do not have to be adapted for this multi-project case (=RCMPSP-AC), since the precedence relations, i.e. the project structure itself, is changed. This is in line with the way multi-project instances have for example been presented by \cite{lova2001analysis}.

The instances for the RCMPSP-ACTF are inspired by the project network data obtained from cooperations with the manufacturing industries described in Section \ref{sec:Introduction}. This network consists of activities such as production (0), cooling (1), processing (2), relocation (3), storaging (4), vehicle relocation (5) and delivery (6). Activities (0) and (3) have a fixed duration, all other activities have variable processing times with minimum and maximum allowed durations. Each lot starts at (0) and ends at (6) via one alternative route. The customer orders are connected to related alternative routes, which may involve different activities at various locations. We use three different lot sizes $|\mathcal{L}| \in \{10,50,100\}$. All lots are sampled from existing customer orders. Processing times, resource capacities with related resource factors \{0.25, 0.50, 0.75, 1.00\} and the resource strength are defined based on \cite{kolisch1995characterization}. Resource demands are equal to one for the real-world situation and only one out of all existing resources is required by each activity. We use an additional, different demand pattern where the demands and the amount of required resources are defined randomly to be able to validate the real-world situation. As a result, we have two demand patterns real-world (\textit{rw}) and random (\textit{rand}). With the 3 lot sizes, the 2 demand patterns and the 4 resource strengths, we have 24 instance groups. For each group we generate 5 instances with varying random seed, which results in a total of 120 instances. The whole data generation procedure is presented in detail in Appendix \ref{AppendixB}. 

For all instances, we note that we limit the large constant $M$ (``Big M'') in the MIP models to the maximum allowed project duration $T$ in order to support better relaxations and integer solvability and a less required computation time \citep{camm1990cutting}. 

\subsection{RCPSP-AC and RCMPSP-AC: Optimization results}\label{sec:RCPSP-AC_Results}
In Table \ref{FIG_ResultsRCPSP-AC}, we provide the results for the CP and the MIP models on the generated single-project instances for the RCPSP-AC of \cite{tao2017scheduling} and the multi-project instances for the RCMPSP-AC. Each entry is an average value across 15 instances per data set. The first column gives the instance size. In columns 2-6, CP solutions are presented: The second column provides the best bound and the best solution is found in the third column. In the fourth column, the runtime is shown in seconds, followed by the number of instances for which an optimal solution was found and the number of instances for which a feasible solution was found in columns 5 and 6. Bold letters indicate the optimal solution. Columns 7-11, where MIP solutions are shown, follow the same logic as columns 2-6. We note that in order to validate our optimization results for the RCPSP-AC, we implemented the MIP model presented by \cite{tao2017scheduling} and tested it with the benchmark instance $\mathcal{J}=30$ which they presented in their paper. Since we had to add and change several constraints of their MIP model to obtain the same results as presented in their paper for this instance, we provide the modified MIP model in Appendix \ref{AppendixC}. The detailed results for every examined benchmark instance are given in Appendix \ref{AppendixD}.
\begin{table}[!h]
	\caption{Results for the RCPSP-AC and the RCMPSP-AC.}
	\vspace{1.0ex}
	\label{FIG_ResultsRCPSP-AC}		
	\centering
	\includegraphics[width=1.0\textwidth]{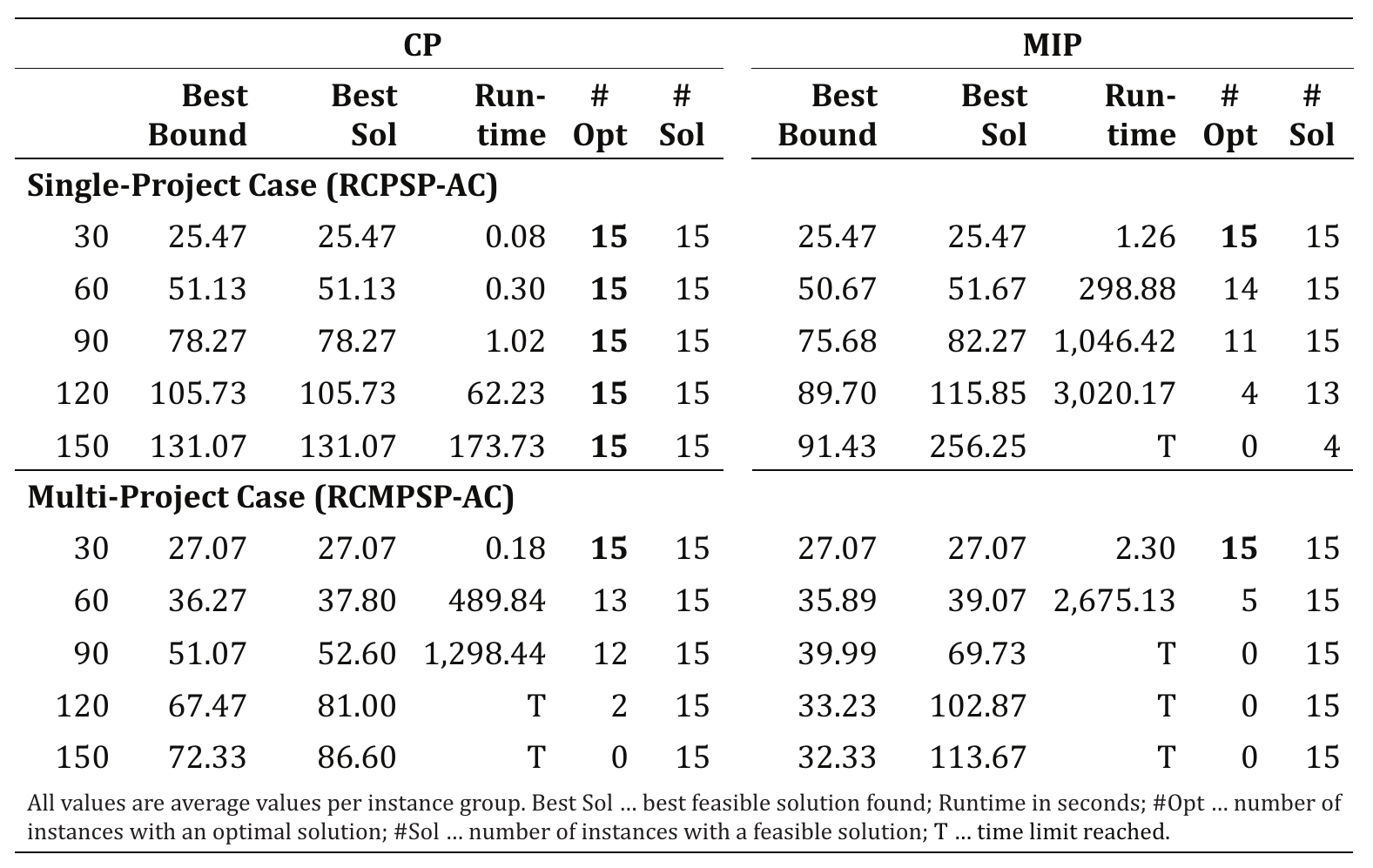}
\end{table}

For the single-project case presented in Table \ref{FIG_ResultsRCPSP-AC}, all benchmarks are solved to optimality by using the CP model but not when using the MIP model. The CP approach provides optimal solutions to dimensions of increasing difficulty with very little effort while in case of the MIP model the solver struggles to solve instances with more than 90 nodes within the allotted runtime. The results also show that the CP Optimizer solves instances of size 150 in less time on average than the MIP solver for dimension 60.

Given these results, we have generated additional new multi-project instances which are computationally more challenging. For this new RCMPSP-AC case, the CP and the MIP solver struggle to solve problems of increasing size to optimality. Although they both get feasible solutions for all problem instances, the CP Optimizer finds better bounds and better solutions on average than the MIP solver. Interestingly, for the MIP solver, finding feasible solutions for multi-project instances appears to be easier than for single-project instances. Nevertheless, proving their optimality is considerably more difficult.

As explained in Section \ref{sec:InstanceGeneration}, benchmarks are generated equal to the description of \cite{tao2017scheduling} with the only difference of a parallel project structure for the multi-project case. This means that the same resource capacities are used for the multi-project case where a lot more activities have to be scheduled in parallel than in the single-project case of \cite{tao2017scheduling}. Thus, the resource strength, which is an indicator of instance hardness \citep{kolisch1995characterization} is much higher.

\subsection{RCMPSP-ACTF: Optimization results}\label{sec:RCMPSP-ACTF_Results}
\begin{figure}%[!h]
	\centering
	\includegraphics[width=1.0\textwidth]{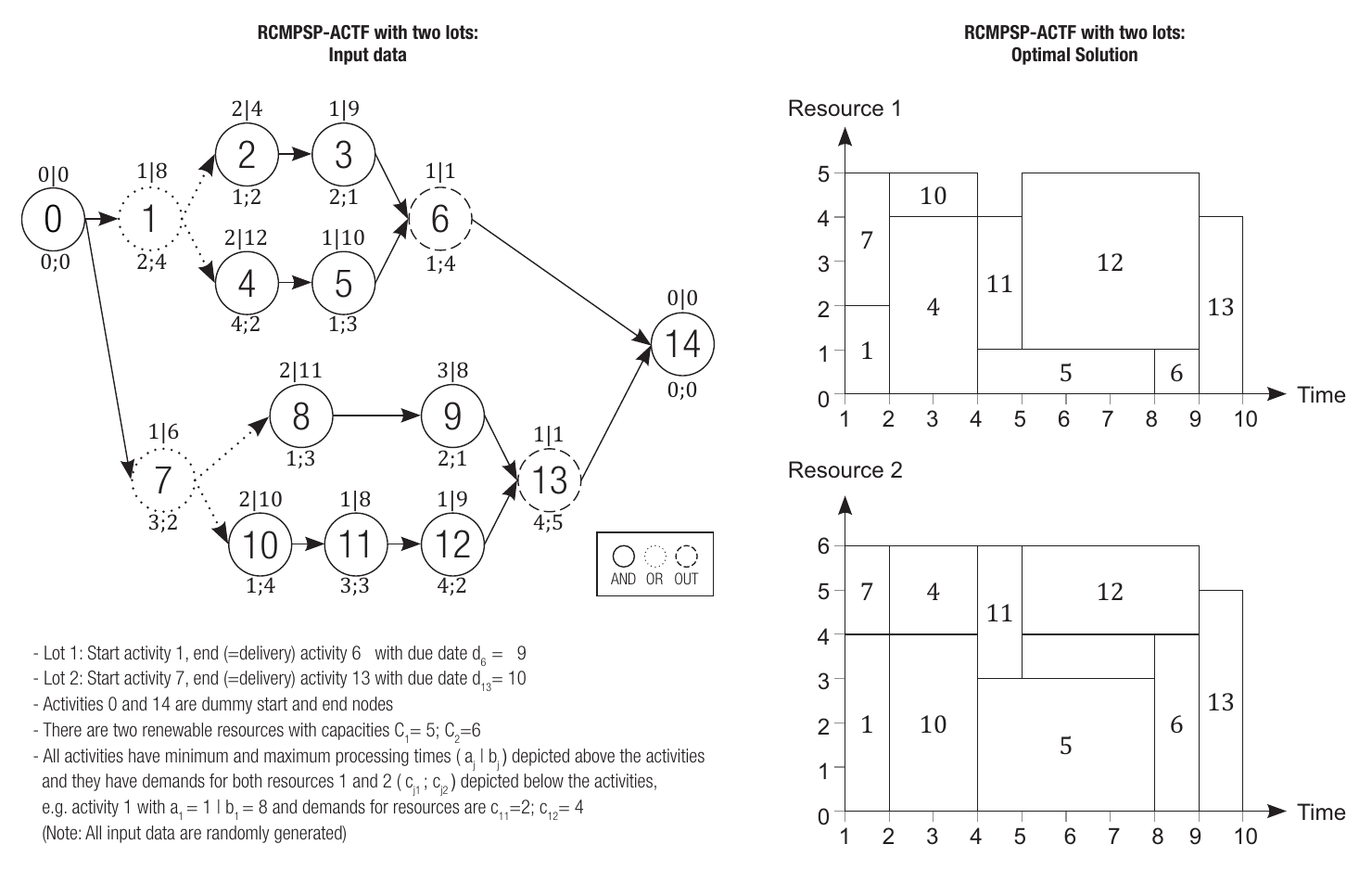}
	\caption{Illustrative optimization example for an RCMPSP-ACTF.}
	\label{FIG_ResultsGantt}
\end{figure}

In Table \ref{FIG_ResultsRCMPSP-ACTF}, we present the optimization results for the benchmark data generated for the RCMPSP-ACTF. They follow the same logic as the results for the RCPSP-AC and RCMPSP-AC shown in Table \ref{FIG_ResultsRCPSP-AC} with the only difference that in the second column now the respective demand patterns (real-world and random) are given in addition. The detailed results for every examined instance can be found in Appendix \ref{AppendixE}. For illustration purposes, we provide the optimal solution for the exemplarily toy instance described in Figure \ref{FIG_Illustrative-AON_RCMPSP-ACTF}, Section \ref{MIP-Notations} in the form of a Gantt chart (see Figure \ref{FIG_ResultsGantt}).

For the objectives makespan and time balance presented in Table \ref{FIG_ResultsRCMPSP-ACTF}, the CP approach solves all instances to optimality while the MIP solver only solves all instances of lot size 10 and some of size 50 and 100. In all cases, it is much slower (more than two orders of magnitude) than the CP solver. Another difference between the two solution approaches is the runtime, as it could already be detected for the optimization results in Table \ref{FIG_ResultsRCPSP-AC}. On average, the MIP solver needs considerably more runtime or even reaches the time limit in contrary to the CP Optimizer.
\begin{table}[!h]
	\caption{Results for the RCMPSP-ACTF.}
	\vspace{1.0ex}
	\label{FIG_ResultsRCMPSP-ACTF}
	\centering
	\includegraphics[width=1.0\textwidth]{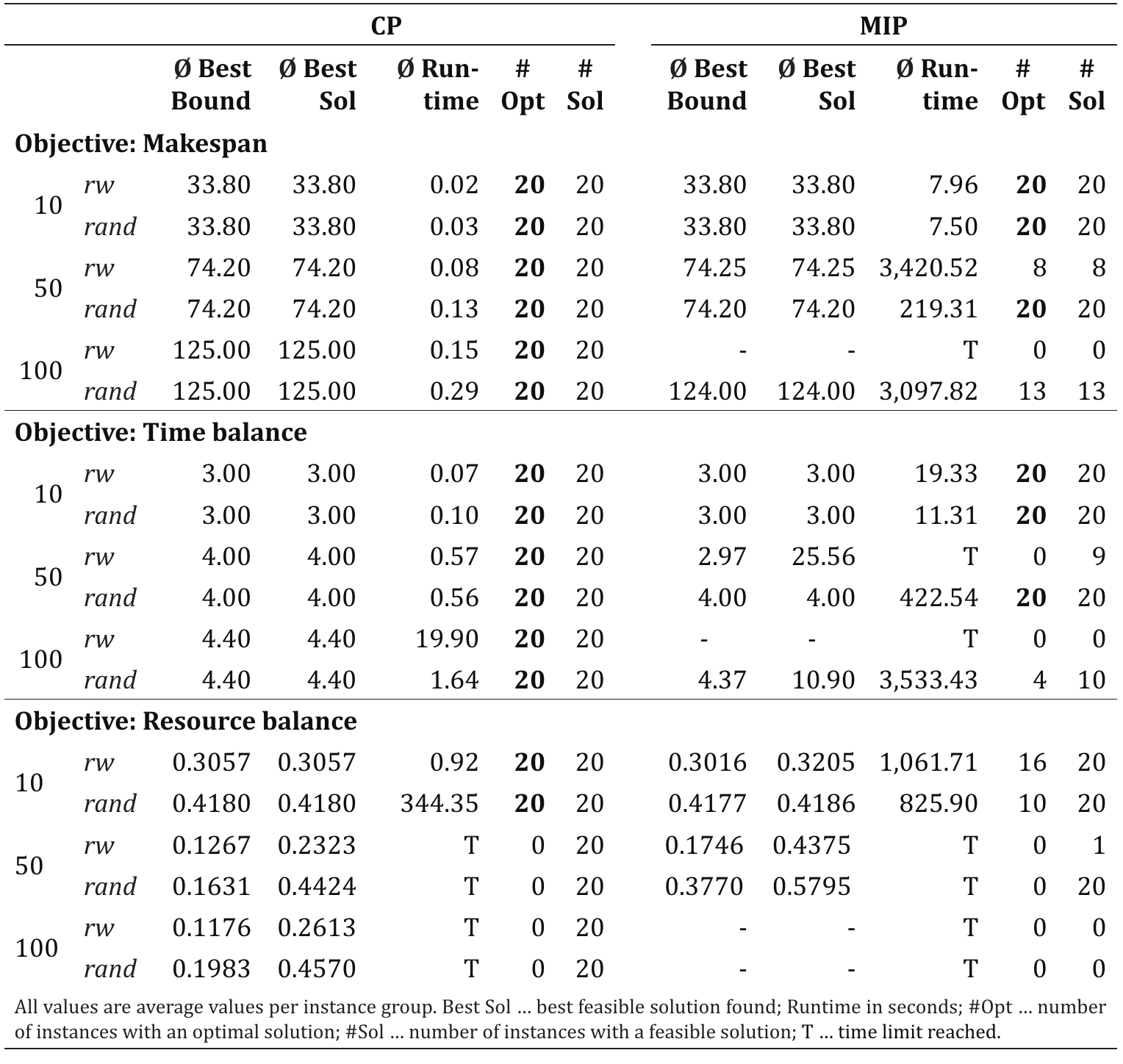}
\end{table} 

Concerning the third objective of resource balance, the CP Optimizer overall, again finds better solutions than the MIP solver. However, it struggles a lot more to prove optimality within the allotted runtime than with the other objectives and does not find any optimal solution for instances of size 50 and 100. For the instances of size 50 and the demand pattern random, the MIP solver finds better bounds than the CP Optimizer does. Nevertheless, in contrary to the MIP solver, the CP Optimizer provides integer solutions for all instances.

Overall, two things particularly stand out in Table \ref{FIG_ResultsRCMPSP-ACTF}. First, the CP Optimizer solves larger instances to optimality than in the case of the RCPSP-AC in Table \ref{FIG_ResultsRCPSP-AC}. We assume that the reason for this difference lies in the project structure of real-world manufacturing situations. For the generated benchmarks in Table \ref{FIG_ResultsRCPSP-AC}, nested AND/OR relations and a parallel structure are considered as described in \cite{tao2017scheduling}. For the industry situations which motivate the RCMPSP-ACTF, a production schedule has to be generated for a serial project structure with one OR relation per sub-project. Second, the CP Optimizer appears to be a lot more competitive than the MIP solver concerning flexible RCPSP. We assume that one major reason is the different modeling approach concerning decision variables, resulting in a strongly divergent amount of constraints that have to be considered. For example taking a benchmark instance with 100 lots and 1,658 activities with the objective makespan minimization, the MIP solver has to take 490,765 constraints and 679,288 variables into account but the CP solver only considers 6,647 constraints and 1,658 variables. 
Moreover, CP and MIP do not apply the same optimization strategies as explained in the literature review in Section \ref{sec:Literature}, which can also lead to very different results. 
However, its performance seems to depend on the considered objective function. A general observation is that instances following a random demand pattern are easier to solve than those mimicking real-world demand.

Since the CP Optimizer solves all instances for the makespan and time balance objectives to optimality, we decide to test two additional very large scale instances with $l$=1,000 and $l$=10,000 lots. With the makespan objective, optimal solutions are available for both instances in 21.76 and 315.61 seconds. However, with the time balance objective, optimality can only be proven for $l$=1,000 in 364.05 seconds; for $l$=10,000 a gap of 58,54\% is left after the allowed runtime of one hour. Although the time balance objective cannot compete with the makespan for the largest instance, our results show that CP works far better for interval-related objectives (especially the makespan) than for the resource balance objective on the tested benchmarks.

\subsection{Industry case study}\label{sec:CaseStudy_Results}
In the following, we present our results of a case study for a globally operating steel producer. The considered products are steel slabs, which are large and bulky artefacts cast out of different sorts of metal. The manufacturer requires an optimized schedule starting with predefined continuous casting programs and ending with customer deliveries. The objective of this study is threefold: We evaluate our proposed models in terms of satisfying all constraints, providing optimal solutions and giving insights into the impact of the respective objectives on processing time lengths and resource utilizations. The manufacturer requests a maximum runtime of 1 hour. The reason is that the optimization results serve as a management decision support, which has to be available at short notice for a large-scale instance size. Since up to now, the operational production planning of the company is partially triggered manually or with spreadsheet software, unfortunately, we cannot compare our results with existing schedules. As it is not possible to publish the whole real-world data instance with the detailed project structure, we give the following company-released information.

The continuous casting plan for the following 2.5 hours has to be taken into account for the optimization, i.e. all customer orders (=lots), which are cast in the next 2.5 hours have to be included in the scheduling plan. However, the overall planning horizon, which has to be considered for the optimization is 3 days (and not 2.5 hours), since the delivery dates for the produced lots vary up to 3 days. This inclusion of the whole production cycle is also necessary since the company wants to generate new production schedules in this make-to-order environment as often as demanded for the already mentioned management decision support, as it is for example also explained in \cite{voss2007hybrid}. The steel producer asks for a minute-by-minute planning, resulting in a total planning horizon of $\mathcal{T}$=4,088 minutes. Our partner produces $\mathcal{L}$=50 lots with related 556 activities. Whereas 21 lots have three route alternatives, 23 have two and six lots have to follow a fixed route. Some production activities have fixed processing times, such as automated handover activities. For other activities, strongly varying minimum ($a_i$) and maximum allowed processing times ($b_i$) are defined by the steel producer, e.g. having a storaging activity with $a_i$=24 and $b_i$=506 minutes. The manufacturer has eight renewable resources with very different capacities $C_r$ = [10; 10; 30; 10; 50; 240; 220; 80], since vehicles, a cooling bed and an automated handover resource with a much lower capacity than large warehouses, cooling and warming boxes are included. For some production resources such as the automated handover resource, concurrent use is not possible, i.e., it is always fully used by a maximum of one activity. Other resources such as warehouses have high and not fully utilized capacities. The company only considers their two conventional warehouses $r \in \{6,8\}$ as appropriate for load balancing purposes $\mathcal{R^*} \subseteq \mathcal{R}$. All other resources have very different technical purposes, i.e. they are not considered since e.g. a load balancing between a handover and a vehicle resource would not make sense for the company.

With the presented MIP models, it is not possible to generate a feasible solution within the allowed runtime. We assume that the major reason is the time index $t$ for the time-related decision variables together with the very long planning horizon of $\mathcal{T}$=4,088 minutes, since preliminary tests with a very low (unrealistic) time horizon provide at least a feastible solution. Thus, all presented results are obtained by means of the CP Optimizer.

\begin{figure}[!h]
	\centering\includegraphics[width=0.9\textwidth]{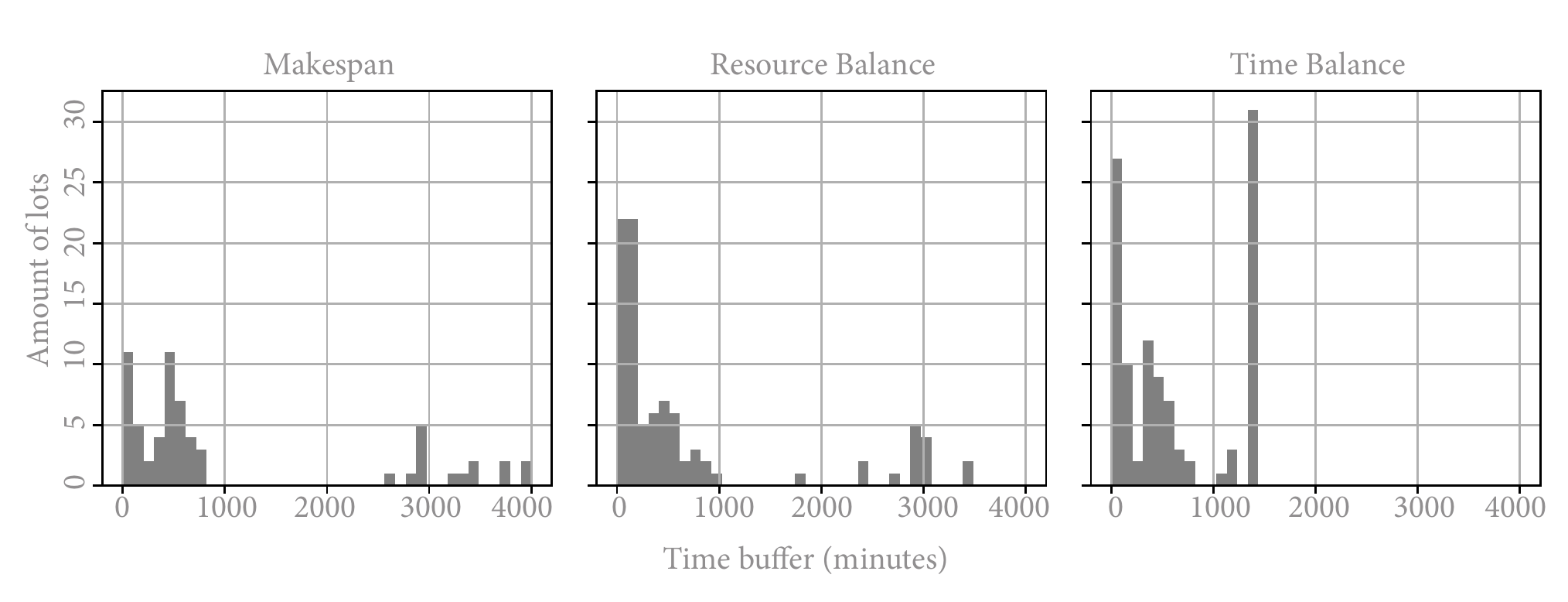}
	\caption{Time balance variations for all considered objectives.}
	\label{FIG_ResultsTimeBalanceVariations}
\end{figure}

\begin{figure}[!h]
	\centering\includegraphics[width=0.5\textwidth]{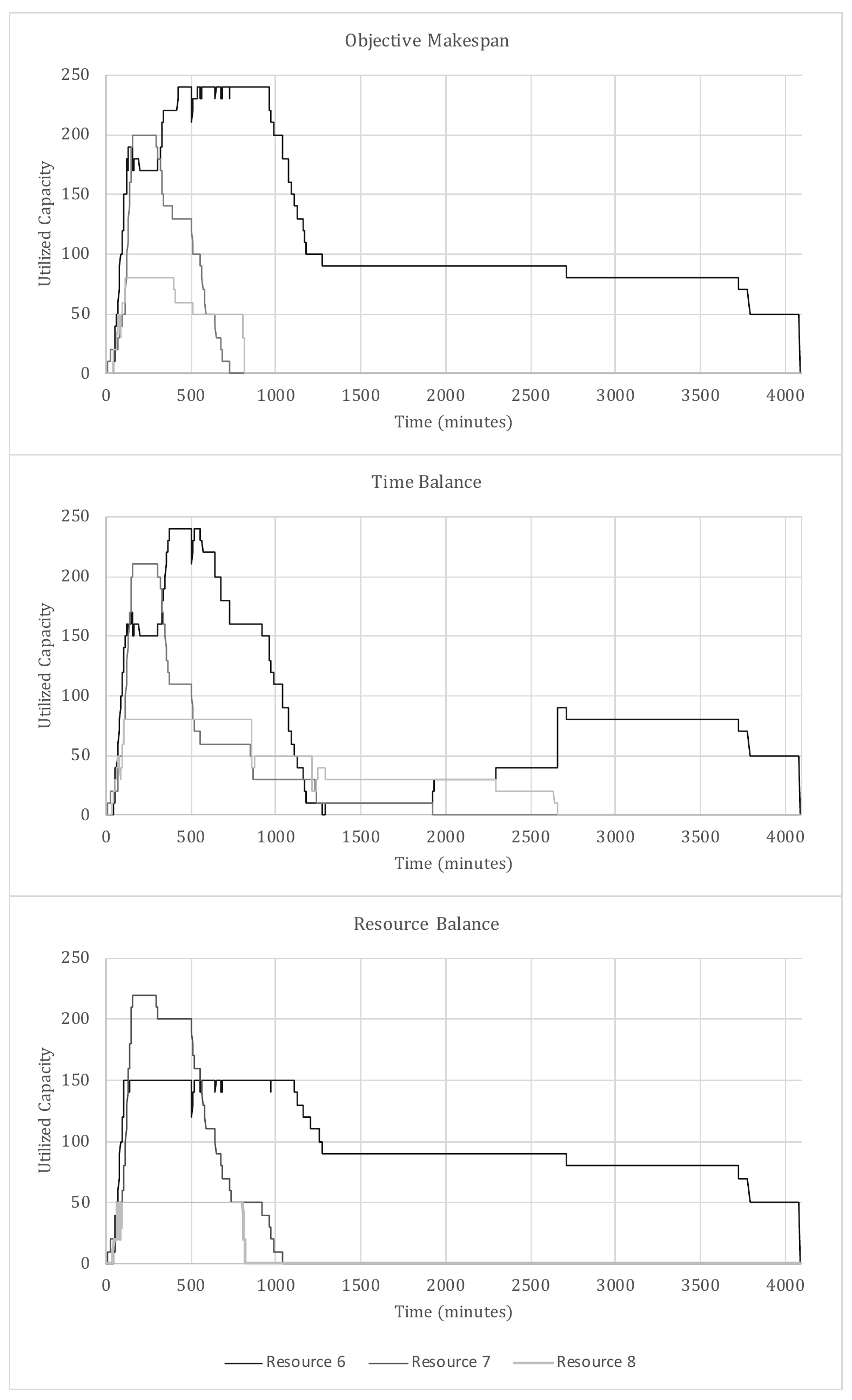}
	\caption{Resource utilization diagrams for all considered objectives.}
	\label{FIG_ResultsResourceUtilization}
\end{figure}

For the CP optimization, 50 additional dummy meta nodes which are necessary for the route selection and 1 dummy start and 1 dummy end node are added, resulting in a total of $\mathcal{J}$=608 activities (for the MIP optimization only 1 dummy start and 1 dummy end node were added). Nevertheless, all three CP models are solved to optimality within seconds, as shown in Table \ref{FIG_ResultsCase}, where the following information is provided for each run: the optimization runtime, the makespan of the obtained solution, the time buffer and the peak usage of resources 6 and 8 (R6 and R8). The optimal makespan is achieved with all three objectives. The reason for this equality is that the due date of the last lot is reached without any tardiness by all three objectives. However, the goal of balancing the peak usages of resources $\mathcal{R^*}$ is much better reached with the resource balance objective than with the others. The same result applies to the time balance objective: the aim of the company to distribute processing times in such a way that buffer time variations are minimized is far better reached with this objective in contrary to the other ones. A related closer examination of the time buffer and peak usage variations is depicted in Figures \ref{FIG_ResultsTimeBalanceVariations}-\ref{FIG_ResultsResourceUtilization}.
\begin{table}[!h]
	\caption{Constraint programming results for the real-world case study.}
	\vspace{1.0ex}
	\label{FIG_ResultsCase}
	\centering\includegraphics[width=1.0\textwidth]{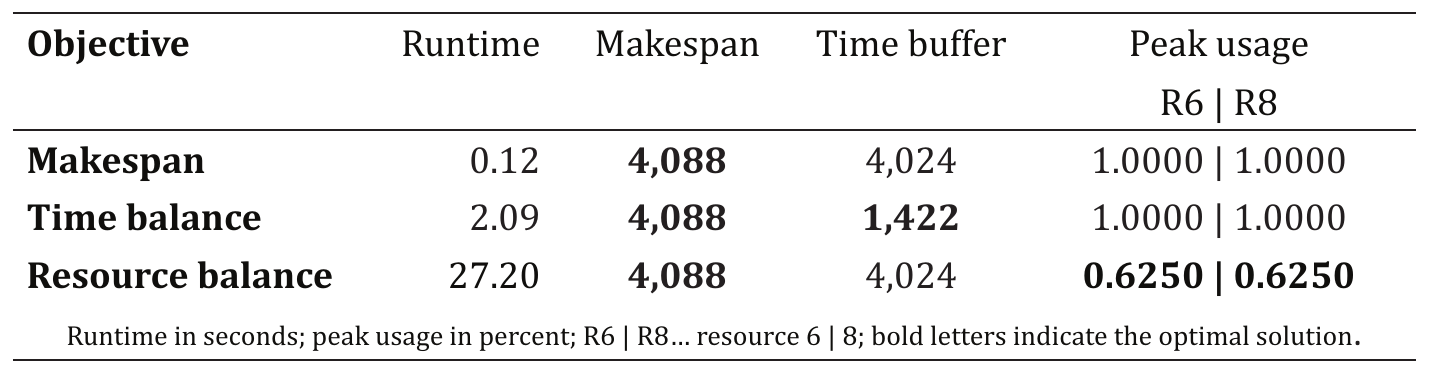}
\end{table}

In principle, our industry partner gives no tardiness restrictions except the overall planning horizon of three days due to special agreements with their customers. However, the company is interested in the amount of customer orders, which are provided too late, and the length of every delay. We decide to carry out an additional analysis to examine if it is even necessary to allow tardiness or if a full delivery date reliability or early delivery was possible with the existing alternative route and capacity restrictions. As a result, we study the following three scenarios: (a) tardiness is allowed, earliness is not allowed (current situation); (b) no earliness or tardiness is allowed; (c) earliness is allowed, tardiness is not allowed.

Scenario (a) exactly complies with constraints (\ref{NB45}) presented in Section \ref{CP-makespan} and corresponds to the real-world case study examined so far (see Table \ref{FIG_ResultsCase}). The makespan and time balance objectives do not bring any tardiness. For the resource balance objective, we have three delayed lots with a maximum delay of 6.16\% with respect to the delivery time of the delayed lot. For scenario (b), we introduce constraints (\ref{NB45b}b) in order to prescribe on-time delivery: 

\makeatletter
\let\reftagform@=\tagform@
\def\tagform@#1{\maketag@@@{(#1b\unskip\@@italiccorr)}} 
\renewcommand{\eqref}[1]{\textup{\reftagform@{\ref{#1}}}}
\makeatother
\nolinenumbers\setcounter{equation}{44}
\nolinenumbers
\vspace{-0,3cm}\begin{equation}
\texttt{endOf}(w_i)=d_i \qquad \forall \; i \in \mathcal{L}  \label{NB45b} \vspace{-0,3cm}
\end{equation}
The solutions for scenario (b) are presented in Table \ref{TAB_FIG_ResultsEarliness1}. All lots are delivered on time, all constraints are satisfied and the problems are solved to optimality. Due to the required on-time delivery of all lots, we now have a different optimal result for the resource balance objective (a higher peak usage), i.e. the capacity utilizations have been changed during the optimization process in order to guarantee the demanded on-time delivery. Moreover, the resource balance optimization needs nearly five times longer than for scenario (a) but is still finished in under three minutes.
\begin{table}[!h]
	\caption{Optimization results for scenario (b).}
	\vspace{1.0ex}
	\label{TAB_FIG_ResultsEarliness1}
	\centering
	\includegraphics[width=0.9\textwidth]{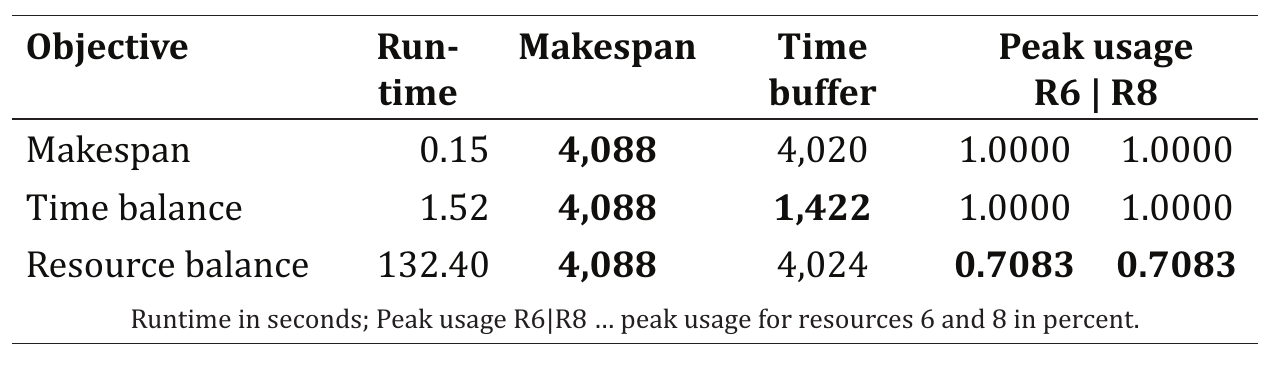}
\end{table}

In scenario (c), we want to find out if the produced lots could be delivered earlier. Therefore, we introduce the weighting factor $v$ to allow a reduction of the delivery time $d_i$. It indicates the maximum allowed earliness $e_i=d_i \cdot (1- v)$ compared to the original due date $d_i$. E.g. having the original delivery time $d_i=50$ with the factor $v=0.10$, a lot can be delivered 10\% prior to the delivery time, resulting in the maximum allowed earliness of $e_i=50 \cdot (1-0.1)=45$.

\makeatletter
\let\reftagform@=\tagform@
\def\tagform@#1{\maketag@@@{(#1c\unskip\@@italiccorr)}} 
\renewcommand{\eqref}[1]{\textup{\reftagform@{\ref{#1}}}}
\makeatother
\nolinenumbers\setcounter{equation}{44}
\nolinenumbers
\vspace{-0,5cm}\begin{equation}
e_i \leq \texttt{endOf}(w_i) \leq d_i \qquad \forall \; i \in \mathcal{L}  \label{NB45c} \vspace{-0,3cm}
\end{equation}
With constraints (\ref{NB45}c), all lots can be delivered earlier and must be at least delivered on time. In Table \ref{TAB_FIG_ResultsEarliness2_ADAPT}, the optimization results for the earliness restrictions (45c) are presented, including an allowed maximum early delivery of 10\%, 50\% and 90\% (=maximum allowed early delivery of 408, 2044 and 3679 minutes) for a comprehensive evaluation of existing earliness possibilities. In columns 1-3, the considered earliness factor $v$, the employed objective and the optimization runtime in seconds is given. Column 4 shows the amount of lots that are delivered early. The makespan, time buffer and peak usage results for every employed objective are presented in columns 5-8; the peak usages are specified in percent for balancing resources 6 and 8. The utilized earliness in columns 9-10 represents the achieved minimum and maximum early delivery in absolute numbers (minutes), e.g. having $v=0.10$ as the first case in Table \ref{TAB_FIG_ResultsEarliness2_ADAPT}, we observe that for the objective makespan, all lots are delivered between 13 and 408 minutes prior to their respective due date.

The results in Table \ref{TAB_FIG_ResultsEarliness2_ADAPT} show that it is possible to introduce earliness for all lots. All restrictions are satisfied and all runs lead to optimal solutions for the respective employed objective. It is also shown that the optimal solution for the resource balance objective (0.6250) is the same for all runs in contrary to the results of the other two objectives. We assume that the reason is the possibility of a minimized equally proportioned peak usage at the cost of activity selection, starting time postponement and processing time variations in combination with very different determined early delivery dates.

Moreover, the maximum allowed earliness is exploited in all solutions. However, the results of the minimum utilized earliness are not the same for all test cases. It is much lower than the maximum one and the higher the earliness factor $v$ is, the higher the minimum utilized earliness gets. We think that the reason lies in the very dissimilar capacity utilizations (and related peak usages) of the three employed objectives. Depending on the objective function, the selection of the activities, the starting times and the durations of single activities are very different. This results in very diverse earliness values for the due dates and resource utilizations for all objectives. We therefore assume that capacity efficiency suffers losses at the cost of time efficiency and vice versa, i.e. that the time objectives on the one hand and the resource objective on the other hand are contradictory goals. Overall, it can be concluded that the analysis of different tardiness scenarios also allows the satisfaction of all constraints and the generation of optimal production schedules and additionally gives insights into the possibility of postponing delivery times.
\begin{table}[!h]
	\caption{Optimization results for tardiness scenario (c): earliness is allowed, tardiness not.}
	\vspace{1.0ex}
	\label{TAB_FIG_ResultsEarliness2_ADAPT}
	\centering
	\includegraphics[width=1.0\textwidth]{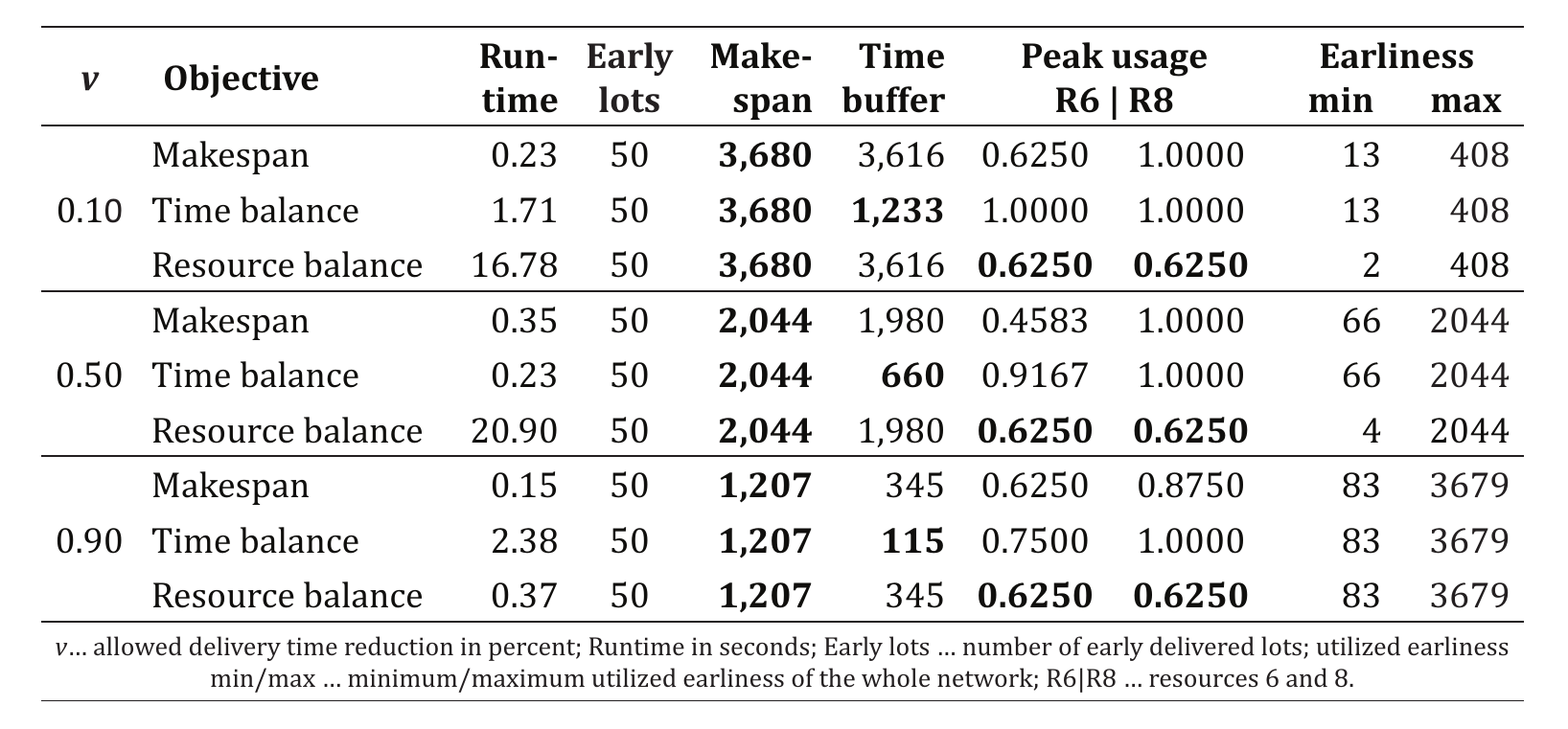}
\end{table}

\section{Conclusion}\label{sec:Conclusion}
In this work, a new resource-constrained multi-project scheduling problem with alternative activity chains and time flexibility (RCMPSP-ACTF) has been proposed. With this integrated problem, inspired by different manufacturing industries and based on the RCPSP-AC of \cite{tao2017scheduling}, it is possible to integrate several decisions on flexibility. Besides the RCPSP typical choice on a start time for every activity, multiple projects are regarded, it is decided on the selection of alternative activities and on the length of the processing times of selected activities and early deliverys are not allowed. Moreover, two new objective functions considering time balance and resource balance maximization have been developed besides the consideration of the popular objective of makespan minimization. New MIP and CP models have been presented and in a comprehensive numerical study, the strong potential of our developed CP approach has been demonstrated in terms of solution quality and runtime: we solve all benchmarks for the RCPSP-AC to optimality and therefore provide an additional set where the computationally more expensive multi-project environment is considered. Moreover, many instances of the newly developed problem and the industry case are solved to optimality.

Future research will address several topics concerning the advancement of flexibility in a scheduling context. First, new methods, which provide better solutions for the multi-project RCPSP-AC and the RCMPSP-ACTF maximizing resource balance, should be addressed. Second, the mutual influence of the investigated three objectives should be examined in a multi-objective context. Third, an additional consideration of the MP approach for the new problem and a comparison to the established SP approach would be an interesting topic of investigation. Moreover, the robustness of the achieved optimization results with regard to disruptive incidents, such as the sudden breakdown of resources, can significantly influence the competitiveness of organizations. In order to meet this challenge, disturbance variables and related methods to obtain robust solutions should be investigated.

\begin{singlespace}
	%\bibliography{CaIE_bibfile}

\end{singlespace}

\setstretch{1.25}
\counterwithin{table}{section}
\newpage
\begin{appendices}
	\section{Online appendix: Constraint programming model explanations}\label{AppendixA}
	The CP modeling process works with different functions and expressions. At the beginning of the solution process of the CP Optimizer, constraint propagation is employed and several search heuristics are applied. For the presented CP models, one can distinguish between decision variables, expressions for decision variables and resource functions \citep{laborie2018ibm,laborie2009ibm,vilim2015failure}. Moreover, a CP model can be solved with and without an objective function. If an objective function is given, it is considered as another constraint within the solution process and the solver tries to find the optimal solution for this objective function. Besides the \textit{interval}, the \texttt{alternative} expression, the \texttt{span} expression and the resource function \texttt{cumulFunction}, which are introduced in the main paper, logical relation expressions and time expressions are used for the CP models in this work:
	
	\textit{Logical relation expressions}: With the expression $\texttt{presenceOf}(w_j)$, the mandatory presence of interval variables is defined.
	
	\textit{Time expressions}: With the expressions $\texttt{endAtStart}(w_i,w_j)$ and $\texttt{endBefore}$ $\texttt{Start}(w_i,w_j)$, time positions of intervals are defined. Hence, two consecutive activities are processed without or with allowed idle times. With the expressions $\texttt{startOf}(w_j)$, $\texttt{endOf}(w_j)$, and $\texttt{lengthOf}(w_j)$, the start and end time and the exact processing time (=duration or length) of an $\textit{interval}(w_j)$ are determined.
	
	For a further detailed description of constraint programming and the CP Optimizer, we refer to \cite{laborie2018ibm,laborie2009ibm,vilim2015failure} and the online tutorial of the CP Optimizer Tutorial\footnote{https://www.ibm.com/analytics/cplex-cp-optimizer}.

	\newpage 
	\makeatletter
	\let\reftagform@=\tagform@
	\def\tagform@#1{\maketag@@@{(b.#1\unskip\@@italiccorr)}} 
	\renewcommand{\eqref}[1]{\textup{\reftagform@{\ref{#1}}}}
	\makeatother
	%\counterwithin{equation}{section}
	\section{Online appendix: Test design procedure for the RCMPSP-ACTF benchmarks}\label{AppendixB}
	For the evaluation of our models, three different test instance classes $|\mathcal{L} \in \{10,50,100\}|$ are used. Depending on the customer orders, different alternative activities and thus, alternative routes (=activity chains) are necessary. All lots and related alternative routes are sampled out of a steel company’s customer orders. Since alternative routes can consist of different numbers of activities, the overall number of activities per lot (and thus, per instance class) varies. It can for example be the case that the first lot has three different alternative routes with involved activities 3, 4, and 5 and the second lot has two different routes, involving activities 3 and 5. Therefore, the number of non-dummy activities per instance class is depicted as an average number (rounded up to the nearest integer) in Table \ref{TAB_B.1}.
	
	\begin{table}[!h]
		\caption{Parameters for test classes of test instances for the RCMPSP-ACTF.}
		\label{TAB_B.1}
		\centering
		\includegraphics[width=0.65\textwidth]{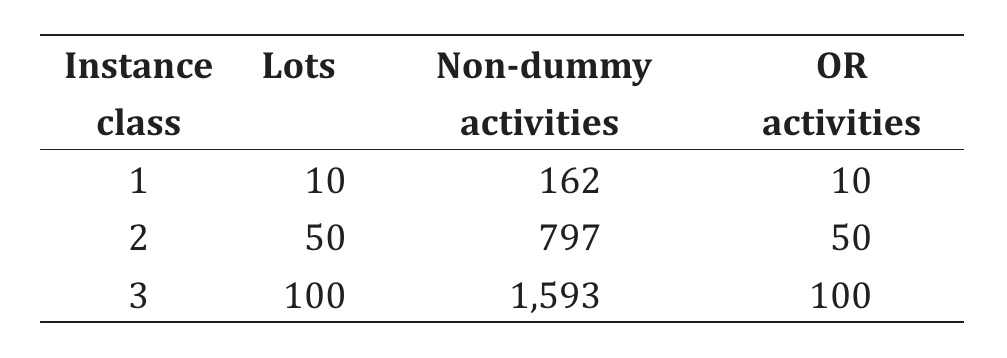}
	\end{table}
	
In the real-world situation, every activity only demands a single resource and the demand is normalized to 1. Thus, the demand pattern random (\textit{rand}) is introduced additionally to the real-world (\textit{rw}) situation. In the demand pattern \textit{rand}, every activity demands random amounts of all resources. The demand, the related resource factor $RF_R$, the resource strength $RS_R$, the calculation for the shortest processing time per activity $a_j$ and the slack $\mathfrak{s}$ are presented in Table \ref{TAB_B.2}. The resource factor 
	$RF_R \in [0,1]$ is defined as explained in \cite{kolisch1995characterization}. It describes the number of resources $r$ used by each activity $j$:
	
	\nolinenumbers\setcounter{equation}{0}
	\begin{equation}
	RF_R=\frac{1}{|\mathcal{J}|}\frac{1}{|\mathcal{R}|}\sum_{j \in \mathcal{J}}\sum_{r \in \mathcal{R}} 
	\begin{cases} 
	1, & \textit{if } c_{jr} > 0 \\ 
	0, & \textit{else} 
	\end{cases}
	\qquad  \forall \; r \in \mathcal{R}
	\label{b.1}
	\end{equation}
	
	The resource strength $RS_R\in[0,1]$ determines the resource availability and is used as a scaling parameter to determine the resource availability $C_r$ \citep{kolisch1995characterization}. 
	
	\renewcommand{\arraystretch}{1.35}
	\begin{table}[!h]
	\caption{Parameters for test instance classes real-world (\textit{rw}) and random (\textit{rand}).}
	\label{TAB_B.2}
		\centering
		\begin{tabular}{lcc}
			\hline
			& \textbf{\textit{rw}} & \textbf{\textit{rand}}
			\\
			\hline
			$c_{jr}$ 		& 	1 				&	random[1;10)
			\\
			$RF_R$			&	0.11			&	1
			\\
			$RS_R$			&	\multicolumn{2}{c}{\{0.25,0.50,0.75,1.00\}}
			\\
			$a_j$			&	\multicolumn{2}{c}{random[1;5)}	
			\\
			$\mathfrak{s}$	&	\multicolumn{2}{c}{random[10;20]}		
			\\
			\hline
		\end{tabular}
	\end{table}
	
	In addition to the parameters in Table \ref{TAB_B.2}, we compute the necessary values for the due date of every lot $d_l$, the maximum duration $b_j$, and the resource availability $C_r$. The due date is calculated by the earliest possible release date per lot $l$ plus a randomly generated value between the completion time in an earliest schedule $\mathfrak{t_l}$ and the double value of $\mathfrak{t_l}$:
	
	\nolinenumbers \begin{equation}
	d_l=\mathfrak{t_l}+\text{random}[\mathfrak{t_l};2\mathfrak{t_l}] \qquad \forall \; l \in \mathcal{L}
	\label{b.2}
	\end{equation}
	The maximum duration $b_j$ for the activities with flexible processing time lengths is given by the difference between the lot’s due date $d_l$ and its completion time $\mathfrak{t_l}$ in an earliest schedule with a slack $\mathfrak{s}$ to allow tardiness: 
	\begin{equation}
	b_j=d_l-\mathfrak{t_l}+\mathfrak{s} \qquad \forall \; j \in \mathcal{J}, \; l \in \mathcal{L}, \qquad if \; \textbf{1}_{jl}=1
	\label{b.3}
	\end{equation}
	where $\textbf{1}_{jl}$ denotes the indicator function, i.e. activity $j$ and $l$ belonging to the same lot (=project). For the computation of resource availability $C_r$, \cite{kolisch1995characterization} define the minimum and maximum demands $C_r^{\;min}$ and $C_r^{\;max}$ as input parameters besides the already explained resource strength $RS_R$:
	
	\nolinenumbers \begin{equation}
	C_r^{\;min}=\text{max}\{c_{jr}\;|\;j=2,...,J\} \qquad \forall \; r \in \mathcal{R}
	\label{b.4}
	\end{equation}
	\begin{equation}
	C_r^{\;max}=\text{max}\bigg\{\sum_{j \in \mathcal{J}}c_{jr}\;|\;j=2,...,J\bigg\} \qquad \forall \; r \in \mathcal{R}
	\label{b.5}
	\end{equation}
	
	The minimum capacity $C_r^{\;min}$ in (b.\ref{b.4}) is determined as the maximum demand of an activity $j$ for resource $r$. The maximum capacity $C_r^{\;max}$ in (b.\ref{b.5}) is the peak demand calculated out of the earliest start time schedule under consideration of all precedence relations \citep{kolisch1995characterization}. In our project structure, multiple projects can run in parallel. Moreover, the selection of one alternative route per lot and the decision on the processing time lengths of single activities have to be regarded. Thus, resource availabilities would be too low when only considering (b.\ref{b.3})-(b.\ref{b.4}). They have to be adapted in a way that they are high enough to get feasible solutions. Therefore, we determine the average amount of projects $L^{par}$ that would be active concurrently. We compare the [release;due) intervals of all lots to satisfy the necessary consideration of parallel running projects. As a result, we extend $C_r^{\;min}$ and $C_r^{\;max}$ of \cite{kolisch1995characterization} for the RCMPSP-ACTF to the values $C_r^{\;lower}$ and $C_r^{\;upper}$ to generate feasible resource availabilities $C_r$ in the following way:
	
	\nolinenumbers \begin{equation}
	C_r^{\;lower}=C_r^{\;min} \cdot L^{par}
	\label{b.6}
	\end{equation}
	\begin{equation}
	C_r^{\;upper}=C_r^{\;lower} + C_r^{\;max}
	\label{b.7}
	\end{equation}
	\begin{equation}
	C_r=C_r^{\;lower}+RS_R \cdot (C_r^{\;upper}-C_r^{\;lower})
	\label{b.8}
	\end{equation}
	
	By multiplying $C_r^{\;min}$ with $L^{par}$ in (b.\ref{b.6}), resource capacities are set in such a way that the parallel execution of different projects is possible. Since the $C_r^{\;upper}$ in (b.\ref{b.7}) has to be higher than $C_r^{\;lower}$, this lower bound is added to $C_r^{\;max}$ to guarantee this requirement. In (b.\ref{b.8}) it can be seen that $C_r^{\;lower}$ and $C_r^{\;upper}$ are used instead of the originally introduced $C_r^{\;min}$ and $C_r^{\;max}$ of \cite{kolisch1995characterization} to calculate feasible resource availabilities. \\ For the real-world (\textit{rw}) instances, the resource availabilities $C_r$ for non-load balancing resources $\mathcal{R} \setminus \mathcal{R}^*$ are calculated in a different way. They are set to two times $C_r^{\;min}$ instead of using (b.\ref{b.8}) since this corresponds to the real-world case. However, for all \textit{rand} instances, formulae (b.\ref{b.6})-(b.\ref{b.8}) are applied.

	\newpage
	\makeatletter
	\let\reftagform@=\tagform@
	\def\tagform@#1{\maketag@@@{(c.#1\unskip\@@italiccorr)}} 
	\renewcommand{\eqref}[1]{\textup{\reftagform@{\ref{#1}}}}
	\makeatother
	\section{Online appendix: Adapted MIP model for the RCPSP-AC}\label{AppendixC}	
	With the following MIP model, alternative activity chains are considered for the optimization of one or multiple projects. In order to validate the optimization results for the RCPSP with alternative activity chains (RCPSP-AC), we implemented the MIP model presented by \cite{tao2017scheduling} and tested it with the presented benchmark instance $\mathcal{J}=30$ in their paper. It is an extended version of the RCPSP-AC originally presented by \cite{tao2017scheduling}, since we added constraint (c.\ref{c.2}) and adapted constraints (c.\ref{c.4})-(c.\ref{c.6}) to obtain the following model and hence, the same solutions as presented in their paper:
	\\ \\
	\setcounter{equation}{0}
	\nolinenumbers \noindent \textit{Minimize} %\vspace{-0,5cm}
	\begin{equation}
	\sum_{t \in \mathcal{T}} t \cdot  y_{n+1t}  								\label{c.1}
	\end{equation}%\vspace{-0,5cm}
	\\ \textit{subject to}%\vspace{-0,5cm}
	\begin{equation}
	x_{0} = 1,														\label{c.2}
	\end{equation}%\vspace{-0,5cm}
	\begin{equation}
	\sum_{t \in \mathcal{T}} y_{it}= x_i 		\qquad  \forall \; i \in \mathcal{J},	\label{c.3}
	\end{equation}%\vspace{-0,5cm}
	\begin{equation}
	\sum_{j \in \mathcal{J}} A_{ij} \cdot x_j = x_i \qquad \forall \; i \in \mathcal{J}, \; if \; p_i=0, \label{c.4}	
	\end{equation}%\vspace{-0,5cm}
	\begin{equation}
	A_{ij} \cdot x_i \leq x_j 		\qquad \forall \; i,j \in \mathcal{J}, \; if \; p_i=1,		\label{c.5}
	\end{equation}%\vspace{-0,5cm}
	\begin{equation}
	A_{ij} \bigg(\sum_{t\in\mathcal{T}}t \cdot y_{it}\bigg)+(x_j+x_i-2)\cdot M \leq \sum_{t\in\mathcal{T}}t \cdot y_{jt} - D_j \qquad \forall \; i,j \in \mathcal{J},
	\label{c.6}
	\end{equation}%\vspace{-0,5cm}
	\begin{equation}
	\sum_{i \in \mathcal{J}}\sum_{\tau=t}^{t+D_i-1}y_{i\tau} \cdot c_{ir} \leq C_r	\qquad r \in \mathcal{R}, t \in \mathcal{T},														\label{c.7}
	\end{equation}%\vspace{-0,5cm}
	\begin{equation}
	\sum_{i \in \mathcal{J}} x_{i} \cdot c_{ir} \leq C_r	\qquad r \in \mathcal{N}, t \in \mathcal{T},												\label{c.8}
	\end{equation}%\vspace{-0,5cm}
	\begin{equation}
	x_i \in \{0,1\} 										\qquad \forall \; i \in \mathcal{J},														\label{c.9}
	\end{equation}%\vspace{-0,5cm}
	\begin{equation}
	y_{it} \in \{0,1\}  \qquad \forall \; i \in \mathcal{J}, t \in \mathcal{T}.				\label{c.10}
	\end{equation}
	
	Objective function (c.\ref{c.1}) minimizes the makespan of the project. With the newly added constraint (c.\ref{c.2}), the project (production process) has to be started. Without the consideration of this condition, an optimization leads to a result of 0. Restrictions (c.\ref{c.3}) define that every selected activity has to be finished exactly once. With altered constraints (c.\ref{c.4})-(c.\ref{c.5}) activity selection flexibility relations are considered. If an activity is an OR node, only one of its successors in the project network must be selected. If an activity is an AND node, all successors have to be selected. Modified restrictions (c.\ref{c.6}) guarantee that no activity within one production route can be started before the predecessor activities of this route are finished and that only activities can be selected which are related to each other. Idle times are allowed, also between successor activities and not only between those of different lots. Conditions (c.\ref{c.7})-(c.\ref{c.8}) make sure that capacity restrictions for renewable and non-renewable resources are met. Constraints (c.\ref{c.9})-(c.\ref{c.10}) define decision variables as binary ones.
	
	We note that with the consideration of the two activity types AND/OR ($p_j=1 / p_j=0$), it can happen that additional nodes appear in the solution of an optimization although they do not belong to the chosen alternative route of the optimizer. This is possible since there is no restriction in the MIP model to select exactly one activity route after an OR activity and no additional activities out of other alternatives. These additional selected activities do not increase or influence a minimization objective since they are considered as a separate schedule by the MIP model optimization. They can be deleted in a manual post-processing step. Alternatively, a third activity type $p_j=2$ for every customer delivery node $j$ and a related new constraint, which forbids the explained additional node selection, can be introduced:
	\begin{equation}
	\sum_{i \in \mathcal{J}} A_{ij} \cdot x_j = x_j \qquad \forall \; j \in \mathcal{J}, \; if \; p_j=2. \label{c.11}	
	\end{equation}
	With constraint (c.\ref{c.11}) it is guaranteed that if one production route is selected, no additional activities of other routes within the project (lot) can be selected. Preliminary experiments showed that it is more efficient to use the post-processing step. Therefore, we use this approach in our experimental results. (We note that in contrary to the here presented model where this activity type $p_j=2$ is a free choice, it is a mandatory activity type for the newly presented RCMPSP-ACTF in this work as explained in Section \ref{MIP-Notations}.)

	\newpage
	\section{Online appendix: Optimization results for the RCPSP-AC and the RCMPSP-AC}\label{AppendixD}	
	On the following pages, we present the detailed examination of all designed benchmark instances for the resource-constrained project scheduling problem with alternative activity chains (RCPSP-AC) and its multi-project version, the RCMPSP-AC. The columns in Table D.1 give (1) the name of every instance, (2) the solver (MIP or CP) which has been applied, (3) the project-scheduling type, i.e. if it is a single- (RCPSP-AC) or a multi-project(RCMPSP-AC) instance, (4) the considered amount of nodes, (5) an identification number from 1-10 per project and node case, (6) the best objective value found, (7) the best bound found, and (8) the runtime in seconds. Bold letters represent the optimal solution.
	
	\newpage
	\begin{table}[!h]
	\caption{Optimization results for the RCPSP-AC \& the RCMPSP-AC for every instance.}
	\vspace{1.0ex}
	\label{TAB_D.1}		
	\centering\includepdf[pages={1},scale=0.74]{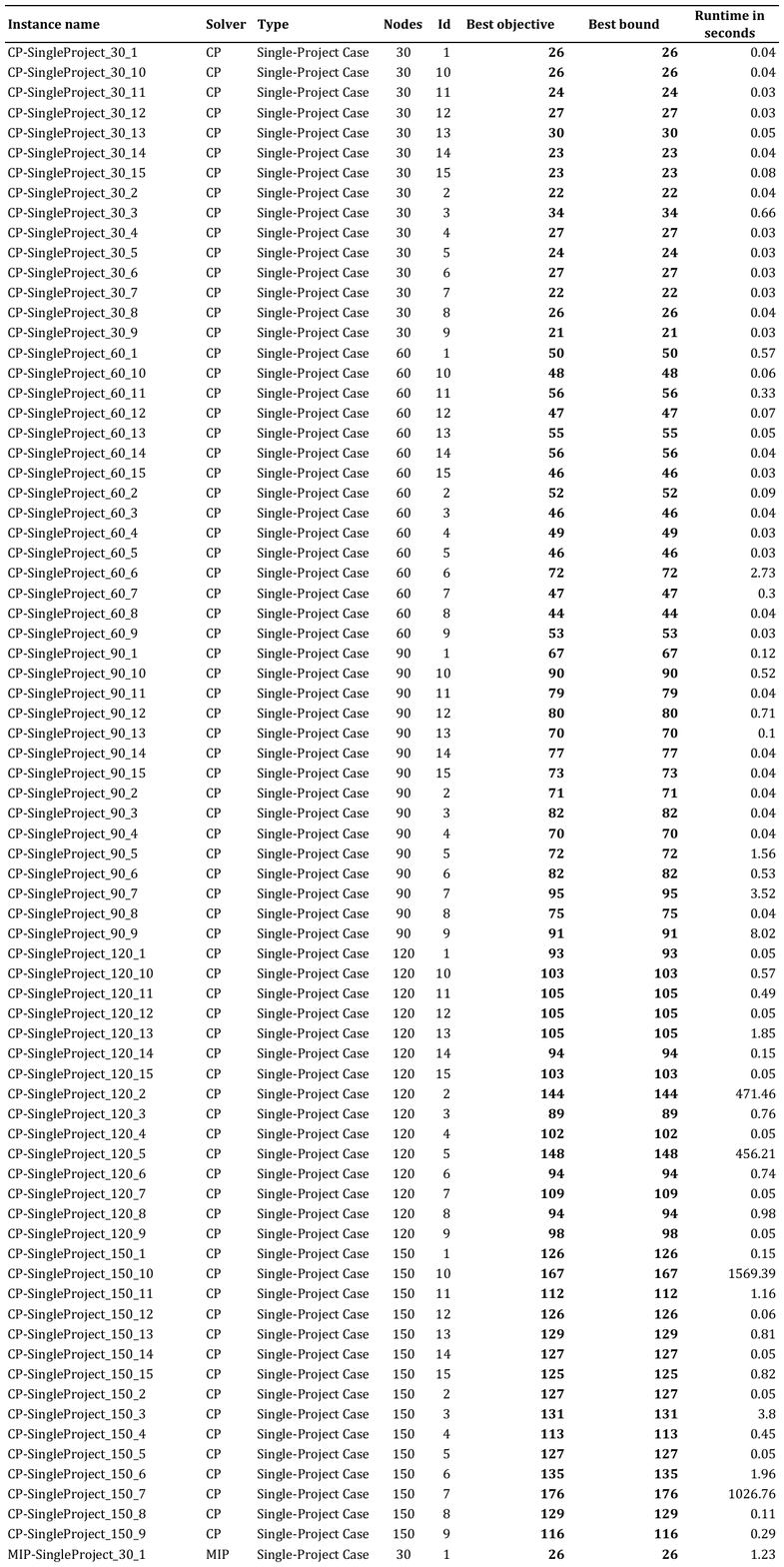}
	\end{table}
	\includepdf[pages={2-3},scale=0.74]{FIG_RCPSP-AC_Detailed-Results}
	\begin{table}[!h]
		\includepdf[pages={4},scale=0.74]{FIG_RCPSP-AC_Detailed-Results}
	\end{table}
	
	\newpage
	\section{Online appendix: Optimization results for the RCMPSP-ACTF}\label{AppendixE}	
	On the following pages, we present the detailed examination of all designed benchmark instances for the newly developed resource-constrained multi-project scheduling problem with alternative activity chains and time flexibility (RCMPSP-ACTF). The columns in Table E.1 give (1) the name of every instance, (2) the solver (MIP or CP) which has been applied, (3) the considered amount of nodes, (4) the demand patterns real-world (rw) or random (rand), (5) an identification number from 1-5 per project and node case, (6) the considered objective function, (7) the best objective value found, (8) the best bound found, and (9) the runtime in seconds. Bold letters represent the optimal solution.
	
	\newpage
	\begin{table}[!h]
	\caption{Optimization results for the RCMPSP-ACTF for every instance.}
	\vspace{1.0ex}
	\label{TAB_E.1}
	\centering\includepdf[pages={1},scale=0.74]{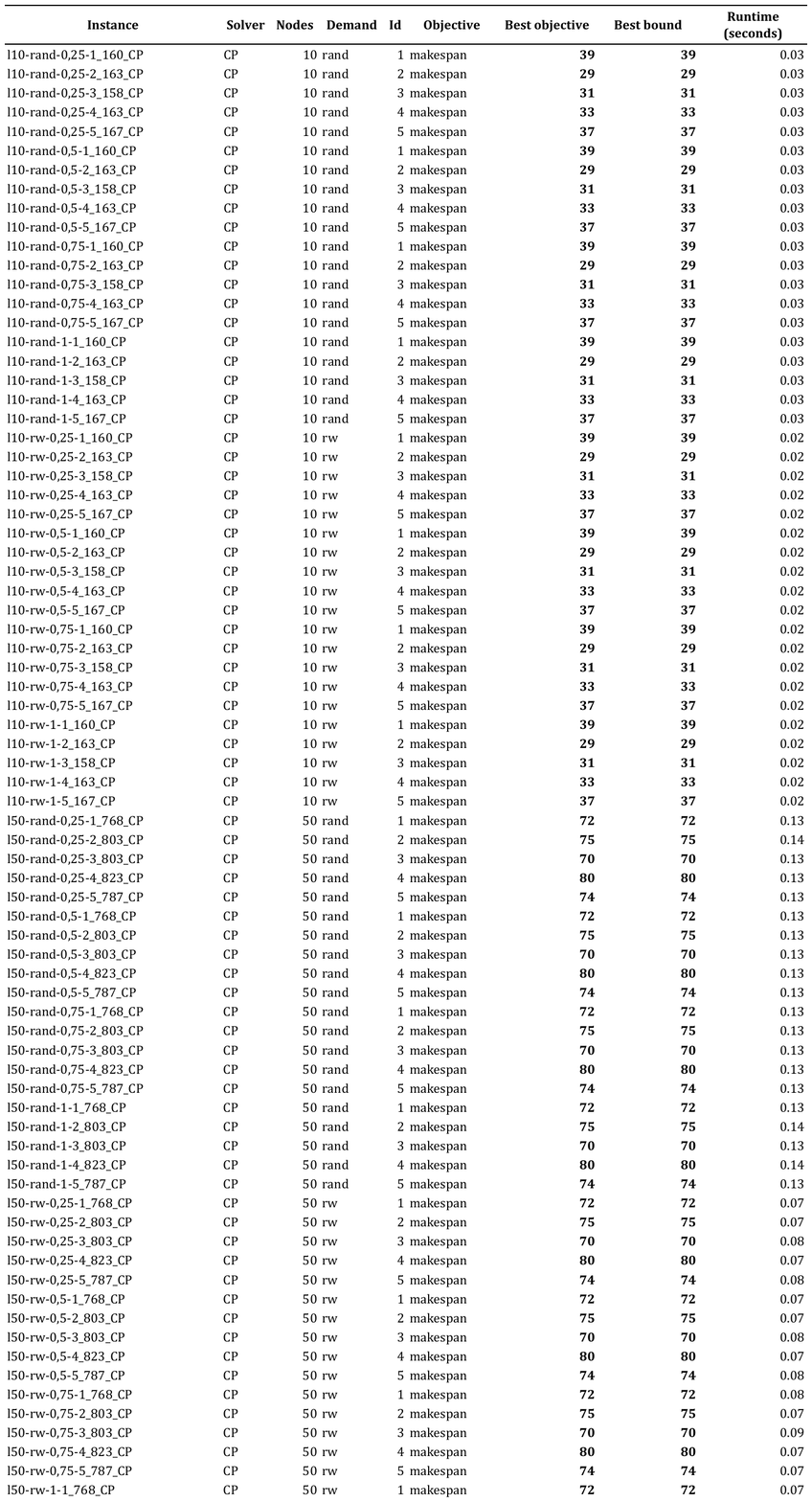}
	\end{table}
	\includepdf[pages={2-9},scale=0.74]{FIG_RCMPSP-ACTF_Detailed-Results}
	\begin{table}[!h]
	

\section*{Supplementary material}

Table E.1: Optimization results for the RCMPSP-ACTF for every instance.

| Instance | Solver | Nodes | Demand | Id | Objective | Best objective | Best bound | Runtime (seconds) |
|---|---|---|---|---|---|---|---|---|
| l10-rand-0,25-1_160_CP | CP | 10 | rand | 1 | makespan | 39 | 39 | 0.03 |
| l10-rand-0,25-2_163_CP | CP | 10 | rand | 2 | makespan | 29 | 29 | 0.03 |
| l10-rand-0,25-3_158_CP | CP | 10 | rand | 3 | makespan | 31 | 31 | 0.03 |
| l10-rand-0,25-4_163_CP | CP | 10 | rand | 4 | makespan | 33 | 33 | 0.03 |
| l10-rand-0,25-5_167_CP | CP | 10 | rand | 5 | makespan | 37 | 37 | 0.03 |
| l10-rand-0,5-1_160_CP | CP | 10 | rand | 1 | makespan | 39 | 39 | 0.03 |
| l10-rand-0,5-2_163_CP | CP | 10 | rand | 2 | makespan | 29 | 29 | 0.03 |
| l10-rand-0,5-3_158_CP | CP | 10 | rand | 3 | makespan | 31 | 31 | 0.03 |
| l10-rand-0,5-4_163_CP | CP | 10 | rand | 4 | makespan | 33 | 33 | 0.03 |
| l10-rand-0,5-5_167_CP | CP | 10 | rand | 5 | makespan | 37 | 37 | 0.03 |
| l10-rand-0,75-1_160_CP | CP | 10 | rand | 1 | makespan | 39 | 39 | 0.03 |
| l10-rand-0,75-2_163_CP | CP | 10 | rand | 2 | makespan | 29 | 29 | 0.03 |
| l10-rand-0,75-3_158_CP | CP | 10 | rand | 3 | makespan | 31 | 31 | 0.03 |
| l10-rand-0,75-4_163_CP | CP | 10 | rand | 4 | makespan | 33 | 33 | 0.03 |
| l10-rand-0,75-5_167_CP | CP | 10 | rand | 5 | makespan | 37 | 37 | 0.03 |
| l10-rand-1-1_160_CP | CP | 10 | rand | 1 | makespan | 39 | 39 | 0.03 |
| l10-rand-1-2_163_CP | CP | 10 | rand | 2 | makespan | 29 | 29 | 0.03 |
| l10-rand-1-3_158_CP | CP | 10 | rand | 3 | makespan | 31 | 31 | 0.03 |
| l10-rand-1-4_163_CP | CP | 10 | rand | 4 | makespan | 33 | 33 | 0.03 |
| l10-rand-1-5_167_CP | CP | 10 | rand | 5 | makespan | 37 | 37 | 0.03 |
| l10-rw-0,25-1_160_CP | CP | 10 | rw | 1 | makespan | 39 | 39 | 0.02 |
| l10-rw-0,25-2_163_CP | CP | 10 | rw | 2 | makespan | 29 | 29 | 0.02 |
| l10-rw-0,25-3_158_CP | CP | 10 | rw | 3 | makespan | 31 | 31 | 0.02 |
| l10-rw-0,25-4_163_CP | CP | 10 | rw | 4 | makespan | 33 | 33 | 0.02 |
| l10-rw-0,25-5_167_CP | CP | 10 | rw | 5 | makespan | 37 | 37 | 0.02 |
| l10-rw-0,5-1_160_CP | CP | 10 | rw | 1 | makespan | 39 | 39 | 0.02 |
| l10-rw-0,5-2_163_CP | CP | 10 | rw | 2 | makespan | 29 | 29 | 0.02 |
| l10-rw-0,5-3_158_CP | CP | 10 | rw | 3 | makespan | 31 | 31 | 0.02 |
| l10-rw-0,5-4_163_CP | CP | 10 | rw | 4 | makespan | 33 | 33 | 0.02 |
| l10-rw-0,5-5_167_CP | CP | 10 | rw | 5 | makespan | 37 | 37 | 0.02 |
| l10-rw-0,75-1_160_CP | CP | 10 | rw | 1 | makespan | 39 | 39 | 0.02 |
| l10-rw-0,75-2_163_CP | CP | 10 | rw | 2 | makespan | 29 | 29 | 0.02 |
| l10-rw-0,75-3_158_CP | CP | 10 | rw | 3 | makespan | 31 | 31 | 0.02 |
| l10-rw-0,75-4_163_CP | CP | 10 | rw | 4 | makespan | 33 | 33 | 0.02 |
| l10-rw-0,75-5_167_CP | CP | 10 | rw | 5 | makespan | 37 | 37 | 0.02 |
| l10-rw-1-1_160_CP | CP | 10 | rw | 1 | makespan | 39 | 39 | 0.02 |
| l10-rw-1-2_163_CP | CP | 10 | rw | 2 | makespan | 29 | 29 | 0.02 |
| l10-rw-1-3_158_CP | CP | 10 | rw | 3 | makespan | 31 | 31 | 0.02 |
| l10-rw-1-4_163_CP | CP | 10 | rw | 4 | makespan | 33 | 33 | 0.02 |
| l10-rw-1-5_167_CP | CP | 10 | rw | 5 | makespan | 37 | 37 | 0.02 |
| l50-rand-0,25-1_768_CP | CP | 50 | rand | 1 | makespan | 72 | 72 | 0.13 |
| l50-rand-0,25-2_803_CP | CP | 50 | rand | 2 | makespan | 75 | 75 | 0.14 |
| l50-rand-0,25-3_803_CP | CP | 50 | rand | 3 | makespan | 70 | 70 | 0.13 |
| l50-rand-0,25-4_823_CP | CP | 50 | rand | 4 | makespan | 80 | 80 | 0.13 |
| l50-rand-0,25-5_787_CP | CP | 50 | rand | 5 | makespan | 74 | 74 | 0.13 |
| l50-rand-0,5-1_768_CP | CP | 50 | rand | 1 | makespan | 72 | 72 | 0.13 |
| l50-rand-0,5-2_803_CP | CP | 50 | rand | 2 | makespan | 75 | 75 | 0.13 |
| l50-rand-0,5-3_803_CP | CP | 50 | rand | 3 | makespan | 70 | 70 | 0.13 |
| l50-rand-0,5-4_823_CP | CP | 50 | rand | 4 | makespan | 80 | 80 | 0.13 |
| l50-rand-0,5-5_787_CP | CP | 50 | rand | 5 | makespan | 74 | 74 | 0.13 |
| l50-rand-0,75-1_768_CP | CP | 50 | rand | 1 | makespan | 72 | 72 | 0.13 |
| l50-rand-0,75-2_803_CP | CP | 50 | rand | 2 | makespan | 75 | 75 | 0.13 |
| l50-rand-0,75-3_803_CP | CP | 50 | rand | 3 | makespan | 70 | 70 | 0.13 |
| l50-rand-0,75-4_823_CP | CP | 50 | rand | 4 | makespan | 80 | 80 | 0.13 |
| l50-rand-0,75-5_787_CP | CP | 50 | rand | 5 | makespan | 74 | 74 | 0.13 |
| l50-rand-1-1_768_CP | CP | 50 | rand | 1 | makespan | 72 | 72 | 0.13 |
| l50-rand-1-2_803_CP | CP | 50 | rand | 2 | makespan | 75 | 75 | 0.14 |
| l50-rand-1-3_803_CP | CP | 50 | rand | 3 | makespan | 70 | 70 | 0.13 |
| l50-rand-1-4_823_CP | CP | 50 | rand | 4 | makespan | 80 | 80 | 0.14 |
| l50-rand-1-5_787_CP | CP | 50 | rand | 5 | makespan | 74 | 74 | 0.13 |
| l50-rw-0,25-1_768_CP | CP | 50 | rw | 1 | makespan | 72 | 72 | 0.07 |
| l50-rw-0,25-2_803_CP | CP | 50 | rw | 2 | makespan | 75 | 75 | 0.07 |
| l50-rw-0,25-3_803_CP | CP | 50 | rw | 3 | makespan | 70 | 70 | 0.08 |
| l50-rw-0,25-4_823_CP | CP | 50 | rw | 4 | makespan | 80 | 80 | 0.07 |
| l50-rw-0,25-5_787_CP | CP | 50 | rw | 5 | makespan | 74 | 74 | 0.08 |
| l50-rw-0,5-1_768_CP | CP | 50 | rw | 1 | makespan | 72 | 72 | 0.07 |
| l50-rw-0,5-2_803_CP | CP | 50 | rw | 2 | makespan | 75 | 75 | 0.07 |
| l50-rw-0,5-3_803_CP | CP | 50 | rw | 3 | makespan | 70 | 70 | 0.08 |
| l50-rw-0,5-4_823_CP | CP | 50 | rw | 4 | makespan | 80 | 80 | 0.07 |
| l50-rw-0,5-5_787_CP | CP | 50 | rw | 5 | makespan | 74 | 74 | 0.08 |
| l50-rw-0,75-1_768_CP | CP | 50 | rw | 1 | makespan | 72 | 72 | 0.08 |
| l50-rw-0,75-2_803_CP | CP | 50 | rw | 2 | makespan | 75 | 75 | 0.07 |
| l50-rw-0,75-3_803_CP | CP | 50 | rw | 3 | makespan | 70 | 70 | 0.09 |
| l50-rw-0,75-4_823_CP | CP | 50 | rw | 4 | makespan | 80 | 80 | 0.07 |
| l50-rw-0,75-5_787_CP | CP | 50 | rw | 5 | makespan | 74 | 74 | 0.07 |
| l50-rw-1-1_768_CP | CP | 50 | rw | 1 | makespan | 72 | 72 | 0.07 |

| Instance | Solver | Nodes | Demand | Id | Objective | Best objective | Best bound | Runtime (seconds) |
|---|---|---|---|---|---|---|---|---|
| l50-rw-1-2_803_CP | CP | 50 | rw | 2 | makespan | 75 | 75 | 0.07 |
| l50-rw-1-3_803_CP | CP | 50 | rw | 3 | makespan | 70 | 70 | 0.09 |
| l50-rw-1-4_823_CP | CP | 50 | rw | 4 | makespan | 80 | 80 | 0.07 |
| l50-rw-1-5_787_CP | CP | 50 | rw | 5 | makespan | 74 | 74 | 0.08 |
| l100-rand-0,25-1_1557_CP | CP | 100 | rand | 1 | makespan | 125 | 125 | 0.29 |
| l100-rand-0,25-2_1605_CP | CP | 100 | rand | 2 | makespan | 121 | 121 | 0.28 |
| l100-rand-0,25-3_1580_CP | CP | 100 | rand | 3 | makespan | 128 | 128 | 0.29 |
| l100-rand-0,25-4_1649_CP | CP | 100 | rand | 4 | makespan | 129 | 129 | 0.29 |
| l100-rand-0,25-5_1573_CP | CP | 100 | rand | 5 | makespan | 122 | 122 | 0.30 |
| l100-rand-0,5-1_1557_CP | CP | 100 | rand | 1 | makespan | 125 | 125 | 0.27 |
| l100-rand-0,5-2_1605_CP | CP | 100 | rand | 2 | makespan | 121 | 121 | 0.28 |
| l100-rand-0,5-3_1580_CP | CP | 100 | rand | 3 | makespan | 128 | 128 | 0.29 |
| l100-rand-0,5-4_1649_CP | CP | 100 | rand | 4 | makespan | 129 | 129 | 0.28 |
| l100-rand-0,5-5_1573_CP | CP | 100 | rand | 5 | makespan | 122 | 122 | 0.28 |
| l100-rand-0,75-1_1557_CP | CP | 100 | rand | 1 | makespan | 125 | 125 | 0.28 |
| l100-rand-0,75-2_1605_CP | CP | 100 | rand | 2 | makespan | 121 | 121 | 0.29 |
| l100-rand-0,75-3_1580_CP | CP | 100 | rand | 3 | makespan | 128 | 128 | 0.28 |
| l100-rand-0,75-4_1649_CP | CP | 100 | rand | 4 | makespan | 129 | 129 | 0.29 |
| l100-rand-0,75-5_1573_CP | CP | 100 | rand | 5 | makespan | 122 | 122 | 0.29 |
| l100-rand-1-1_1557_CP | CP | 100 | rand | 1 | makespan | 125 | 125 | 0.28 |
| l100-rand-1-2_1605_CP | CP | 100 | rand | 2 | makespan | 121 | 121 | 0.29 |
| l100-rand-1-3_1580_CP | CP | 100 | rand | 3 | makespan | 128 | 128 | 0.28 |
| l100-rand-1-4_1649_CP | CP | 100 | rand | 4 | makespan | 129 | 129 | 0.30 |
| l100-rand-1-5_1573_CP | CP | 100 | rand | 5 | makespan | 122 | 122 | 0.28 |
| l100-rw-0,25-1_1557_CP | CP | 100 | rw | 1 | makespan | 125 | 125 | 0.14 |
| l100-rw-0,25-2_1605_CP | CP | 100 | rw | 2 | makespan | 121 | 121 | 0.15 |
| l100-rw-0,25-3_1580_CP | CP | 100 | rw | 3 | makespan | 128 | 128 | 0.15 |
| l100-rw-0,25-4_1649_CP | CP | 100 | rw | 4 | makespan | 129 | 129 | 0.15 |
| l100-rw-0,25-5_1573_CP | CP | 100 | rw | 5 | makespan | 122 | 122 | 0.16 |
| l100-rw-0,5-1_1557_CP | CP | 100 | rw | 1 | makespan | 125 | 125 | 0.14 |
| l100-rw-0,5-2_1605_CP | CP | 100 | rw | 2 | makespan | 121 | 121 | 0.15 |
| l100-rw-0,5-3_1580_CP | CP | 100 | rw | 3 | makespan | 128 | 128 | 0.15 |
| l100-rw-0,5-4_1649_CP | CP | 100 | rw | 4 | makespan | 129 | 129 | 0.15 |
| l100-rw-0,5-5_1573_CP | CP | 100 | rw | 5 | makespan | 122 | 122 | 0.16 |
| l100-rw-0,75-1_1557_CP | CP | 100 | rw | 1 | makespan | 125 | 125 | 0.14 |
| l100-rw-0,75-2_1605_CP | CP | 100 | rw | 2 | makespan | 121 | 121 | 0.15 |
| l100-rw-0,75-3_1580_CP | CP | 100 | rw | 3 | makespan | 128 | 128 | 0.15 |
| l100-rw-0,75-4_1649_CP | CP | 100 | rw | 4 | makespan | 129 | 129 | 0.15 |
| l100-rw-0,75-5_1573_CP | CP | 100 | rw | 5 | makespan | 122 | 122 | 0.16 |
| l100-rw-1-1_1557_CP | CP | 100 | rw | 1 | makespan | 125 | 125 | 0.14 |
| l100-rw-1-2_1605_CP | CP | 100 | rw | 2 | makespan | 121 | 121 | 0.15 |
| l100-rw-1-3_1580_CP | CP | 100 | rw | 3 | makespan | 128 | 128 | 0.16 |
| l100-rw-1-4_1649_CP | CP | 100 | rw | 4 | makespan | 129 | 129 | 0.15 |
| l100-rw-1-5_1573_CP | CP | 100 | rw | 5 | makespan | 122 | 122 | 0.16 |
| l10-rand-0,25-1_160_CP | CP | 10 | rand | 1 | timebalance | 4 | 4 | 0.10 |
| l10-rand-0,25-2_163_CP | CP | 10 | rand | 2 | timebalance | 2 | 2 | 0.10 |
| l10-rand-0,25-3_158_CP | CP | 10 | rand | 3 | timebalance | 3 | 3 | 0.12 |
| l10-rand-0,25-4_163_CP | CP | 10 | rand | 4 | timebalance | 3 | 3 | 0.11 |
| l10-rand-0,25-5_167_CP | CP | 10 | rand | 5 | timebalance | 3 | 3 | 0.11 |
| l10-rand-0,5-1_160_CP | CP | 10 | rand | 1 | timebalance | 4 | 4 | 0.10 |
| l10-rand-0,5-2_163_CP | CP | 10 | rand | 2 | timebalance | 2 | 2 | 0.09 |
| l10-rand-0,5-3_158_CP | CP | 10 | rand | 3 | timebalance | 3 | 3 | 0.11 |
| l10-rand-0,5-4_163_CP | CP | 10 | rand | 4 | timebalance | 3 | 3 | 0.09 |
| l10-rand-0,5-5_167_CP | CP | 10 | rand | 5 | timebalance | 3 | 3 | 0.11 |
| l10-rand-0,75-1_160_CP | CP | 10 | rand | 1 | timebalance | 4 | 4 | 0.10 |
| l10-rand-0,75-2_163_CP | CP | 10 | rand | 2 | timebalance | 2 | 2 | 0.10 |
| l10-rand-0,75-3_158_CP | CP | 10 | rand | 3 | timebalance | 3 | 3 | 0.10 |
| l10-rand-0,75-4_163_CP | CP | 10 | rand | 4 | timebalance | 3 | 3 | 0.08 |
| l10-rand-0,75-5_167_CP | CP | 10 | rand | 5 | timebalance | 3 | 3 | 0.11 |
| l10-rand-1-1_160_CP | CP | 10 | rand | 1 | timebalance | 4 | 4 | 0.10 |
| l10-rand-1-2_163_CP | CP | 10 | rand | 2 | timebalance | 2 | 2 | 0.10 |
| l10-rand-1-3_158_CP | CP | 10 | rand | 3 | timebalance | 3 | 3 | 0.10 |
| l10-rand-1-4_163_CP | CP | 10 | rand | 4 | timebalance | 3 | 3 | 0.08 |
| l10-rand-1-5_167_CP | CP | 10 | rand | 5 | timebalance | 3 | 3 | 0.10 |
| l10-rw-0,25-1_160_CP | CP | 10 | rw | 1 | timebalance | 4 | 4 | 0.06 |
| l10-rw-0,25-2_163_CP | CP | 10 | rw | 2 | timebalance | 2 | 2 | 0.07 |
| l10-rw-0,25-3_158_CP | CP | 10 | rw | 3 | timebalance | 3 | 3 | 0.07 |
| l10-rw-0,25-4_163_CP | CP | 10 | rw | 4 | timebalance | 3 | 3 | 0.07 |
| l10-rw-0,25-5_167_CP | CP | 10 | rw | 5 | timebalance | 3 | 3 | 0.07 |
| l10-rw-0,5-1_160_CP | CP | 10 | rw | 1 | timebalance | 4 | 4 | 0.06 |
| l10-rw-0,5-2_163_CP | CP | 10 | rw | 2 | timebalance | 2 | 2 | 0.07 |
| l10-rw-0,5-3_158_CP | CP | 10 | rw | 3 | timebalance | 3 | 3 | 0.07 |
| l10-rw-0,5-4_163_CP | CP | 10 | rw | 4 | timebalance | 3 | 3 | 0.08 |
| l10-rw-0,5-5_167_CP | CP | 10 | rw | 5 | timebalance | 3 | 3 | 0.08 |
| l10-rw-0,75-1_160_CP | CP | 10 | rw | 1 | timebalance | 4 | 4 | 0.06 |
| l10-rw-0,75-2_163_CP | CP | 10 | rw | 2 | timebalance | 2 | 2 | 0.07 |

| Instance | Solver | Nodes | Demand | Id | Objective | Best objective | Best bound | Runtime (seconds) |
|---|---|---|---|---|---|---|---|---|
| l10-rw-0,75-3_158_CP | CP | 10 | rw | 3 | timebalance | **3** | **3** | 0.07 |
| l10-rw-0,75-4_163_CP | CP | 10 | rw | 4 | timebalance | **3** | **3** | 0.07 |
| l10-rw-0,75-5_167_CP | CP | 10 | rw | 5 | timebalance | **3** | **3** | 0.08 |
| l10-rw-1-1_160_CP | CP | 10 | rw | 1 | timebalance | **4** | **4** | 0.06 |
| l10-rw-1-2_163_CP | CP | 10 | rw | 2 | timebalance | **2** | **2** | 0.07 |
| l10-rw-1-3_158_CP | CP | 10 | rw | 3 | timebalance | **3** | **3** | 0.06 |
| l10-rw-1-4_163_CP | CP | 10 | rw | 4 | timebalance | **3** | **3** | 0.07 |
| l10-rw-1-5_167_CP | CP | 10 | rw | 5 | timebalance | **3** | **3** | 0.08 |
| l50-rand-0,25-1_768_CP | CP | 50 | rand | 1 | timebalance | **4** | **4** | 0.56 |
| l50-rand-0,25-2_803_CP | CP | 50 | rand | 2 | timebalance | **4** | **4** | 0.54 |
| l50-rand-0,25-3_803_CP | CP | 50 | rand | 3 | timebalance | **4** | **4** | 0.60 |
| l50-rand-0,25-4_823_CP | CP | 50 | rand | 4 | timebalance | **4** | **4** | 0.58 |
| l50-rand-0,25-5_787_CP | CP | 50 | rand | 5 | timebalance | **4** | **4** | 0.54 |
| l50-rand-0,5-1_768_CP | CP | 50 | rand | 1 | timebalance | **4** | **4** | 0.54 |
| l50-rand-0,5-2_803_CP | CP | 50 | rand | 2 | timebalance | **4** | **4** | 0.54 |
| l50-rand-0,5-3_803_CP | CP | 50 | rand | 3 | timebalance | **4** | **4** | 0.59 |
| l50-rand-0,5-4_823_CP | CP | 50 | rand | 4 | timebalance | **4** | **4** | 0.67 |
| l50-rand-0,5-5_787_CP | CP | 50 | rand | 5 | timebalance | **4** | **4** | 0.56 |
| l50-rand-0,75-1_768_CP | CP | 50 | rand | 1 | timebalance | **4** | **4** | 0.52 |
| l50-rand-0,75-2_803_CP | CP | 50 | rand | 2 | timebalance | **4** | **4** | 0.51 |
| l50-rand-0,75-3_803_CP | CP | 50 | rand | 3 | timebalance | **4** | **4** | 0.58 |
| l50-rand-0,75-4_823_CP | CP | 50 | rand | 4 | timebalance | **4** | **4** | 0.61 |
| l50-rand-0,75-5_787_CP | CP | 50 | rand | 5 | timebalance | **4** | **4** | 0.55 |
| l50-rand-1-1_768_CP | CP | 50 | rand | 1 | timebalance | **4** | **4** | 0.53 |
| l50-rand-1-2_803_CP | CP | 50 | rand | 2 | timebalance | **4** | **4** | 0.51 |
| l50-rand-1-3_803_CP | CP | 50 | rand | 3 | timebalance | **4** | **4** | 0.56 |
| l50-rand-1-4_823_CP | CP | 50 | rand | 4 | timebalance | **4** | **4** | 0.58 |
| l50-rand-1-5_787_CP | CP | 50 | rand | 5 | timebalance | **4** | **4** | 0.53 |
| l50-rw-0,25-1_768_CP | CP | 50 | rw | 1 | timebalance | **4** | **4** | 0.51 |
| l50-rw-0,25-2_803_CP | CP | 50 | rw | 2 | timebalance | **4** | **4** | 0.44 |
| l50-rw-0,25-3_803_CP | CP | 50 | rw | 3 | timebalance | **4** | **4** | 0.64 |
| l50-rw-0,25-4_823_CP | CP | 50 | rw | 4 | timebalance | **4** | **4** | 0.59 |
| l50-rw-0,25-5_787_CP | CP | 50 | rw | 5 | timebalance | **4** | **4** | 0.53 |
| l50-rw-0,5-1_768_CP | CP | 50 | rw | 1 | timebalance | **4** | **4** | 0.52 |
| l50-rw-0,5-2_803_CP | CP | 50 | rw | 2 | timebalance | **4** | **4** | 0.59 |
| l50-rw-0,5-3_803_CP | CP | 50 | rw | 3 | timebalance | **4** | **4** | 0.64 |
| l50-rw-0,5-4_823_CP | CP | 50 | rw | 4 | timebalance | **4** | **4** | 0.59 |
| l50-rw-0,5-5_787_CP | CP | 50 | rw | 5 | timebalance | **4** | **4** | 0.53 |
| l50-rw-0,75-1_768_CP | CP | 50 | rw | 1 | timebalance | **4** | **4** | 0.45 |
| l50-rw-0,75-2_803_CP | CP | 50 | rw | 2 | timebalance | **4** | **4** | 0.54 |
| l50-rw-0,75-3_803_CP | CP | 50 | rw | 3 | timebalance | **4** | **4** | 0.83 |
| l50-rw-0,75-4_823_CP | CP | 50 | rw | 4 | timebalance | **4** | **4** | 0.68 |
| l50-rw-0,75-5_787_CP | CP | 50 | rw | 5 | timebalance | **4** | **4** | 0.52 |
| l50-rw-1-1_768_CP | CP | 50 | rw | 1 | timebalance | **4** | **4** | 0.52 |
| l50-rw-1-2_803_CP | CP | 50 | rw | 2 | timebalance | **4** | **4** | 0.43 |
| l50-rw-1-3_803_CP | CP | 50 | rw | 3 | timebalance | **4** | **4** | 0.64 |
| l50-rw-1-4_823_CP | CP | 50 | rw | 4 | timebalance | **4** | **4** | 0.67 |
| l50-rw-1-5_787_CP | CP | 50 | rw | 5 | timebalance | **4** | **4** | 0.52 |
| l100-rand-0,25-1_1557_CP | CP | 100 | rand | 1 | timebalance | **4** | **4** | 1.75 |
| l100-rand-0,25-2_1605_CP | CP | 100 | rand | 2 | timebalance | **4** | **4** | 1.79 |
| l100-rand-0,25-3_1580_CP | CP | 100 | rand | 3 | timebalance | **4** | **4** | 1.97 |
| l100-rand-0,25-4_1649_CP | CP | 100 | rand | 4 | timebalance | **5** | **5** | 1.87 |
| l100-rand-0,25-5_1573_CP | CP | 100 | rand | 5 | timebalance | **5** | **5** | 1.55 |
| l100-rand-0,5-1_1557_CP | CP | 100 | rand | 1 | timebalance | **4** | **4** | 1.66 |
| l100-rand-0,5-2_1605_CP | CP | 100 | rand | 2 | timebalance | **4** | **4** | 2.03 |
| l100-rand-0,5-3_1580_CP | CP | 100 | rand | 3 | timebalance | **4** | **4** | 1.71 |
| l100-rand-0,5-4_1649_CP | CP | 100 | rand | 4 | timebalance | **5** | **5** | 1.66 |
| l100-rand-0,5-5_1573_CP | CP | 100 | rand | 5 | timebalance | **5** | **5** | 1.55 |
| l100-rand-0,75-1_1557_CP | CP | 100 | rand | 1 | timebalance | **4** | **4** | 1.80 |
| l100-rand-0,75-2_1605_CP | CP | 100 | rand | 2 | timebalance | **4** | **4** | 1.45 |
| l100-rand-0,75-3_1580_CP | CP | 100 | rand | 3 | timebalance | **4** | **4** | 1.63 |
| l100-rand-0,75-4_1649_CP | CP | 100 | rand | 4 | timebalance | **5** | **5** | 1.46 |
| l100-rand-0,75-5_1573_CP | CP | 100 | rand | 5 | timebalance | **5** | **5** | 1.24 |
| l100-rand-1-1_1557_CP | CP | 100 | rand | 1 | timebalance | **4** | **4** | 1.86 |
| l100-rand-1-2_1605_CP | CP | 100 | rand | 2 | timebalance | **4** | **4** | 1.49 |
| l100-rand-1-3_1580_CP | CP | 100 | rand | 3 | timebalance | **4** | **4** | 1.52 |
| l100-rand-1-4_1649_CP | CP | 100 | rand | 4 | timebalance | **5** | **5** | 1.61 |
| l100-rand-1-5_1573_CP | CP | 100 | rand | 5 | timebalance | **5** | **5** | 1.26 |
| l100-rw-0,25-1_1557_CP | CP | 100 | rw | 1 | timebalance | **4** | **4** | 10.02 |
| l100-rw-0,25-2_1605_CP | CP | 100 | rw | 2 | timebalance | **4** | **4** | 1.98 |
| l100-rw-0,25-3_1580_CP | CP | 100 | rw | 3 | timebalance | **4** | **4** | 28.36 |
| l100-rw-0,25-4_1649_CP | CP | 100 | rw | 4 | timebalance | **5** | **5** | 5.19 |
| l100-rw-0,25-5_1573_CP | CP | 100 | rw | 5 | timebalance | **5** | **5** | 10.18 |
| l100-rw-0,5-1_1557_CP | CP | 100 | rw | 1 | timebalance | **4** | **4** | 47.87 |
| l100-rw-0,5-2_1605_CP | CP | 100 | rw | 2 | timebalance | **4** | **4** | 15.53 |
| l100-rw-0,5-3_1580_CP | CP | 100 | rw | 3 | timebalance | **4** | **4** | 14.23 |

| Instance | Solver | Nodes | Demand | Id | Objective | Best objective | Best bound | Runtime (seconds) |
|---|---|---|---|---|---|---|---|---|
| l100-rw-0,5-4_1649_CP | CP | 100 | rw | 4 | timebalance | **5** | **5** | 2.03 |
| l100-rw-0,5-5_1573_CP | CP | 100 | rw | 5 | timebalance | **5** | **5** | 6.83 |
| l100-rw-0,75-1_1557_CP | CP | 100 | rw | 1 | timebalance | **4** | **4** | 186.91 |
| l100-rw-0,75-2_1605_CP | CP | 100 | rw | 2 | timebalance | **4** | **4** | 2.09 |
| l100-rw-0,75-3_1580_CP | CP | 100 | rw | 3 | timebalance | **4** | **4** | 1.91 |
| l100-rw-0,75-4_1649_CP | CP | 100 | rw | 4 | timebalance | **5** | **5** | 1.62 |
| l100-rw-0,75-5_1573_CP | CP | 100 | rw | 5 | timebalance | **5** | **5** | 18.54 |
| l100-rw-1-1_1557_CP | CP | 100 | rw | 1 | timebalance | **4** | **4** | 21.33 |
| l100-rw-1-2_1605_CP | CP | 100 | rw | 2 | timebalance | **4** | **4** | 16.60 |
| l100-rw-1-3_1580_CP | CP | 100 | rw | 3 | timebalance | **4** | **4** | 1.52 |
| l100-rw-1-4_1649_CP | CP | 100 | rw | 4 | timebalance | **5** | **5** | 2.01 |
| l100-rw-1-5_1573_CP | CP | 100 | rw | 5 | timebalance | **5** | **5** | 3.34 |
| l10-rand-0,25-1_160_CP | CP | 10 | rand | 1 | loadbalance | **0.571429** | **0.571429** | 339.89 |
| l10-rand-0,25-2_163_CP | CP | 10 | rand | 2 | loadbalance | **0.614286** | **0.614286** | 146.76 |
| l10-rand-0,25-3_158_CP | CP | 10 | rand | 3 | loadbalance | **0.633803** | **0.633803** | 238.99 |
| l10-rand-0,25-4_163_CP | CP | 10 | rand | 4 | loadbalance | **0.60274** | **0.60274** | 524.80 |
| l10-rand-0,25-5_167_CP | CP | 10 | rand | 5 | loadbalance | **0.597403** | **0.597403** | 426.21 |
| l10-rand-0,5-1_160_CP | CP | 10 | rand | 1 | loadbalance | **0.415094** | **0.415094** | 144.90 |
| l10-rand-0,5-2_163_CP | CP | 10 | rand | 2 | loadbalance | **0.452632** | **0.452632** | 172.13 |
| l10-rand-0,5-3_158_CP | CP | 10 | rand | 3 | loadbalance | **0.463918** | **0.463918** | 244.66 |
| l10-rand-0,5-4_163_CP | CP | 10 | rand | 4 | loadbalance | **0.438776** | **0.438776** | 291.07 |
| l10-rand-0,5-5_167_CP | CP | 10 | rand | 5 | loadbalance | **0.425926** | **0.425926** | 888.89 |
| l10-rand-0,75-1_160_CP | CP | 10 | rand | 1 | loadbalance | **0.323529** | **0.323529** | 201.64 |
| l10-rand-0,75-2_163_CP | CP | 10 | rand | 2 | loadbalance | **0.358333** | **0.358333** | 288.08 |
| l10-rand-0,75-3_158_CP | CP | 10 | rand | 3 | loadbalance | **0.366667** | **0.366667** | 372.53 |
| l10-rand-0,75-4_163_CP | CP | 10 | rand | 4 | loadbalance | **0.34375** | **0.34375** | 259.90 |
| l10-rand-0,75-5_167_CP | CP | 10 | rand | 5 | loadbalance | **0.330935** | **0.330935** | 289.32 |
| l10-rand-1-1_160_CP | CP | 10 | rand | 1 | loadbalance | **0.26506** | **0.26506** | 402.17 |
| l10-rand-1-2_163_CP | CP | 10 | rand | 2 | loadbalance | **0.296552** | **0.296552** | 151.17 |
| l10-rand-1-3_158_CP | CP | 10 | rand | 3 | loadbalance | **0.303448** | **0.303448** | 255.18 |
| l10-rand-1-4_163_CP | CP | 10 | rand | 4 | loadbalance | **0.283871** | **0.283871** | 721.19 |
| l10-rand-1-5_167_CP | CP | 10 | rand | 5 | loadbalance | **0.272109** | **0.272109** | 527.45 |
| l10-rw-0,25-1_160_CP | CP | 10 | rw | 1 | loadbalance | **0.333333** | **0.333333** | 0.52 |
| l10-rw-0,25-2_163_CP | CP | 10 | rw | 2 | loadbalance | **0.333333** | **0.333333** | 0.53 |
| l10-rw-0,25-3_158_CP | CP | 10 | rw | 3 | loadbalance | **0.333333** | **0.333333** | 0.37 |
| l10-rw-0,25-4_163_CP | CP | 10 | rw | 4 | loadbalance | **0.5** | **0.5** | 0.12 |
| l10-rw-0,25-5_167_CP | CP | 10 | rw | 5 | loadbalance | **0.333333** | **0.333333** | 0.53 |
| l10-rw-0,5-1_160_CP | CP | 10 | rw | 1 | loadbalance | **0.285714** | **0.285714** | 0.46 |
| l10-rw-0,5-2_163_CP | CP | 10 | rw | 2 | loadbalance | **0.285714** | **0.285714** | 1.69 |
| l10-rw-0,5-3_158_CP | CP | 10 | rw | 3 | loadbalance | **0.285714** | **0.285714** | 0.76 |
| l10-rw-0,5-4_163_CP | CP | 10 | rw | 4 | loadbalance | **0.428571** | **0.428571** | 0.73 |
| l10-rw-0,5-5_167_CP | CP | 10 | rw | 5 | loadbalance | **0.285714** | **0.285714** | 0.56 |
| l10-rw-0,75-1_160_CP | CP | 10 | rw | 1 | loadbalance | **0.25** | **0.25** | 0.37 |
| l10-rw-0,75-2_163_CP | CP | 10 | rw | 2 | loadbalance | **0.25** | **0.25** | 0.55 |
| l10-rw-0,75-3_158_CP | CP | 10 | rw | 3 | loadbalance | **0.25** | **0.25** | 1.00 |
| l10-rw-0,75-4_163_CP | CP | 10 | rw | 4 | loadbalance | **0.375** | **0.375** | 0.52 |
| l10-rw-0,75-5_167_CP | CP | 10 | rw | 5 | loadbalance | **0.25** | **0.25** | 0.44 |
| l10-rw-1-1_160_CP | CP | 10 | rw | 1 | loadbalance | **0.25** | **0.25** | 0.86 |
| l10-rw-1-2_163_CP | CP | 10 | rw | 2 | loadbalance | **0.25** | **0.25** | 1.10 |
| l10-rw-1-3_158_CP | CP | 10 | rw | 3 | loadbalance | **0.25** | **0.25** | 2.67 |
| l10-rw-1-4_163_CP | CP | 10 | rw | 4 | loadbalance | **0.333333** | **0.333333** | 0.90 |
| l10-rw-1-5_167_CP | CP | 10 | rw | 5 | loadbalance | **0.25** | **0.25** | 3.70 |
| l50-rand-0,25-1_768_CP | CP | 50 | rand | 1 | loadbalance | 0.548387 | 0.210327 | T |
| l50-rand-0,25-2_803_CP | CP | 50 | rand | 2 | loadbalance | 0.554217 | 0.200357 | T |
| l50-rand-0,25-3_803_CP | CP | 50 | rand | 3 | loadbalance | 0.586826 | 0.213456 | T |
| l50-rand-0,25-4_823_CP | CP | 50 | rand | 4 | loadbalance | 0.532544 | 0.193621 | T |
| l50-rand-0,25-5_787_CP | CP | 50 | rand | 5 | loadbalance | 0.541667 | 0.197566 | T |
| l50-rand-0,5-1_768_CP | CP | 50 | rand | 1 | loadbalance | 0.472826 | 0.176667 | T |
| l50-rand-0,5-2_803_CP | CP | 50 | rand | 2 | loadbalance | 0.462687 | 0.168552 | T |
| l50-rand-0,5-3_803_CP | CP | 50 | rand | 3 | loadbalance | 0.482927 | 0.179592 | T |
| l50-rand-0,5-4_823_CP | CP | 50 | rand | 4 | loadbalance | 0.450495 | 0.162887 | T |
| l50-rand-0,5-5_787_CP | CP | 50 | rand | 5 | loadbalance | 0.458537 | 0.168001 | T |
| l50-rand-0,75-1_768_CP | CP | 50 | rand | 1 | loadbalance | 0.403756 | 0.152312 | T |
| l50-rand-0,75-2_803_CP | CP | 50 | rand | 2 | loadbalance | 0.394737 | 0.145418 | T |
| l50-rand-0,75-3_803_CP | CP | 50 | rand | 3 | loadbalance | 0.416667 | 0.154862 | T |
| l50-rand-0,75-4_823_CP | CP | 50 | rand | 4 | loadbalance | 0.391489 | 0.140582 | T |
| l50-rand-0,75-5_787_CP | CP | 50 | rand | 5 | loadbalance | 0.402715 | 0.145789 | T |
| l50-rand-1-1_768_CP | CP | 50 | rand | 1 | loadbalance | 0.352941 | 0.133864 | T |
| l50-rand-1-2_803_CP | CP | 50 | rand | 2 | loadbalance | 0.338521 | 0.128168 | T |
| l50-rand-1-3_803_CP | CP | 50 | rand | 3 | loadbalance | 0.360153 | 0.136294 | T |
| l50-rand-1-4_823_CP | CP | 50 | rand | 4 | loadbalance | 0.339552 | 0.123717 | T |
| l50-rand-1-5_787_CP | CP | 50 | rand | 5 | loadbalance | 0.357143 | 0.129285 | T |
| l50-rw-0,25-1_768_CP | CP | 50 | rw | 1 | loadbalance | 0.266667 | 0.133333 | T |
| l50-rw-0,25-2_803_CP | CP | 50 | rw | 2 | loadbalance | 0.25 | 0.125 | T |
| l50-rw-0,25-3_803_CP | CP | 50 | rw | 3 | loadbalance | 0.25 | 0.125 | T |
| l50-rw-0,25-4_823_CP | CP | 50 | rw | 4 | loadbalance | 0.25 | 0.125 | T |

| Instance | Solver | Nodes | Demand | Id | Objective | Best objective | Best bound | Runtime (seconds) |
|---|---|---|---|---|---|---|---|---|
| l50-rw-0,25-5_787_CP | CP | 50 | rw | 5 | loadbalance | 0.25 | 0.125 | T |
| l50-rw-0,5-1_768_CP | CP | 50 | rw | 1 | loadbalance | 0.25 | 0.133333 | T |
| l50-rw-0,5-2_803_CP | CP | 50 | rw | 2 | loadbalance | 0.235294 | 0.125 | T |
| l50-rw-0,5-3_803_CP | CP | 50 | rw | 3 | loadbalance | 0.235294 | 0.125 | T |
| l50-rw-0,5-4_823_CP | CP | 50 | rw | 4 | loadbalance | 0.235294 | 0.125 | T |
| l50-rw-0,5-5_787_CP | CP | 50 | rw | 5 | loadbalance | 0.235294 | 0.125 | T |
| l50-rw-0,75-1_768_CP | CP | 50 | rw | 1 | loadbalance | 0.235294 | 0.133333 | T |
| l50-rw-0,75-2_803_CP | CP | 50 | rw | 2 | loadbalance | 0.222222 | 0.125 | T |
| l50-rw-0,75-3_803_CP | CP | 50 | rw | 3 | loadbalance | 0.222222 | 0.125 | T |
| l50-rw-0,75-4_823_CP | CP | 50 | rw | 4 | loadbalance | 0.222222 | 0.125 | T |
| l50-rw-0,75-5_787_CP | CP | 50 | rw | 5 | loadbalance | 0.222222 | 0.125 | T |
| l50-rw-1-1_768_CP | CP | 50 | rw | 1 | loadbalance | 0.222222 | 0.133333 | T |
| l50-rw-1-2_803_CP | CP | 50 | rw | 2 | loadbalance | 0.210526 | 0.125 | T |
| l50-rw-1-3_803_CP | CP | 50 | rw | 3 | loadbalance | 0.210526 | 0.125 | T |
| l50-rw-1-4_823_CP | CP | 50 | rw | 4 | loadbalance | 0.210526 | 0.125 | T |
| l50-rw-1-5_787_CP | CP | 50 | rw | 5 | loadbalance | 0.210526 | 0.125 | T |
| l100-rand-0,25-1_1557_CP | CP | 100 | rand | 1 | loadbalance | 0.554286 | 0.244761 | T |
| l100-rand-0,25-2_1605_CP | CP | 100 | rand | 2 | loadbalance | 0.565714 | 0.254743 | T |
| l100-rand-0,25-3_1580_CP | CP | 100 | rand | 3 | loadbalance | 0.586592 | 0.238098 | T |
| l100-rand-0,25-4_1649_CP | CP | 100 | rand | 4 | loadbalance | 0.550562 | 0.233447 | T |
| l100-rand-0,25-5_1573_CP | CP | 100 | rand | 5 | loadbalance | 0.560694 | 0.251437 | T |
| l100-rand-0,5-1_1557_CP | CP | 100 | rand | 1 | loadbalance | 0.487805 | 0.208326 | T |
| l100-rand-0,5-2_1605_CP | CP | 100 | rand | 2 | loadbalance | 0.5 | 0.216216 | T |
| l100-rand-0,5-3_1580_CP | CP | 100 | rand | 3 | loadbalance | 0.485577 | 0.202001 | T |
| l100-rand-0,5-4_1649_CP | CP | 100 | rand | 4 | loadbalance | 0.473934 | 0.197304 | T |
| l100-rand-0,5-5_1573_CP | CP | 100 | rand | 5 | loadbalance | 0.477612 | 0.215531 | T |
| l100-rand-0,75-1_1557_CP | CP | 100 | rand | 1 | loadbalance | 0.414938 | 0.181234 | T |
| l100-rand-0,75-2_1605_CP | CP | 100 | rand | 2 | loadbalance | 0.418803 | 0.18762 | T |
| l100-rand-0,75-3_1580_CP | CP | 100 | rand | 3 | loadbalance | 0.422594 | 0.175258 | T |
| l100-rand-0,75-4_1649_CP | CP | 100 | rand | 4 | loadbalance | 0.404255 | 0.170982 | T |
| l100-rand-0,75-5_1573_CP | CP | 100 | rand | 5 | loadbalance | 0.411523 | 0.188082 | T |
| l100-rand-1-1_1557_CP | CP | 100 | rand | 1 | loadbalance | 0.364662 | 0.160552 | T |
| l100-rand-1-2_1605_CP | CP | 100 | rand | 2 | loadbalance | 0.369403 | 0.166089 | T |
| l100-rand-1-3_1580_CP | CP | 100 | rand | 3 | loadbalance | 0.362963 | 0.154981 | T |
| l100-rand-1-4_1649_CP | CP | 100 | rand | 4 | loadbalance | 0.355072 | 0.150907 | T |
| l100-rand-1-5_1573_CP | CP | 100 | rand | 5 | loadbalance | 0.372093 | 0.167529 | T |
| l100-rw-0,25-1_1557_CP | CP | 100 | rw | 1 | loadbalance | 0.294118 | 0.117647 | T |
| l100-rw-0,25-2_1605_CP | CP | 100 | rw | 2 | loadbalance | 0.235294 | 0.117647 | T |
| l100-rw-0,25-3_1580_CP | CP | 100 | rw | 3 | loadbalance | 0.294118 | 0.117647 | T |
| l100-rw-0,25-4_1649_CP | CP | 100 | rw | 4 | loadbalance | 0.294118 | 0.117647 | T |
| l100-rw-0,25-5_1573_CP | CP | 100 | rw | 5 | loadbalance | 0.294118 | 0.117647 | T |
| l100-rw-0,5-1_1557_CP | CP | 100 | rw | 1 | loadbalance | 0.277778 | 0.117647 | T |
| l100-rw-0,5-2_1605_CP | CP | 100 | rw | 2 | loadbalance | 0.277778 | 0.117647 | T |
| l100-rw-0,5-3_1580_CP | CP | 100 | rw | 3 | loadbalance | 0.235294 | 0.117647 | T |
| l100-rw-0,5-4_1649_CP | CP | 100 | rw | 4 | loadbalance | 0.235294 | 0.117647 | T |
| l100-rw-0,5-5_1573_CP | CP | 100 | rw | 5 | loadbalance | 0.277778 | 0.117647 | T |
| l100-rw-0,75-1_1557_CP | CP | 100 | rw | 1 | loadbalance | 0.263158 | 0.117647 | T |
| l100-rw-0,75-2_1605_CP | CP | 100 | rw | 2 | loadbalance | 0.235294 | 0.117647 | T |
| l100-rw-0,75-3_1580_CP | CP | 100 | rw | 3 | loadbalance | 0.235294 | 0.117647 | T |
| l100-rw-0,75-4_1649_CP | CP | 100 | rw | 4 | loadbalance | 0.263158 | 0.117647 | T |
| l100-rw-0,75-5_1573_CP | CP | 100 | rw | 5 | loadbalance | 0.263158 | 0.117647 | T |
| l100-rw-1-1_1557_CP | CP | 100 | rw | 1 | loadbalance | 0.25 | 0.117647 | T |
| l100-rw-1-2_1605_CP | CP | 100 | rw | 2 | loadbalance | 0.25 | 0.117647 | T |
| l100-rw-1-3_1580_CP | CP | 100 | rw | 3 | loadbalance | 0.25 | 0.117647 | T |
| l100-rw-1-4_1649_CP | CP | 100 | rw | 4 | loadbalance | 0.25 | 0.117647 | T |
| l100-rw-1-5_1573_CP | CP | 100 | rw | 5 | loadbalance | 0.25 | 0.117647 | T |
| l10-rand-0.25-1_160_MIP | MIP | 10 | rand | 1 | makespan | **39** | **39** | 8.88 |
| l10-rand-0.25-2_163_MIP | MIP | 10 | rand | 2 | makespan | **29** | **29** | 6.49 |
| l10-rand-0.25-3_158_MIP | MIP | 10 | rand | 3 | makespan | **31** | **31** | 6.31 |
| l10-rand-0.25-4_163_MIP | MIP | 10 | rand | 4 | makespan | **33** | **33** | 7.77 |
| l10-rand-0.25-5_167_MIP | MIP | 10 | rand | 5 | makespan | **37** | **37** | 8.80 |
| l10-rand-0.5-1_160_MIP | MIP | 10 | rand | 1 | makespan | **39** | **39** | 9.52 |
| l10-rand-0.5-2_163_MIP | MIP | 10 | rand | 2 | makespan | **29** | **29** | 6.98 |
| l10-rand-0.5-3_158_MIP | MIP | 10 | rand | 3 | makespan | **31** | **31** | 5.92 |
| l10-rand-0.5-4_163_MIP | MIP | 10 | rand | 4 | makespan | **33** | **33** | 7.72 |
| l10-rand-0.5-5_167_MIP | MIP | 10 | rand | 5 | makespan | **37** | **37** | 8.72 |
| l10-rand-0.75-1_160_MIP | MIP | 10 | rand | 1 | makespan | **39** | **39** | 7.73 |
| l10-rand-0.75-2_163_MIP | MIP | 10 | rand | 2 | makespan | **29** | **29** | 6.28 |
| l10-rand-0.75-3_158_MIP | MIP | 10 | rand | 3 | makespan | **31** | **31** | 5.97 |
| l10-rand-0.75-4_163_MIP | MIP | 10 | rand | 4 | makespan | **33** | **33** | 7.52 |
| l10-rand-0.75-5_167_MIP | MIP | 10 | rand | 5 | makespan | **37** | **37** | 8.78 |
| l10-rand-1-1_160_MIP | MIP | 10 | rand | 1 | makespan | **39** | **39** | 8.61 |
| l10-rand-1-2_163_MIP | MIP | 10 | rand | 2 | makespan | **29** | **29** | 5.97 |
| l10-rand-1-3_158_MIP | MIP | 10 | rand | 3 | makespan | **31** | **31** | 5.83 |
| l10-rand-1-4_163_MIP | MIP | 10 | rand | 4 | makespan | **33** | **33** | 7.16 |
| l10-rand-1-5_167_MIP | MIP | 10 | rand | 5 | makespan | **37** | **37** | 9.00 |

| Instance | Solver | Nodes | Demand | Id | Objective | Best objective | Best bound | Runtime (seconds) |
|---|---|---|---|---|---|---|---|---|
| l10-rw-0.25-1_160_MIP | MIP | 10 | rw | 1 | makespan | **39** | **39** | 7.94 |
| l10-rw-0.25-2_163_MIP | MIP | 10 | rw | 2 | makespan | **29** | **29** | 5.69 |
| l10-rw-0.25-3_158_MIP | MIP | 10 | rw | 3 | makespan | **31** | **31** | 6.27 |
| l10-rw-0.25-4_163_MIP | MIP | 10 | rw | 4 | makespan | **33** | **33** | 9.19 |
| l10-rw-0.25-5_167_MIP | MIP | 10 | rw | 5 | makespan | **37** | **37** | 8.48 |
| l10-rw-0.5-1_160_MIP | MIP | 10 | rw | 1 | makespan | **39** | **39** | 10.69 |
| l10-rw-0.5-2_163_MIP | MIP | 10 | rw | 2 | makespan | **29** | **29** | 7.17 |
| l10-rw-0.5-3_158_MIP | MIP | 10 | rw | 3 | makespan | **31** | **31** | 7.53 |
| l10-rw-0.5-4_163_MIP | MIP | 10 | rw | 4 | makespan | **33** | **33** | 7.63 |
| l10-rw-0.5-5_167_MIP | MIP | 10 | rw | 5 | makespan | **37** | **37** | 7.92 |
| l10-rw-0.75-1_160_MIP | MIP | 10 | rw | 1 | makespan | **39** | **39** | 8.45 |
| l10-rw-0.75-2_163_MIP | MIP | 10 | rw | 2 | makespan | **29** | **29** | 5.83 |
| l10-rw-0.75-3_158_MIP | MIP | 10 | rw | 3 | makespan | **31** | **31** | 7.13 |
| l10-rw-0.75-4_163_MIP | MIP | 10 | rw | 4 | makespan | **33** | **33** | 8.78 |
| l10-rw-0.75-5_167_MIP | MIP | 10 | rw | 5 | makespan | **37** | **37** | 8.14 |
| l10-rw-1-1_160_MIP | MIP | 10 | rw | 1 | makespan | **39** | **39** | 7.75 |
| l10-rw-1-2_163_MIP | MIP | 10 | rw | 2 | makespan | **29** | **29** | 6.02 |
| l10-rw-1-3_158_MIP | MIP | 10 | rw | 3 | makespan | **31** | **31** | 5.83 |
| l10-rw-1-4_163_MIP | MIP | 10 | rw | 4 | makespan | **33** | **33** | 8.30 |
| l10-rw-1-5_167_MIP | MIP | 10 | rw | 5 | makespan | **37** | **37** | 14.36 |
| l50-rand-0.25-1_768_MIP | MIP | 50 | rand | 1 | makespan | **72** | **72** | 211.91 |
| l50-rand-0.25-2_803_MIP | MIP | 50 | rand | 2 | makespan | **75** | **75** | 188.30 |
| l50-rand-0.25-3_803_MIP | MIP | 50 | rand | 3 | makespan | **70** | **70** | 188.55 |
| l50-rand-0.25-4_823_MIP | MIP | 50 | rand | 4 | makespan | **80** | **80** | 258.22 |
| l50-rand-0.25-5_787_MIP | MIP | 50 | rand | 5 | makespan | **74** | **74** | 195.38 |
| l50-rand-0.5-1_768_MIP | MIP | 50 | rand | 1 | makespan | **72** | **72** | 221.73 |
| l50-rand-0.5-2_803_MIP | MIP | 50 | rand | 2 | makespan | **75** | **75** | 241.75 |
| l50-rand-0.5-3_803_MIP | MIP | 50 | rand | 3 | makespan | **70** | **70** | 199.23 |
| l50-rand-0.5-4_823_MIP | MIP | 50 | rand | 4 | makespan | **80** | **80** | 238.17 |
| l50-rand-0.5-5_787_MIP | MIP | 50 | rand | 5 | makespan | **74** | **74** | 200.50 |
| l50-rand-0.75-1_768_MIP | MIP | 50 | rand | 1 | makespan | **72** | **72** | 265.77 |
| l50-rand-0.75-2_803_MIP | MIP | 50 | rand | 2 | makespan | **75** | **75** | 216.86 |
| l50-rand-0.75-3_803_MIP | MIP | 50 | rand | 3 | makespan | **70** | **70** | 185.91 |
| l50-rand-0.75-4_823_MIP | MIP | 50 | rand | 4 | makespan | **80** | **80** | 277.17 |
| l50-rand-0.75-5_787_MIP | MIP | 50 | rand | 5 | makespan | **74** | **74** | 196.81 |
| l50-rand-1-1_768_MIP | MIP | 50 | rand | 1 | makespan | **72** | **72** | 174.03 |
| l50-rand-1-2_803_MIP | MIP | 50 | rand | 2 | makespan | **75** | **75** | 210.31 |
| l50-rand-1-3_803_MIP | MIP | 50 | rand | 3 | makespan | **70** | **70** | 185.22 |
| l50-rand-1-4_823_MIP | MIP | 50 | rand | 4 | makespan | **80** | **80** | 263.91 |
| l50-rand-1-5_787_MIP | MIP | 50 | rand | 5 | makespan | **74** | **74** | 266.52 |
| l50-rw-0.25-1_768_MIP | MIP | 50 | rw | 1 | makespan | | | T |
| l50-rw-0.25-2_803_MIP | MIP | 50 | rw | 2 | makespan | | | T |
| l50-rw-0.25-3_803_MIP | MIP | 50 | rw | 3 | makespan | | | T |
| l50-rw-0.25-4_823_MIP | MIP | 50 | rw | 4 | makespan | **80** | **80** | 2,609.80 |
| l50-rw-0.25-5_787_MIP | MIP | 50 | rw | 5 | makespan | | | T |
| l50-rw-0.5-1_768_MIP | MIP | 50 | rw | 1 | makespan | **72** | **72** | 2,199.22 |
| l50-rw-0.5-2_803_MIP | MIP | 50 | rw | 2 | makespan | | | T |
| l50-rw-0.5-3_803_MIP | MIP | 50 | rw | 3 | makespan | **70** | **70** | T |
| l50-rw-0.5-4_823_MIP | MIP | 50 | rw | 4 | makespan | **80** | **80** | 3,388.91 |
| l50-rw-0.5-5_787_MIP | MIP | 50 | rw | 5 | makespan | **74** | **74** | 3,280.91 |
| l50-rw-0.75-1_768_MIP | MIP | 50 | rw | 1 | makespan | **72** | **72** | 3,538.88 |
| l50-rw-0.75-2_803_MIP | MIP | 50 | rw | 2 | makespan | | | T |
| l50-rw-0.75-3_803_MIP | MIP | 50 | rw | 3 | makespan | | | T |
| l50-rw-0.75-4_823_MIP | MIP | 50 | rw | 4 | makespan | | | T |
| l50-rw-0.75-5_787_MIP | MIP | 50 | rw | 5 | makespan | **74** | **74** | 3,011.94 |
| l50-rw-1-1_768_MIP | MIP | 50 | rw | 1 | makespan | **72** | **72** | 3,394.28 |
| l50-rw-1-2_803_MIP | MIP | 50 | rw | 2 | makespan | | | T |
| l50-rw-1-3_803_MIP | MIP | 50 | rw | 3 | makespan | | | T |
| l50-rw-1-4_823_MIP | MIP | 50 | rw | 4 | makespan | | | T |
| l50-rw-1-5_787_MIP | MIP | 50 | rw | 5 | makespan | | | T |
| l100-rand-0.25-1_1557_MIP | MIP | 100 | rand | 1 | makespan | **125** | **125** | T |
| l100-rand-0.25-2_1605_MIP | MIP | 100 | rand | 2 | makespan | **121** | **121** | T |
| l100-rand-0.25-3_1580_MIP | MIP | 100 | rand | 3 | makespan | **128** | **128** | 2,600.80 |
| l100-rand-0.25-4_1649_MIP | MIP | 100 | rand | 4 | makespan | | | 3,621.78 |
| l100-rand-0.25-5_1573_MIP | MIP | 100 | rand | 5 | makespan | **122** | **122** | T |
| l100-rand-0.5-1_1557_MIP | MIP | 100 | rand | 1 | makespan | **125** | **125** | 2,228.09 |
| l100-rand-0.5-2_1605_MIP | MIP | 100 | rand | 2 | makespan | **121** | **121** | 2,472.50 |
| l100-rand-0.5-3_1580_MIP | MIP | 100 | rand | 3 | makespan | | | T |
| l100-rand-0.5-4_1649_MIP | MIP | 100 | rand | 4 | makespan | **129** | **129** | 2,667.38 |
| l100-rand-0.5-5_1573_MIP | MIP | 100 | rand | 5 | makespan | **122** | **122** | T |
| l100-rand-0.75-1_1557_MIP | MIP | 100 | rand | 1 | makespan | | | T |
| l100-rand-0.75-2_1605_MIP | MIP | 100 | rand | 2 | makespan | **121** | **121** | 2,357.38 |
| l100-rand-0.75-3_1580_MIP | MIP | 100 | rand | 3 | makespan | | | T |
| l100-rand-0.75-4_1649_MIP | MIP | 100 | rand | 4 | makespan | | | T |
| l100-rand-0.75-5_1573_MIP | MIP | 100 | rand | 5 | makespan | **122** | **122** | 1,978.94 |
| l100-rand-1-1_1557_MIP | MIP | 100 | rand | 1 | makespan | **125** | **125** | T |

| Instance | Solver | Nodes | Demand | Id | Objective | Best objective | Best bound | Runtime (seconds) |
|---|---|---|---|---|---|---|---|---|
| l100-rand-1-2_1605_MIP | MIP | 100 | rand | 2 | makespan | | | T |
| l100-rand-1-3_1580_MIP | MIP | 100 | rand | 3 | makespan | | | T |
| l100-rand-1-4_1649_MIP | MIP | 100 | rand | 4 | makespan | **129** | **129** | 2,474.47 |
| l100-rand-1-5_1573_MIP | MIP | 100 | rand | 5 | makespan | **122** | **122** | 1,799.42 |
| l100-rw-0.25-1_1557_MIP | MIP | 100 | rw | 1 | makespan | | | T |
| l100-rw-0.25-2_1605_MIP | MIP | 100 | rw | 2 | makespan | | | T |
| l100-rw-0.25-3_1580_MIP | MIP | 100 | rw | 3 | makespan | | | T |
| l100-rw-0.25-4_1649_MIP | MIP | 100 | rw | 4 | makespan | | | T |
| l100-rw-0.25-5_1573_MIP | MIP | 100 | rw | 5 | makespan | | | T |
| l100-rw-0.5-1_1557_MIP | MIP | 100 | rw | 1 | makespan | | | T |
| l100-rw-0.5-2_1605_MIP | MIP | 100 | rw | 2 | makespan | | | T |
| l100-rw-0.5-3_1580_MIP | MIP | 100 | rw | 3 | makespan | | | T |
| l100-rw-0.5-4_1649_MIP | MIP | 100 | rw | 4 | makespan | | | T |
| l100-rw-0.5-5_1573_MIP | MIP | 100 | rw | 5 | makespan | | | T |
| l100-rw-0.75-1_1557_MIP | MIP | 100 | rw | 1 | makespan | | | T |
| l100-rw-0.75-2_1605_MIP | MIP | 100 | rw | 2 | makespan | | | T |
| l100-rw-0.75-3_1580_MIP | MIP | 100 | rw | 3 | makespan | | | T |
| l100-rw-0.75-4_1649_MIP | MIP | 100 | rw | 4 | makespan | | | T |
| l100-rw-0.75-5_1573_MIP | MIP | 100 | rw | 5 | makespan | | | T |
| l100-rw-1-1_1557_MIP | MIP | 100 | rw | 1 | makespan | | | T |
| l100-rw-1-2_1605_MIP | MIP | 100 | rw | 2 | makespan | | | T |
| l100-rw-1-3_1580_MIP | MIP | 100 | rw | 3 | makespan | | | T |
| l100-rw-1-4_1649_MIP | MIP | 100 | rw | 4 | makespan | | | T |
| l100-rw-1-5_1573_MIP | MIP | 100 | rw | 5 | makespan | | | T |
| l10-rand-0.25-1_160_MIP | MIP | 10 | rand | 1 | timebalance | **4** | **4** | 14.47 |
| l10-rand-0.25-2_163_MIP | MIP | 10 | rand | 2 | timebalance | **2** | **2** | 6.97 |
| l10-rand-0.25-3_158_MIP | MIP | 10 | rand | 3 | timebalance | **3** | **3** | 10.19 |
| l10-rand-0.25-4_163_MIP | MIP | 10 | rand | 4 | timebalance | **3** | **3** | 9.03 |
| l10-rand-0.25-5_167_MIP | MIP | 10 | rand | 5 | timebalance | **3** | **3** | 11.72 |
| l10-rand-0.5-1_160_MIP | MIP | 10 | rand | 1 | timebalance | **4** | **4** | 14.41 |
| l10-rand-0.5-2_163_MIP | MIP | 10 | rand | 2 | timebalance | **2** | **2** | 7.06 |
| l10-rand-0.5-3_158_MIP | MIP | 10 | rand | 3 | timebalance | **3** | **3** | 10.84 |
| l10-rand-0.5-4_163_MIP | MIP | 10 | rand | 4 | timebalance | **3** | **3** | 9.69 |
| l10-rand-0.5-5_167_MIP | MIP | 10 | rand | 5 | timebalance | **3** | **3** | 16.92 |
| l10-rand-0.75-1_160_MIP | MIP | 10 | rand | 1 | timebalance | **4** | **4** | 14.31 |
| l10-rand-0.75-2_163_MIP | MIP | 10 | rand | 2 | timebalance | **2** | **2** | 8.63 |
| l10-rand-0.75-3_158_MIP | MIP | 10 | rand | 3 | timebalance | **3** | **3** | 9.13 |
| l10-rand-0.75-4_163_MIP | MIP | 10 | rand | 4 | timebalance | **3** | **3** | 10.73 |
| l10-rand-0.75-5_167_MIP | MIP | 10 | rand | 5 | timebalance | **3** | **3** | 13.98 |
| l10-rand-1-1_160_MIP | MIP | 10 | rand | 1 | timebalance | **4** | **4** | 18.86 |
| l10-rand-1-2_163_MIP | MIP | 10 | rand | 2 | timebalance | **2** | **2** | 7.13 |
| l10-rand-1-3_158_MIP | MIP | 10 | rand | 3 | timebalance | **3** | **3** | 10.13 |
| l10-rand-1-4_163_MIP | MIP | 10 | rand | 4 | timebalance | **3** | **3** | 9.47 |
| l10-rand-1-5_167_MIP | MIP | 10 | rand | 5 | timebalance | **3** | **3** | 12.56 |
| l10-rw-0.25-1_160_MIP | MIP | 10 | rw | 1 | timebalance | **4** | **4** | 16.84 |
| l10-rw-0.25-2_163_MIP | MIP | 10 | rw | 2 | timebalance | **2** | **2** | 13.22 |
| l10-rw-0.25-3_158_MIP | MIP | 10 | rw | 3 | timebalance | **3** | **3** | 12.94 |
| l10-rw-0.25-4_163_MIP | MIP | 10 | rw | 4 | timebalance | **3** | **3** | 13.34 |
| l10-rw-0.25-5_167_MIP | MIP | 10 | rw | 5 | timebalance | **3** | **3** | 35.59 |
| l10-rw-0.5-1_160_MIP | MIP | 10 | rw | 1 | timebalance | **4** | **4** | 18.16 |
| l10-rw-0.5-2_163_MIP | MIP | 10 | rw | 2 | timebalance | **2** | **2** | 7.78 |
| l10-rw-0.5-3_158_MIP | MIP | 10 | rw | 3 | timebalance | **3** | **3** | 25.26 |
| l10-rw-0.5-4_163_MIP | MIP | 10 | rw | 4 | timebalance | **3** | **3** | 15.00 |
| l10-rw-0.5-5_167_MIP | MIP | 10 | rw | 5 | timebalance | **3** | **3** | 23.31 |
| l10-rw-0.75-1_160_MIP | MIP | 10 | rw | 1 | timebalance | **4** | **4** | 28.88 |
| l10-rw-0.75-2_163_MIP | MIP | 10 | rw | 2 | timebalance | **2** | **2** | 10.92 |
| l10-rw-0.75-3_158_MIP | MIP | 10 | rw | 3 | timebalance | **3** | **3** | 19.73 |
| l10-rw-0.75-4_163_MIP | MIP | 10 | rw | 4 | timebalance | **3** | **3** | 15.17 |
| l10-rw-0.75-5_167_MIP | MIP | 10 | rw | 5 | timebalance | **3** | **3** | 38.52 |
| l10-rw-1-1_160_MIP | MIP | 10 | rw | 1 | timebalance | **4** | **4** | 25.84 |
| l10-rw-1-2_163_MIP | MIP | 10 | rw | 2 | timebalance | **2** | **2** | 7.17 |
| l10-rw-1-3_158_MIP | MIP | 10 | rw | 3 | timebalance | **3** | **3** | 15.92 |
| l10-rw-1-4_163_MIP | MIP | 10 | rw | 4 | timebalance | **3** | **3** | 12.88 |
| l10-rw-1-5_167_MIP | MIP | 10 | rw | 5 | timebalance | **3** | **3** | 30.19 |
| l50-rand-0.25-1_768_MIP | MIP | 50 | rand | 1 | timebalance | **4** | **4** | 316.91 |
| l50-rand-0.25-2_803_MIP | MIP | 50 | rand | 2 | timebalance | **4** | **4** | 447.63 |
| l50-rand-0.25-3_803_MIP | MIP | 50 | rand | 3 | timebalance | **4** | **4** | 441.19 |
| l50-rand-0.25-4_823_MIP | MIP | 50 | rand | 4 | timebalance | **4** | **4** | 526.59 |
| l50-rand-0.25-5_787_MIP | MIP | 50 | rand | 5 | timebalance | **4** | **4** | 664.83 |
| l50-rand-0.5-1_768_MIP | MIP | 50 | rand | 1 | timebalance | **4** | **4** | 328.92 |
| l50-rand-0.5-2_803_MIP | MIP | 50 | rand | 2 | timebalance | **4** | **4** | 380.59 |
| l50-rand-0.5-3_803_MIP | MIP | 50 | rand | 3 | timebalance | **4** | **4** | 369.91 |
| l50-rand-0.5-4_823_MIP | MIP | 50 | rand | 4 | timebalance | **4** | **4** | 489.88 |
| l50-rand-0.5-5_787_MIP | MIP | 50 | rand | 5 | timebalance | **4** | **4** | 453.72 |
| l50-rand-0.75-1_768_MIP | MIP | 50 | rand | 1 | timebalance | **4** | **4** | 277.88 |
| l50-rand-0.75-2_803_MIP | MIP | 50 | rand | 2 | timebalance | **4** | **4** | 485.27 |

| Instance | Solver | Nodes | Demand | Id | Objective | Best objective | Best bound | Runtime (seconds) |
|---|---|---|---|---|---|---|---|---|
| l50-rand-0.75-3_803_MIP | MIP | 50 | rand | 3 | timebalance | **4** | **4** | 379.67 |
| l50-rand-0.75-4_823_MIP | MIP | 50 | rand | 4 | timebalance | **4** | **4** | 452.89 |
| l50-rand-0.75-5_787_MIP | MIP | 50 | rand | 5 | timebalance | **4** | **4** | 373.25 |
| l50-rand-1-1_768_MIP | MIP | 50 | rand | 1 | timebalance | **4** | **4** | 313.19 |
| l50-rand-1-2_803_MIP | MIP | 50 | rand | 2 | timebalance | **4** | **4** | 344.67 |
| l50-rand-1-3_803_MIP | MIP | 50 | rand | 3 | timebalance | **4** | **4** | 377.75 |
| l50-rand-1-4_823_MIP | MIP | 50 | rand | 4 | timebalance | **4** | **4** | 434.63 |
| l50-rand-1-5_787_MIP | MIP | 50 | rand | 5 | timebalance | **4** | **4** | 591.33 |
| l50-rw-0.25-1_768_MIP | MIP | 50 | rw | 1 | timebalance | | | T |
| l50-rw-0.25-2_803_MIP | MIP | 50 | rw | 2 | timebalance | | | T |
| l50-rw-0.25-3_803_MIP | MIP | 50 | rw | 3 | timebalance | 27 | 2.176212308 | T |
| l50-rw-0.25-4_823_MIP | MIP | 50 | rw | 4 | timebalance | | | T |
| l50-rw-0.25-5_787_MIP | MIP | 50 | rw | 5 | timebalance | 20 | 3 | T |
| l50-rw-0.5-1_768_MIP | MIP | 50 | rw | 1 | timebalance | | | T |
| l50-rw-0.5-2_803_MIP | MIP | 50 | rw | 2 | timebalance | | | T |
| l50-rw-0.5-3_803_MIP | MIP | 50 | rw | 3 | timebalance | 27 | 2.202663552 | T |
| l50-rw-0.5-4_823_MIP | MIP | 50 | rw | 4 | timebalance | 26 | 3.239925626 | T |
| l50-rw-0.5-5_787_MIP | MIP | 50 | rw | 5 | timebalance | 22 | 3 | T |
| l50-rw-0.75-1_768_MIP | MIP | 50 | rw | 1 | timebalance | | | T |
| l50-rw-0.75-2_803_MIP | MIP | 50 | rw | 2 | timebalance | | | T |
| l50-rw-0.75-3_803_MIP | MIP | 50 | rw | 3 | timebalance | | | T |
| l50-rw-0.75-4_823_MIP | MIP | 50 | rw | 4 | timebalance | 28 | 3.157533472 | T |
| l50-rw-0.75-5_787_MIP | MIP | 50 | rw | 5 | timebalance | | | T |
| l50-rw-1-1_768_MIP | MIP | 50 | rw | 1 | timebalance | 26 | 3.415547878 | T |
| l50-rw-1-2_803_MIP | MIP | 50 | rw | 2 | timebalance | | | T |
| l50-rw-1-3_803_MIP | MIP | 50 | rw | 3 | timebalance | 27 | 2.531647880 | T |
| l50-rw-1-4_823_MIP | MIP | 50 | rw | 4 | timebalance | | | T |
| l50-rw-1-5_787_MIP | MIP | 50 | rw | 5 | timebalance | 27 | 4 | T |
| l100-rand-0.25-1_1557_MIP | MIP | 100 | rand | 1 | timebalance | | | T |
| l100-rand-0.25-2_1605_MIP | MIP | 100 | rand | 2 | timebalance | **11** | **4** | T |
| l100-rand-0.25-3_1580_MIP | MIP | 100 | rand | 3 | timebalance | | | T |
| l100-rand-0.25-4_1649_MIP | MIP | 100 | rand | 4 | timebalance | | | T |
| l100-rand-0.25-5_1573_MIP | MIP | 100 | rand | 5 | timebalance | | | T |
| l100-rand-0.5-1_1557_MIP | MIP | 100 | rand | 1 | timebalance | | | T |
| l100-rand-0.5-2_1605_MIP | MIP | 100 | rand | 2 | timebalance | | | T |
| l100-rand-0.5-3_1580_MIP | MIP | 100 | rand | 3 | timebalance | 18 | 3.666666667 | T |
| l100-rand-0.5-4_1649_MIP | MIP | 100 | rand | 4 | timebalance | | | T |
| l100-rand-0.5-5_1573_MIP | MIP | 100 | rand | 5 | timebalance | 15 | 5 | T |
| l100-rand-0.75-1_1557_MIP | MIP | 100 | rand | 1 | timebalance | **4** | **4** | 3,318.61 |
| l100-rand-0.75-2_1605_MIP | MIP | 100 | rand | 2 | timebalance | **4** | **4** | 3,483.53 |
| l100-rand-0.75-3_1580_MIP | MIP | 100 | rand | 3 | timebalance | **4** | **4** | 3,548.84 |
| l100-rand-0.75-4_1649_MIP | MIP | 100 | rand | 4 | timebalance | 12 | 5 | T |
| l100-rand-0.75-5_1573_MIP | MIP | 100 | rand | 5 | timebalance | 20 | 5 | T |
| l100-rand-1-1_1557_MIP | MIP | 100 | rand | 1 | timebalance | | | T |
| l100-rand-1-2_1605_MIP | MIP | 100 | rand | 2 | timebalance | 16 | 4 | T |
| l100-rand-1-3_1580_MIP | MIP | 100 | rand | 3 | timebalance | | | T |
| l100-rand-1-4_1649_MIP | MIP | 100 | rand | 4 | timebalance | | | T |
| l100-rand-1-5_1573_MIP | MIP | 100 | rand | 5 | timebalance | **5** | **5** | 2,490.03 |
| l100-rw-0.25-1_1557_MIP | MIP | 100 | rw | 1 | timebalance | | | T |
| l100-rw-0.25-2_1605_MIP | MIP | 100 | rw | 2 | timebalance | | | T |
| l100-rw-0.25-3_1580_MIP | MIP | 100 | rw | 3 | timebalance | | | T |
| l100-rw-0.25-4_1649_MIP | MIP | 100 | rw | 4 | timebalance | | | T |
| l100-rw-0.25-5_1573_MIP | MIP | 100 | rw | 5 | timebalance | | | T |
| l100-rw-0.5-1_1557_MIP | MIP | 100 | rw | 1 | timebalance | | | T |
| l100-rw-0.5-2_1605_MIP | MIP | 100 | rw | 2 | timebalance | | | T |
| l100-rw-0.5-3_1580_MIP | MIP | 100 | rw | 3 | timebalance | | | T |
| l100-rw-0.5-4_1649_MIP | MIP | 100 | rw | 4 | timebalance | | | T |
| l100-rw-0.5-5_1573_MIP | MIP | 100 | rw | 5 | timebalance | | | T |
| l100-rw-0.75-1_1557_MIP | MIP | 100 | rw | 1 | timebalance | | | T |
| l100-rw-0.75-2_1605_MIP | MIP | 100 | rw | 2 | timebalance | | | T |
| l100-rw-0.75-3_1580_MIP | MIP | 100 | rw | 3 | timebalance | | | T |
| l100-rw-0.75-4_1649_MIP | MIP | 100 | rw | 4 | timebalance | | | T |
| l100-rw-0.75-5_1573_MIP | MIP | 100 | rw | 5 | timebalance | | | T |
| l100-rw-1-1_1557_MIP | MIP | 100 | rw | 1 | timebalance | | | T |
| l100-rw-1-2_1605_MIP | MIP | 100 | rw | 2 | timebalance | | | T |
| l100-rw-1-3_1580_MIP | MIP | 100 | rw | 3 | timebalance | | | T |
| l100-rw-1-4_1649_MIP | MIP | 100 | rw | 4 | timebalance | | | T |
| l100-rw-1-5_1573_MIP | MIP | 100 | rw | 5 | timebalance | | | T |
| l10-rand-0.25-1_160_MIP | MIP | 10 | rand | 1 | loadbalance | **0.571428571** | **0.571428571** | 348.33 |
| l10-rand-0.25-2_163_MIP | MIP | 10 | rand | 2 | loadbalance | 0.614285714 | 0.614224548 | 1,783.77 |
| l10-rand-0.25-3_158_MIP | MIP | 10 | rand | 3 | loadbalance | 0.633802816 | 0.633766234 | 345.61 |
| l10-rand-0.25-4_163_MIP | MIP | 10 | rand | 4 | loadbalance | **0.602739726** | **0.602739726** | 465.05 |
| l10-rand-0.25-5_167_MIP | MIP | 10 | rand | 5 | loadbalance | 0.597402597 | 0.597343533 | 1,627.02 |
| l10-rand-0.5-1_160_MIP | MIP | 10 | rand | 1 | loadbalance | **0.415094340** | **0.415094340** | 216.91 |
| l10-rand-0.5-2_163_MIP | MIP | 10 | rand | 2 | loadbalance | **0.452631579** | **0.452631579** | 451.09 |
| l10-rand-0.5-3_158_MIP | MIP | 10 | rand | 3 | loadbalance | 0.463917526 | 0.463874346 | 612.44 |

| Instance | Solver | Nodes | Demand | Id | Objective | Best objective | Best bound | Runtime (seconds) |
|---|---|---|---|---|---|---|---|---|
| l10-rand-0.5-4_163_MIP | MIP | 10 | rand | 4 | loadbalance | 0.438775509 | 0.438731857 | 710.13 |
| l10-rand-0.5-5_167_MIP | MIP | 10 | rand | 5 | loadbalance | 0.425925926 | 0.425883568 | 1,307.42 |
| l10-rand-0.75-1_160_MIP | MIP | 10 | rand | 1 | loadbalance | **0.323529412** | **0.323529412** | 206.56 |
| l10-rand-0.75-2_163_MIP | MIP | 10 | rand | 2 | loadbalance | **0.358333333** | **0.358333333** | 470.44 |
| l10-rand-0.75-3_158_MIP | MIP | 10 | rand | 3 | loadbalance | 0.366666667 | 0.366643321 | 196.05 |
| l10-rand-0.75-4_163_MIP | MIP | 10 | rand | 4 | loadbalance | **0.343750000** | **0.343750000** | 703.00 |
| l10-rand-0.75-5_167_MIP | MIP | 10 | rand | 5 | loadbalance | 0.330935252 | 0.330919296 | 1,499.09 |
| l10-rand-1-1_160_MIP | MIP | 10 | rand | 1 | loadbalance | **0.265060241** | **0.265060241** | 471.81 |
| l10-rand-1-2_163_MIP | MIP | 10 | rand | 2 | loadbalance | 0.296551724 | 0.296549919 | 793.09 |
| l10-rand-1-3_158_MIP | MIP | 10 | rand | 3 | loadbalance | **0.303448276** | **0.303448276** | 287.36 |
| l10-rand-1-4_163_MIP | MIP | 10 | rand | 4 | loadbalance | **0.283870968** | **0.283870968** | 420.34 |
| l10-rand-1-5_167_MIP | MIP | 10 | rand | 5 | loadbalance | 0.283950617 | 0.265930870 | T |
| l10-rw-0.25-1_160_MIP | MIP | 10 | rw | 1 | loadbalance | **0.333333333** | **0.333333333** | 66.06 |
| l10-rw-0.25-2_163_MIP | MIP | 10 | rw | 2 | loadbalance | **0.333333333** | **0.333333333** | 909.58 |
| l10-rw-0.25-3_158_MIP | MIP | 10 | rw | 3 | loadbalance | **0.333333333** | **0.333333333** | 236.63 |
| l10-rw-0.25-4_163_MIP | MIP | 10 | rw | 4 | loadbalance | **0.500000000** | **0.500000000** | 10.73 |
| l10-rw-0.25-5_167_MIP | MIP | 10 | rw | 5 | loadbalance | **0.333333333** | **0.333333333** | 36.23 |
| l10-rw-0.5-1_160_MIP | MIP | 10 | rw | 1 | loadbalance | **0.285714286** | **0.285714286** | 574.05 |
| l10-rw-0.5-2_163_MIP | MIP | 10 | rw | 2 | loadbalance | **0.285714286** | **0.285714286** | 414.41 |
| l10-rw-0.5-3_158_MIP | MIP | 10 | rw | 3 | loadbalance | 0.333333332 | 0.253145849 | T |
| l10-rw-0.5-4_163_MIP | MIP | 10 | rw | 4 | loadbalance | **0.428571429** | **0.428571429** | 12.23 |
| l10-rw-0.5-5_167_MIP | MIP | 10 | rw | 5 | loadbalance | **0.285714286** | **0.285714286** | 118.80 |
| l10-rw-0.75-1_160_MIP | MIP | 10 | rw | 1 | loadbalance | **0.250000000** | **0.250000000** | 171.38 |
| l10-rw-0.75-2_163_MIP | MIP | 10 | rw | 2 | loadbalance | **0.250000000** | **0.250000000** | 440.59 |
| l10-rw-0.75-3_158_MIP | MIP | 10 | rw | 3 | loadbalance | 0.333333332 | 0.250000000 | T |
| l10-rw-0.75-4_163_MIP | MIP | 10 | rw | 4 | loadbalance | **0.375000000** | **0.375000000** | 15.63 |
| l10-rw-0.75-5_167_MIP | MIP | 10 | rw | 5 | loadbalance | **0.250000000** | **0.250000000** | 142.28 |
| l10-rw-1-1_160_MIP | MIP | 10 | rw | 1 | loadbalance | **0.250000000** | **0.250000000** | 1,713.78 |
| l10-rw-1-2_163_MIP | MIP | 10 | rw | 2 | loadbalance | 0.333333332 | 0.229477187 | T |
| l10-rw-1-3_158_MIP | MIP | 10 | rw | 3 | loadbalance | 0.333333332 | 0.222222222 | T |
| l10-rw-1-4_163_MIP | MIP | 10 | rw | 4 | loadbalance | **0.333333333** | **0.333333333** | 12.39 |
| l10-rw-1-5_167_MIP | MIP | 10 | rw | 5 | loadbalance | **0.250000000** | **0.250000000** | 1,937.67 |
| l50-rand-0.25-1_768_MIP | MIP | 50 | rand | 1 | loadbalance | 0.834394904 | 0.471791014 | T |
| l50-rand-0.25-2_803_MIP | MIP | 50 | rand | 2 | loadbalance | 0.728915663 | 0.470106371 | T |
| l50-rand-0.25-3_803_MIP | MIP | 50 | rand | 3 | loadbalance | 0.760479042 | 0.490834855 | T |
| l50-rand-0.25-4_823_MIP | MIP | 50 | rand | 4 | loadbalance | 0.668639053 | 0.435022617 | T |
| l50-rand-0.25-5_787_MIP | MIP | 50 | rand | 5 | loadbalance | 0.676829268 | 0.480128353 | T |
| l50-rand-0.5-1_768_MIP | MIP | 50 | rand | 1 | loadbalance | 0.494565216 | 0.397511769 | T |
| l50-rand-0.5-2_803_MIP | MIP | 50 | rand | 2 | loadbalance | 0.566326531 | 0.394969311 | T |
| l50-rand-0.5-3_803_MIP | MIP | 50 | rand | 3 | loadbalance | 0.678571429 | 0.412551739 | T |
| l50-rand-0.5-4_823_MIP | MIP | 50 | rand | 4 | loadbalance | 0.534653465 | 0.369292271 | T |
| l50-rand-0.5-5_787_MIP | MIP | 50 | rand | 5 | loadbalance | 0.557213930 | 0.403976181 | T |
| l50-rand-0.75-1_768_MIP | MIP | 50 | rand | 1 | loadbalance | 0.544600939 | 0.342940348 | T |
| l50-rand-0.75-2_803_MIP | MIP | 50 | rand | 2 | loadbalance | 0.551111111 | 0.343028677 | T |
| l50-rand-0.75-3_803_MIP | MIP | 50 | rand | 3 | loadbalance | 0.536796537 | 0.354972202 | T |
| l50-rand-0.75-4_823_MIP | MIP | 50 | rand | 4 | loadbalance | 0.463302752 | 0.318438736 | T |
| l50-rand-0.75-5_787_MIP | MIP | 50 | rand | 5 | loadbalance | 0.529147982 | 0.348461875 | T |
| l50-rand-1-1_768_MIP | MIP | 50 | rand | 1 | loadbalance | 0.578512397 | 0.300623027 | T |
| l50-rand-1-2_803_MIP | MIP | 50 | rand | 2 | loadbalance | 0.477443609 | 0.302171606 | T |
| l50-rand-1-3_803_MIP | MIP | 50 | rand | 3 | loadbalance | 0.498084291 | 0.313358764 | T |
| l50-rand-1-4_823_MIP | MIP | 50 | rand | 4 | loadbalance | 0.426294821 | 0.281793250 | T |
| l50-rand-1-5_787_MIP | MIP | 50 | rand | 5 | loadbalance | 0.484962406 | 0.308868437 | T |
| l50-rw-0.25-1_768_MIP | MIP | 50 | rw | 1 | loadbalance | | | T |
| l50-rw-0.25-2_803_MIP | MIP | 50 | rw | 2 | loadbalance | | | T |
| l50-rw-0.25-3_803_MIP | MIP | 50 | rw | 3 | loadbalance | | | T |
| l50-rw-0.25-4_823_MIP | MIP | 50 | rw | 4 | loadbalance | | | T |
| l50-rw-0.25-5_787_MIP | MIP | 50 | rw | 5 | loadbalance | | | T |
| l50-rw-0.5-1_768_MIP | MIP | 50 | rw | 1 | loadbalance | 0.437500000 | 0.174603175 | T |
| l50-rw-0.5-2_803_MIP | MIP | 50 | rw | 2 | loadbalance | | | T |
| l50-rw-0.5-3_803_MIP | MIP | 50 | rw | 3 | loadbalance | | | T |
| l50-rw-0.5-4_823_MIP | MIP | 50 | rw | 4 | loadbalance | | | T |
| l50-rw-0.5-5_787_MIP | MIP | 50 | rw | 5 | loadbalance | | | T |
| l50-rw-0.75-1_768_MIP | MIP | 50 | rw | 1 | loadbalance | | | T |
| l50-rw-0.75-2_803_MIP | MIP | 50 | rw | 2 | loadbalance | | | T |
| l50-rw-0.75-3_803_MIP | MIP | 50 | rw | 3 | loadbalance | | | T |
| l50-rw-0.75-4_823_MIP | MIP | 50 | rw | 4 | loadbalance | | | T |
| l50-rw-0.75-5_787_MIP | MIP | 50 | rw | 5 | loadbalance | | | T |
| l50-rw-1-1_768_MIP | MIP | 50 | rw | 1 | loadbalance | | | T |
| l50-rw-1-2_803_MIP | MIP | 50 | rw | 2 | loadbalance | | | T |
| l50-rw-1-3_803_MIP | MIP | 50 | rw | 3 | loadbalance | | | T |
| l50-rw-1-4_823_MIP | MIP | 50 | rw | 4 | loadbalance | | | T |
| l50-rw-1-5_787_MIP | MIP | 50 | rw | 5 | loadbalance | | | T |
| l100-rand-0.25-1_1557_MIP | MIP | 100 | rand | 1 | loadbalance | | | T |
| l100-rand-0.25-2_1605_MIP | MIP | 100 | rand | 2 | loadbalance | | | T |
| l100-rand-0.25-3_1580_MIP | MIP | 100 | rand | 3 | loadbalance | | | T |
| l100-rand-0.25-4_1649_MIP | MIP | 100 | rand | 4 | loadbalance | | | T |

| Instance | Solver | Nodes | Demand | Id | Objective | Best objective | Best bound | Runtime (seconds) |
|---|---|---|---|---|---|---|---|---|
| l100-rand-0.25-5_1573_MIP | MIP | 100 | rand | 5 | loadbalance | | | T |
| l100-rand-0.5-1_1557_MIP | MIP | 100 | rand | 1 | loadbalance | | | T |
| l100-rand-0.5-2_1605_MIP | MIP | 100 | rand | 2 | loadbalance | | | T |
| l100-rand-0.5-3_1580_MIP | MIP | 100 | rand | 3 | loadbalance | | | T |
| l100-rand-0.5-4_1649_MIP | MIP | 100 | rand | 4 | loadbalance | | | T |
| l100-rand-0.5-5_1573_MIP | MIP | 100 | rand | 5 | loadbalance | | | T |
| l100-rand-0.75-1_1557_MIP | MIP | 100 | rand | 1 | loadbalance | | | T |
| l100-rand-0.75-2_1605_MIP | MIP | 100 | rand | 2 | loadbalance | | | T |
| l100-rand-0.75-3_1580_MIP | MIP | 100 | rand | 3 | loadbalance | | | T |
| l100-rand-0.75-4_1649_MIP | MIP | 100 | rand | 4 | loadbalance | | | T |
| l100-rand-0.75-5_1573_MIP | MIP | 100 | rand | 5 | loadbalance | | | T |
| l100-rand-1-1_1557_MIP | MIP | 100 | rand | 1 | loadbalance | | | T |
| l100-rand-1-2_1605_MIP | MIP | 100 | rand | 2 | loadbalance | | | T |
| l100-rand-1-3_1580_MIP | MIP | 100 | rand | 3 | loadbalance | | | T |
| l100-rand-1-4_1649_MIP | MIP | 100 | rand | 4 | loadbalance | | | T |
| l100-rand-1-5_1573_MIP | MIP | 100 | rand | 5 | loadbalance | | | T |
| l100-rw-0.25-1_1557_MIP | MIP | 100 | rw | 1 | loadbalance | | | T |
| l100-rw-0.25-2_1605_MIP | MIP | 100 | rw | 2 | loadbalance | | | T |
| l100-rw-0.25-3_1580_MIP | MIP | 100 | rw | 3 | loadbalance | | | T |
| l100-rw-0.25-4_1649_MIP | MIP | 100 | rw | 4 | loadbalance | | | T |
| l100-rw-0.25-5_1573_MIP | MIP | 100 | rw | 5 | loadbalance | | | T |
| l100-rw-0.5-1_1557_MIP | MIP | 100 | rw | 1 | loadbalance | | | T |
| l100-rw-0.5-2_1605_MIP | MIP | 100 | rw | 2 | loadbalance | | | T |
| l100-rw-0.5-3_1580_MIP | MIP | 100 | rw | 3 | loadbalance | | | T |
| l100-rw-0.5-4_1649_MIP | MIP | 100 | rw | 4 | loadbalance | | | T |
| l100-rw-0.5-5_1573_MIP | MIP | 100 | rw | 5 | loadbalance | | | T |
| l100-rw-0.75-1_1557_MIP | MIP | 100 | rw | 1 | loadbalance | | | T |
| l100-rw-0.75-2_1605_MIP | MIP | 100 | rw | 2 | loadbalance | | | T |
| l100-rw-0.75-3_1580_MIP | MIP | 100 | rw | 3 | loadbalance | | | T |
| l100-rw-0.75-4_1649_MIP | MIP | 100 | rw | 4 | loadbalance | | | T |
| l100-rw-0.75-5_1573_MIP | MIP | 100 | rw | 5 | loadbalance | | | T |
| l100-rw-1-1_1557_MIP | MIP | 100 | rw | 1 | loadbalance | | | T |
| l100-rw-1-2_1605_MIP | MIP | 100 | rw | 2 | loadbalance | | | T |
| l100-rw-1-3_1580_MIP | MIP | 100 | rw | 3 | loadbalance | | | T |
| l100-rw-1-4_1649_MIP | MIP | 100 | rw | 4 | loadbalance | | | T |
| l100-rw-1-5_1573_MIP | MIP | 100 | rw | 5 | loadbalance | | | T |

CP … Constraint Programming
MIP … Mixed Integer Programming
Id … Identification number of instance
T … Maximum allowed computation time exploited (3.600 seconds)
rw … real world
rand … random
bold characters … optimal solution found


	\end{table}
	%\vfills
	
\end{appendices}

\end{document}